%% file: main.tex
\documentclass[11pt,a4paper]{article}

\usepackage[T1]{fontenc}
\usepackage[utf8]{inputenc}
\usepackage[american]{babel}
\usepackage{lmodern}
\usepackage{microtype}
\usepackage{textcomp}
\usepackage{csquotes}

\usepackage{amsmath}
\usepackage{amssymb}
\usepackage{mathtools}
\usepackage{bm}

\usepackage{graphicx}
\usepackage{caption}
\graphicspath{{figures/}}
\usepackage{booktabs}
\usepackage{array}
\usepackage{longtable}
\usepackage{calc}
\usepackage{wrapfig}
\usepackage[normalem]{ulem}          

\makeatletter
\newsavebox\pandoc@box
\newcommand*\pandocbounded[1]{%
  \sbox\pandoc@box{#1}%
  \Gscale@div\@tempa{\textheight}{\dimexpr\ht\pandoc@box+\dp\pandoc@box\relax}%
  \Gscale@div\@tempb{\linewidth}{\wd\pandoc@box}%
  \ifdim\@tempb\p@<\@tempa\p@\let\@tempa\@tempb\fi
  \ifdim\@tempa\p@<\p@\scalebox{\@tempa}{\usebox\pandoc@box}%
  \else\usebox{\pandoc@box}\fi}
\makeatother

\usepackage[margin=2.5cm]{geometry}
\usepackage{parskip}
\usepackage{xcolor}
\usepackage{hyperref}
\hypersetup{hidelinks}

\usepackage[backend=biber,style=numeric,sorting=none]{biblatex}
\usepackage{newunicodechar}
\newunicodechar{∙}{\ensuremath{\cdot}}
\newunicodechar{•}{\ensuremath{\bullet}}
\newunicodechar{→}{\ensuremath{\rightarrow}}
\newunicodechar{√}{\ensuremath{\surd}}
\newunicodechar{≤}{\ensuremath{\leq}}
\newunicodechar{≥}{\ensuremath{\geq}}
\newunicodechar{≠}{\ensuremath{\neq}}
\newunicodechar{≈}{\ensuremath{\approx}}
\newunicodechar{∈}{\ensuremath{\in}}
\newunicodechar{∣}{\ensuremath{\mid}}
\newunicodechar{∑}{\ensuremath{\sum}}
\newunicodechar{∏}{\ensuremath{\prod}}
\newunicodechar{ᵀ}{\ensuremath{^{\top}}}
\newunicodechar{Δ}{\ensuremath{\Delta}}
\newunicodechar{Θ}{\ensuremath{\Theta}}
\newunicodechar{Σ}{\ensuremath{\Sigma}}
\newunicodechar{μ}{\ensuremath{\mu}}
\newunicodechar{σ}{\ensuremath{\sigma}}
\newunicodechar{τ}{\ensuremath{\tau}}
\newunicodechar{η}{\ensuremath{\eta}}

\title{AI at a Turning Point\\[0.4em]\large From ChatGPT through Neural Networks to Neuromorphic AI}

\author{Volker Kempe}

\date{}

\begin{document}
\maketitle
\begin{abstract}
    The tension between the established technologies of connectionist AI and their neuromorphic challengers is attracting growing attention, above all because of the enormous energy demand of the former and the low power consumption of the latter. Developing and running large AI models consumes gigawatt- to terawatt-hours of energy, while the human brain operates on a continuous 20 watts. Spiking neural networks (SNNs) counter this energy demand with a frugal approach and bring elements of self-organization  into the methodological repertoire of AI: local learning instead of global error backpropagation, event-driven computing, and time as a carrier of information. They do not replace today’s AI but extend it; in hybrid systems each approach can be deployed where its strengths lie. This monograph explains the field in a generally accessible yet mathematically rigorous way. Using the text-generation module of ChatGPT as an example, the principle, elegance, and computational cost of transformer-based AI are presented and compared with the coding, learning, and network-design methods of SNNs. Throughout, relevant mechanisms of the human nervous system serve as model  and reference point. Topics include spike-timing-dependent plasticity, population and temporal coding, the Neural Engineering Framework, neuromorphic hardware from Loihi 2 to commercial edge processors, and the prospects for classical and neuromorphic AI to complement and extend each other.

\medskip

\noindent
{\small\itshape
This article is an English translation of the German original
by the same author, entitled
\enquote{KI am Wendepunkt: Von ChatGPT
\"uber neuronale Netze zu neuromorpher KI}, Version 4, October 2026, 
DOI \url{https://doi.org/10.5281/zenodo.23278918}, licensed under the Creative Commons
license CC BY-NC-ND 4.0
(\url{https://creativecommons.org/licenses/by-nc-nd/4.0/}).\\[6pt]}

\end{abstract}

\bigskip

\input{chapters/01_einleitung}

\input{chapters/02_chatgpt}

\input{chapters/03_neuronale_netze}

\input{chapters/04_spiking_neural_networks}

\input{chapters/05_zusammenfassung}

\newpage

\nocite{russell2020aima,bear2018neurowissenschaften,krauss2023ki}
\printbibliography[notkeyword=weiterfuehrend,heading=bibintoc,title={References}]
\printbibliography[keyword=weiterfuehrend,heading=bibintoc,title={Further Reading}]

\vspace{8cm}
\noindent\rule{\textwidth}{0.4pt}\par
\vspace{0.8em}
\noindent
\begin{minipage}{\textwidth}\small
Translated from German by Vera Kempe with the assistance of Claude
Fable 5.1 (Anthropic). The entire translation was carefully reviewed, edited  and verified by both the translator and the original author.\\[6pt]

Apart from the figures produced by the author, some images were created
with the support of generative AI from prompts formulated by the author
and were subsequently reviewed and edited.\\[6pt]
\end{minipage}

\newpage

\tableofcontents
\end{document}

%% file: chapters/01_einleitung.tex
\section{1. Introduction}

\begin{center}

\emph{Learning → Knowledge → Intelligence → Intelligent Action}

\end{center}

This simplified causal chain describes an essential path by which biological systems acquire knowledge through interaction with their environment and apply it to intelligent action. Its most complex known manifestation to date is found in humans. Historically, artificial intelligence arose from the effort to recreate the links of this chain, that is, the abilities to learn, to understand, to acquire knowledge, and to apply it meaningfully.

Building on medieval notions of homunculi, and subsequently all manner of automata that imitated sequences of human actions, leading scientists of the last century took up this baton. Formal logic, explicit knowledge, symbolic representations, inference mechanisms, search algorithms, planning and expert systems, methods of abstraction, and more would be brought to bear on problems that had previously seemed accessible only to human intelligence. These efforts produced results that, only half a century earlier, would have been consigned to the realm of distant fantasy.

Crucial for this progress was a powerful tool: the electronic computer, which from the middle of the last century onwards became the dominant instrument for translating human knowledge and thought into computational processes. The foundation for this was the knowledge acquired over the course of human evolution and continually expanded by human thinking. Its implementation was dominated by digital computation based on logic circuits following the rules of Boolean algebra. Advances in AI have been fueled by ever-increasing computing power, to the point where the energy demands of training large models can now reach levels comparable to the energy consumption of smaller cities.

To produce natural intelligence, nature had to take the opposite path: bottom-up, not top-down. The driving forces were self-organization and evolution. First to emerge were mechanisms of learning and knowledge acquisition. Knowledge could be ordered, interconnected, abstracted, and compressed until systems reached a degree of complexity and organization that we today associate with intelligence. This bottom-up approach shapes the path to the emergence of natural intelligence. At the start of this path is learning as the mechanism of interaction with the environment; at its end, an increasingly generalized knowledge of objects, relations, regularities, and purposes.

Under the pressure of evolution and through processes of self-organization, nature created the human brain as tool and vehicle of natural intelligence: an astonishing web of elementary building blocks for learning and knowledge acquisition. Its basic functional unit is the neuron. Much as the transistor is the basic element of electronic circuits, the neuron is the basic element of neural networks, yet each neuron on its own already exhibits incomparably greater functional complexity. In a technical implementation, some of its individual functions can be reproduced only through the interplay of several electronic components; entire neural networks amount to highly complex, dynamically interacting systems.

Such analogies reveal the remarkable fact that the rapid progress in AI over recent decades has important roots in principles deliberately borrowed from natural processes in living organisms.

As early as 1943, Warren S. McCulloch and Walter Pitts \cite{mcculloch1943logical} laid important foundations for the mathematical description of neural information processing. They showed that networks of idealized neurons can carry out logical functions, thereby creating an essential foundation for the development of artificial neural networks.

In his book published in 1949 \cite{hebb1949organization}, Donald Hebb then formulated the principle that later became known as the Hebbian learning rule:

\emph{``When an axon of cell A is near enough to excite a cell B and repeatedly or persistently takes part in firing it, some growth process or metabolic change takes place in one or both cells such that A's efficiency, as one of the cells firing B, is increased.''}

In a memorable slogan coined later \cite{shatz1992developing}, this principle is often summarized as

\begin{center}

\textbf{``Neurons that fire together, wire together.''}

\end{center}

One of the foundational insights behind artificial neural networks was that learning can be implemented by changing the strength of connections between neural units. Hebb's rule supplied a biologically motivated local learning principle for this. However, how such weight changes could be deployed in artificial networks to solve specific tasks initially remained largely unsolved.

The Hebbian learning rule can thus be regarded as one of the important roots of connectionist AI, which has shaped modern AI in crucial ways.

The Dartmouth conference of 1956 \cite{mccarthy2006dartmouth} is regarded as the birth of artificial intelligence as a research field in its own right. The decades that followed were initially dominated by symbolic AI. Neural networks kept a place in research laboratories but did not achieve widespread practical use. Critical studies of what the artificial neural networks of the day could actually do further dampened the expectations that had been raised. This precipitated the first so-called ``AI winter''.

After some further development and a second ``AI winter,'' machine learning increasingly took center stage. In their groundbreaking 1986 paper \emph{``Learning representations by back-propagating errors''} \cite{rumelhart1986learning}, Rumelhart, Hinton, and Williams described a powerful algorithm that adjusts the internal weights of multilayer neural networks by propagating an error signal backward through the network. This provided a fundamentally different way to optimize the network weights in a targeted fashion with respect to global error. This breakthrough did much to render artificial neural networks fit for practical applications and prepared the ground for the deep learning revolution of the 2000s and 2010s.

The gradient-based backpropagation mechanism became one of the most important pillars of modern AI. It determines the directions of change in network parameters that reduce a loss function, and thus allows the network to adapt step by step toward a learning objective. With growing model size, increasing volumes of data, and ever longer training runs, however, the computational cost becomes enormous. As a consequence, AI factories with vast data centers are being built. Although backpropagation is extraordinarily successful, the combination of global gradient optimization with ever larger models, immense quantities of data, and gigantic data centers increasingly resembles brute force. Given the energy and resources it consumes, one may well ask whether this is the only route to ever more capable artificial intelligence.

Yet advances of recent decades, above all in neurobiology and cognitive science, show ever more clearly that nature has further principles for shaping learning processes at its disposal; harnessing them in technology may offer substantially more efficient solutions for many tasks.

Prominent among them is a neuronal mechanism established with experimental precision in the 1990s, in which the change in synaptic efficacy depends on the \textbf{causally connected temporal order} of pre- and postsynaptic neuronal events: spike-timing-dependent plasticity (STDP) \cite{Bi1998Synaptic,markram1997regulation}.

The underlying principle can be condensed into another slogan:

\begin{center}

\textbf{``Fire together, wire together.''}\\
\textbf{``Out of sync, weaken the link.''}

\end{center}

Here a local learning rule, operating without central external intervention, is tied directly to the temporal events in the neural network. Classical artificial neural networks have largely lost exactly this: the explicit role of individual events and their timing. Exploiting it directly requires a representation that preserves the temporal dynamics of neuronal events. This casts the spotlight on event-based neural models, and spiking neural networks in particular.

The experimental characterization of STDP thus helped to give rise to biologically oriented learning mechanisms in modern spiking neural networks.

These mechanisms hold the potential for a fundamentally different mode of operation. Temporal dynamics, short-term memory, and the immediate effect of co-occurring neuronal events can make the network an active co-designer of its own configuration. This opens the door to an alternative developmental path for AI, one that does not replace the path taken so far but extends it. Whether this will lead to a fundamental shift, how far and how fast such a development can advance, depends not least on an understanding of its potential and on confidence in the possibilities for its technical implementation.

What follows is intended to give an interested readership a clear picture of the characteristic capabilities of classical modern AI, represented by a selected example, and to compare them with what spiking neural networks have achieved so far and where their potential lies. The emphasis is on the efficiency of the two approaches.

This exposition is set against the background of a discussion of selected operating principles of the human nervous system. That discussion serves a second purpose: to encourage reflection on which further insights from neurobiology and cognitive science might be harnessed for the development of neuromorphic systems.

My aim has been to write the following account in a generally accessible way while, for readers with a deeper interest, allowing the argument to be followed in greater depth through the integration of technical and mathematical detail. This balancing act may make some sections not entirely easy to read. For this I ask the reader's indulgence.

%% file: chapters/02_chatgpt.tex
\section{2. ChatGPT}

Let us begin the comparison by examining one of the most representative examples of modern AI. The system chosen is an LLM-based language generation system (LLM: large language model), represented here by the text generation part of OpenAI's chatbot ``ChatGPT.''

ChatGPT is among the most capable current systems built on large artificial neural networks, which is why it strikes me as particularly well suited to an illustrative comparison with neuromorphic approaches.

\subsection{2.1 Text processing}

At the core of every chatbot are its text processing and text generation modules. They are what make communication with the system feel natural to a human at all.

A distinctive feature of the task is that the system must infer linguistic structures and semantic relations from the data themselves. From the outset, pure language models lack the immediate anchoring of linguistic concepts in the physical world that humans acquire through perception and experience.

A person hearing the word ``house'' associates with it a wealth of sensory experiences, mental images, and memories. These associations are the product of that person's own interaction with the environment. A language model has no access to such a world of first-hand experience. It initially learns the concept ``house'' from the relations in which the word and related concepts occur in its training data.

Even so, astonishingly rich semantic representations can arise in this way. The model captures relations between concepts without ever having seen, touched, or experienced the things they refer to as a human has. It is precisely this missing immediate anchoring, known as grounding, that makes the comparison between human language understanding and machine language processing so interesting.

Among the historical starting points of today's chatbots were early attempts in computational linguistics to reproduce linguistic communication with formalized rules, patterns, and dialogue structures. In the mid-1960s, Joseph Weizenbaum caused a sensation with ELIZA, one of the first widely known natural-language dialogue systems \cite{weizenbaum1966eliza}.

The system worked essentially using keywords, pattern matching, and pattern substitution. It was unable to learn, and it had no deeper semantic understanding.

A particularly well-known dialogue configuration was one called ``DOCTOR,'' which simulated a conversation in the style of a non-directive psychotherapy session.

It was striking how quickly people came to ascribe meaning and understanding to the machine-generated replies, and in some cases even placed their trust in the system.

The public's amazement at ELIZA's seemingly intelligent behavior, and the discussion it ignited about the possibility of ``making machines intelligent,'' set off a series of debates that continues to this day: about the nature of intelligence, about whether it can have a place in the real ``world of machines,'' about its potential and prospects, and finally about the opportunities and dangers of further development.

This typical reaction of society to achievements in artificial intelligence has changed remarkably little since then. Amazement, incomprehension, or half-comprehension of AI and of its present and future role, along with optimistic or pessimistic projections, still characterize the public mood even today. This stands in the way of sober, balanced conclusions, and at present it frequently produces both inflated expectations and an underestimation of possible dangers, up to and including the malicious exploitation of a loss of control over powerful AI systems.

In the development of machine language processing, rule-based and symbolic methods gave way increasingly to statistical methods and finally to artificial neural networks (ANNs). Words, and later larger linguistic units, came to be represented as embeddings, vectors in continuous spaces where linguistic and semantic relations could be captured mathematically.

Recurrent neural networks, and later LSTM (long short-term memory) networks in particular, made it possible to process temporal and sequential dependencies. The introduction of attention mechanisms was a further crucial step; one influential version was proposed in 2014 by Dzmitry Bahdanau, Kyunghyun Cho, and Yoshua Bengio \cite{bahdanau2015neural}.

The actual breakthrough to a new network architecture came in 2017, when Vaswani et al.~introduced the transformer architecture \cite{vaswani2017attention} under the pointed title:

\begin{center}

\textbf{``Attention Is All You Need''}

\end{center}

While processing one element of a sequence, the attention mechanism allows the model to weight its relations to the other elements selectively. Relevant connections can thus be taken into account even across large distances. At the same time, sequences no longer had to be processed strictly step by step through a recurrent network. Training could be parallelized far more extensively, which became a crucial precondition for scaling up today's large language models.

Why ChatGPT?

The abbreviation GPT stands for \emph{Generative Pre-trained Transformer}: a generative, pre-trained model built on the transformer architecture. ``Generative'' means that the model does not merely classify or analyze existing data; it can produce new content on the basis of learned structures.

Figuratively, one might call the transformer a master of relationships. It does not process the elements of a context in isolation but relates them to one another through attention. During training, the statistical and semantic structures learned in this way are condensed into the parameters of the network, where they remain available for generating new text.

The transformer architecture was presented in 2017 by researchers at Google. With GPT-1 in 2018, OpenAI applied the principle to a generative, pre-trained language model and developed the approach further over several generations.

The breakthrough in public awareness came at the end of 2022 with ChatGPT, which combined an already mature language model technology with a dialogue interface anyone could use. The text generation part thus became the interface between the mathematical world of the model and human language.

It was, in essence, precisely the part that turned a powerful AI technology into a system directly usable by millions of people.

Since ChatGPT's breakthrough, numerous comparable systems have followed. Large language models are now used for information retrieval, text processing, programming, analysis, and many other tasks, and are increasingly combined with additional tools and multimodal capabilities.

The example of ChatGPT was chosen above all because such systems have spread like an avalanche across vast areas of application, reaching far into private life. Large language models are trained on vast quantities of text and other data and, depending on the system and its technical integration, can also draw on external information sources and tools.

They thus increasingly shape communication and organization in government, business, science, and society, and not everywhere for the better. Texts and scholarly works can be produced with AI support; marketing is a major field of application; and the aggregation of large amounts of information can support decision-making. At the same time, generative methods open new avenues for abuse. Deepfakes, particularly through image, video, and speech generation, are a widely known example.

The following explanation of the purpose, structure, and function of the text generation module is meant to give readers who do not work in the field a brief glimpse into the conceptual world of the AI community.

I have deliberately not written this book in the language of insiders; it is tailored to readers with a general scientific education. Nor does it retrace, step by step, the history of the developers' reasoning, which I have deliberately not followed in detail. Rather, my aim is to present the system, stripped of excess jargon, so that its basic mode of operation can be understood by essentially anyone.

\subsection{2.2 Brief refreshers}

Before turning to the structure of ChatGPT, it is necessary to settle on a few terms and to review some basics of artificial neural networks.

The artificial neurons used in classical artificial neural networks adopt only a narrow selection of the properties of natural neurons. Chief among them is the idea that the effects of several input signals are combined, and that the neuron is activated depending on the result. A second biologically inspired idea is that learning can proceed through changes in the connection strengths between neurons. The Hebbian learning rule, as mentioned earlier, provided an early biological model for this. Later research extended this picture substantially, as discussed further below.

An important foundation of artificial neural networks is the perceptron, developed from 1957 onward by Frank Rosenblatt \cite{rosenblatt1958perceptron} on the basis of earlier work (including that of McCulloch and Pitts), and its later extensions (Figs. 2.1a and 2.1b).

\begin{figure}[htbp]
\centering
\begin{minipage}[t]{0.4\textwidth}
\centering
\includegraphics[width=\linewidth]{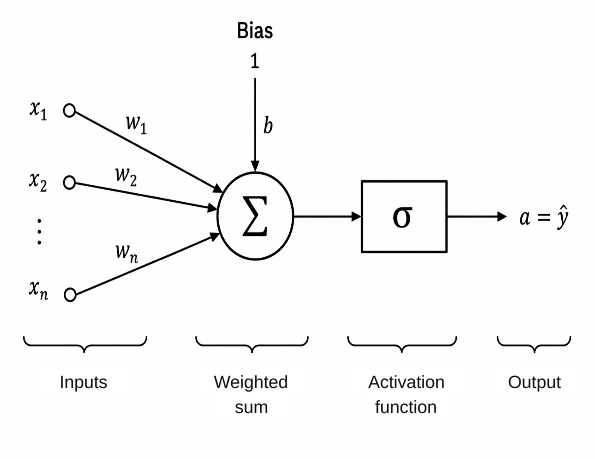}
\captionof*{figure}{Fig. 2.1a The perceptron}
\end{minipage}\hfill
\begin{minipage}[t]{0.54\textwidth}
\centering
\includegraphics[width=\linewidth]{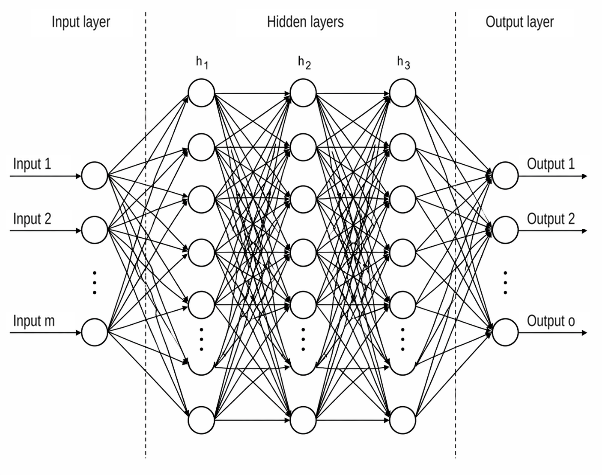}
\captionof*{figure}{Fig. 2.1b Network detail}
\end{minipage}
\end{figure}

In such networks, the capacity to learn comes from adjusting the connection weights. For the classical perceptron, the adjustment follows the perceptron learning rule:
\begin{equation*}
\Delta \mathbf{W}
 = \eta \cdot (\text{target} - \text{estimate})
   \cdot \text{input},
\qquad
\eta = \text{learning rate}.
\end{equation*}
It is not the same as the Hebbian learning rule, but it shares with it the basic idea that learning can be realized by modifying connection strengths.

In a first step, the artificial neuron computes the weighted sum of its input values. The weights are the crucial trainable parameters of the network (see the network detail in the right-hand figure). Adding a bias \(b\) shifts the operating point of the activation function and can thereby influence, for example, the input sum at which a neuron becomes appreciably activated. The bias, too, is trainable.

The output value is then determined by the activation function \(\sigma( \cdot )\). For a node with index \(j\):
\begin{equation}\tag{2.2.1}
y_j =
\sigma\!\left(
\sum_{i=1}^{n} W_{ij}x_i + b_j
\right).
\end{equation}
The original perceptron used a threshold function. Later artificial neural networks use a variety of activation functions, many of them continuous or differentiable.

A frequently used function is the ReLU (rectified linear unit):
\begin{equation*}
\operatorname{ReLU}(x)=\max(0,x).
\end{equation*}
It sets all values for \(x \leq 0\) to zero and passes positive values through unchanged.

Transformer models frequently use the GELU function (Gaussian error linear unit) \cite{hendrycks2016gelu}; it has been described for GPT-3, for example. It is defined as
\begin{equation*}
\operatorname{GELU}(x)=x\,\Phi(x),
\end{equation*}
where \(\Phi(x)\) is the cumulative distribution function of the standard normal distribution. Intuitively, GELU is a smooth, input-dependent weighting: negative values are not cut off abruptly but increasingly damped, while large positive values pass through almost unchanged. Its smoothness is an advantage for gradient-based optimization.

For what follows, it is convenient to write the artificial neuron in vector form. Individual scalar inputs and outputs are replaced by vectors, and the connection weights are collected in a weight matrix. This gives the compact notation
\begin{equation*}
\mathbf{y}
=
\sigma\!\left(
\mathbf{W}\mathbf{x}+\mathbf{b}
\right),
\end{equation*}
where \(\mathbf{x}\) is the input vector, \(\mathbf{y}\) the output vector, \(\mathbf{W}\) the weight matrix, and \(\mathbf{b}\) the bias vector. The nonlinear activation function \(\sigma( \cdot )\) is applied componentwise to the resulting vector.

From here on, vectors are set in bold lowercase and matrices in bold uppercase.

We will meet this vector notation again and again when examining the transformer. There, vectors are mapped into new representations by multiplication with trainable matrices. The query, key, and value vectors introduced later also arise in this way, through linear projections. For a single token they are written as vectors \(\mathbf{q}_{i}\), \(\mathbf{k}_{i}\), and \(\mathbf{v}_{i}\); collected for a whole token sequence, as matrices \(\mathbf{Q}\), \(\mathbf{K}\), and \(\mathbf{V}\).

Input and output representations play an important role in neural networks. The input data must first be converted into a mathematical representation the network can process. The network proper, whether simple, recurrent, or many-layered as in deep learning, processes this representation and finally produces an output, which must in turn be related back to the task at hand.

In supervised training, the output is compared with a given target, and an error or loss function \(L\) is computed from the difference. Optimization methods, usually gradient-based, then modify the trainable parameters so that the loss decreases step by step. Backpropagation computes the gradients of the loss function with respect to the individual trainable parameters by tracing dependencies backward through the network, allowing them to be adjusted.

Translating a real-world problem into a suitable mathematical representation, and designing the network architecture that goes with it, are among the key tasks in developing AI systems. The true art is not simply to let a network learn, but first to transfer the real task into the network's mathematical world in a way that makes learning possible at all.

This demands both a deep understanding of the available AI tools and extensive knowledge of the application domain. This is precisely where domain knowledge and AI knowledge meet.

\subsection{2.3 ChatGPT: structure and function}

For automated chat systems, the interface between user and system is of crucial importance. To humans, natural language is far more than a string of characters: words and sentences trigger associations and activate knowledge acquired through perception, experience, and learning.

As noted, pure language models lack this immediate anchoring of linguistic concepts in a personally experienced world. Modern language models instead extract, from vast quantities of text, the syntactic, semantic, and structural relations manifested in language use. Their fundamental training principle is to determine, from a given context, the probability distribution over its possible continuations.

The entire structure of an autoregressive text generation model is geared toward learning such context-dependent predictions. During training, the model receives a sequence of text elements and learns, for each position, to predict the next element from those preceding it. The element to be predicted is never part of the input used to predict it.

To manage this, the model must capture relations between the elements of a text sequence. The attention mechanism plays the central role. It produces context-dependent representations by assessing, for each text element, which other elements of the available context matter for its processing.

In the structures that can then form in a high-dimensional representation space, similar information, or information used in comparable contexts, is represented by geometrically related vectors. How such structures and clusters develop is examined more closely later.

The concept of attention-driven neural networks thus proved a crucial step on the way to today's large language models.

\subsubsection{2.3.1 The token concept}

For a neural network to process natural language, the language must first be decomposed into discrete, numerically addressable units. These units are called tokens.

Depending on the tokenization method, a token may correspond to a word, a part of a word, a single character, or a special character. The word ``unbelievable,'' for example, might be decomposed into several sub-tokens. Which decomposition actually occurs depends on the specific tokenizer and its vocabulary.

Each token in the vocabulary carries a unique identification number, its token ID. Typical token vocabularies range from a few tens of thousands to over a hundred thousand entries.

\textbf{Vocabulary and embedding matrix:}

Let \(V\) be the number of tokens in the vocabulary. Each token with ID \(j\) is assigned a row of a trainable embedding matrix \(\mathbf{E}\):
\begin{equation*}
\mathbf{E}
=
\begin{pmatrix}
\mathbf{e}_1^{\top}\\
\vdots\\
\mathbf{e}_j^{\top}\\
\vdots\\
\mathbf{e}_V^{\top}
\end{pmatrix},
\qquad
\mathbf{E}\in
\mathbb{R}^{V\times d_{\text{model}}}.
\end{equation*}
The vector
\begin{equation*}
\mathbf{e}_j
=
(E_{j1},\ldots,E_{j\,d_{\text{model}}})^{\top}
\end{equation*}
is the learnable initial representation of the token with ID \(j\). Figuratively, the embedding matrix can be pictured as a large ``embedding shelf.'' Every token, through its ID, has a fixed compartment. What lies in the compartment, however, is not the token itself but its currently 
learned vector \(\mathbf{e}_{j}\).

Provided no special tokenizer is used (which we assume here for the sake of simplicity), the embedding matrix contains no learned language structure at the start of training. Its values are initialized and then modified by training, together with the model's other parameters.

\textbf{The goal of the entire text generation system is to adjust these representations of all tokens, step by step, so that the next token can be predicted as reliably as possible from any given text context. The foundation for building these representations is the attention mechanism described below.}

In the course of this process, representations arise whose geometric relations increasingly reflect regularities of language use.

\textbf{The positioned token sequence:}

A token sequence
\begin{equation*}
(t_1,t_2,\ldots,t_L)
\end{equation*}
has two fundamentally different organizational attributes. Every occurring token has

\begin{enumerate}

\item

an \emph{identity}, given by its token ID, and

\item

a \emph{position} \(i\) within the current sequence.

\end{enumerate}

This distinction is essential for what follows. The same token can occur at different positions in a sequence, and it initially always receives the same embedding vector from \(\mathbf{E}\). Its position, and later its specific context, then cause its representation to change during processing.

For a given token sequence, the corresponding rows of the embedding matrix are first selected:

\begin{equation*}
\mathbf{x}_i^{(0)}
=
\mathbf{e}_{\mathrm{ID}(t_i)}.
\end{equation*}
This yields the input matrix

\begin{equation}\tag{2.3.1}
\mathbf{X}^{(0)}
=
\begin{pmatrix}
(\mathbf{x}_1^{(0)})^{\top}\\
\vdots\\
(\mathbf{x}_L^{(0)})^{\top}
\end{pmatrix},
\qquad
\mathbf{X}^{(0)}
\in
\mathbb{R}^{L\times d_{\text{model}}}.
\end{equation}
And with it, a remarkable transition takes place:

\begin{center}

\emph{Text is translated into geometry.}

\end{center}

The initially discrete linguistic symbols become points, or vectors, in a continuous high-dimensional space. Training shapes the geometry of this space so that learned relations within language can be represented in it.

The global embedding matrix \(\mathbf{E}\) must be distinguished, however, from the representations of a specific token sequence. The embedding matrix is a store of trainable parameters of the whole model. The vectors of the current sequence, by contrast, form the current working state, which the successive transformer blocks process further.

For the following account it is useful to distinguish, as a didactic organizing scheme, between three domains:

\begin{enumerate}

\item

the \textbf{vocabulary space}, comprising the discrete tokens and their IDs,

\item

the \textbf{representation space}, in which tokens are represented by vectors of dimension \(d_{\text{model}}\),

\item

the \textbf{model or working space}, in which these representations are modified, depending on context, by attention and further transformations.

\end{enumerate}

These are not mathematically separate spaces in the strict sense; they serve to distinguish conceptually the different functions within the model.

The vocabulary space provides the link to the discrete linguistic symbols. The representation space translates these symbols into continuous vectors. In the model or working space, context-dependent representations are then derived from these.

At the model's output, the reverse step takes place: a context-dependent representation is mapped onto a vector of scores over the tokens of the vocabulary. Applying the softmax function (see below) turns this into a probability distribution over possible next tokens. The loop is thus closed: from the discrete token, via its high-dimensional representation, back to the vocabulary.

\textbf{Positional information:}

So far, the assignment of a token to its embedding vector carries no information about where in the sequence the token occurs. This positional information must be supplied to the model separately.

In classical absolute positional encoding, each token vector is therefore assigned a position vector \(\mathbf{p}_{i}\):
\begin{equation*}
\widetilde{\mathbf{x}}_i^{(0)}
=
\mathbf{e}_{\mathrm{ID}(t_i)}
+
\mathbf{p}_i.
\end{equation*}
The input representation then carries information both about the token's identity and about its position in the sequence.

The technical specifics of how positional information is introduced depend on the particular transformer architecture. Besides absolute position vectors that are added to the embeddings, relative positional encodings and rotary position embeddings (RoPE) are in use today. Common to all of them is the goal of making the order and spacing of the tokens accessible to the attention mechanism.

\subsubsection{2.3.2 Processing a token sequence}

The next question is what length of token sequences can be processed efficiently for training. In principle, one could slide a window over a very long token sequence and run a separate training step at each window position. But this would repeat many nearly identical computations and make poor use of parallel processing.

Instead, long text corpora are cut into sequences, or training segments, of a fixed length \(L\). Several such sequences are collected into a batch and processed in parallel within one training step. The contributions to the loss function from the individual sequences and token positions are combined and used to adjust the model parameters jointly.

The actual sequence length and batch size depend on the model, the hardware, and the training strategy. In today's large language models, a single training step can involve very large numbers of tokens and correspondingly vast quantities of activation values. Since the exact training parameters of proprietary models such as GPT-4 have not been fully published, speculative figures are avoided here. What matters is the order of magnitude: many token sequences are processed simultaneously to exploit the transformer architecture's high degree of parallelizability.

What, then, does a model need if it is to predict a meaningful continuation from a given context?

The underlying working hypothesis is as simple as it is far-reaching: in very large quantities of natural language, regularities occur with sufficient frequency. They include grammar, syntax, and orthography, but also semantic relations, factual connections, typical patterns of argument, and many further regularities that can only partly be stated explicitly.

If more and more of these relations can be brought to bear on prediction from the context, the probability of a correct and meaningful continuation should rise as well. The model must therefore learn to capture as many prediction-relevant relations as possible in its parameters, and to activate them when processing a specific context.

\textbf{The probabilistic estimate:}

An autoregressive language model assigns a conditional probability to every possible next token. For a token sequence
\[
t_1,t_2,\ldots,t_L
\]
the joint probability can be decomposed exactly, by the chain rule of probability theory, as

\begin{equation}\tag{2.3.2}
P(t_1,t_2,\ldots,t_L)
=
\prod_{i=1}^{L}
P(t_i\mid t_1,\ldots,t_{i-1}),
\end{equation}
where \(P\left( t_{1} \mid \varnothing \right) = P\left( t_{1} \right)\).

As it stands, this equation describes an exact decomposition of a joint probability and not yet a Bayesian estimate in the narrower sense. From a stochastic point of view, however, training can be understood as an estimation problem: the model should approximate the unknown conditional distributions of natural language as well as possible by parameterized probability distributions

\[
P_{\theta}(t_i\mid t_1,\ldots,t_{i-1})
\]
The parameter vector \(\theta\) comprises all trainable model parameters.

Crucially, a training sequence is not used to predict just its final token. Causal masking makes it possible to generate, for every position \(i\) simultaneously, a prediction of the next token from the context before it:
\[
t_1
\rightarrow t_2,
\qquad
(t_1,t_2)
\rightarrow t_3,
\qquad \ldots \qquad
(t_1,\ldots,t_{L-1})
\rightarrow t_L.
\]
The transformer computes these predictions in parallel during training. The causal attention mask (see below) ensures that each position can access only the tokens preceding it. The token to be predicted thus remains invisible to its own prediction and serves purely as the target value.

This is the essence of the principle of \emph{self-supervised learning}: the target values need not be labeled by humans; they come directly from the text itself.

One distinction matters here: the model does not try to reproduce the components of the original token's embedding vector directly. At its output, it produces, for each prediction position, a distribution over the entire token vocabulary. The token that actually follows is uniquely identified by its token ID and serves as the target for that distribution.

Every position thus yields a prediction error. The contributions of all the positions evaluated, and of all sequences in a batch, are combined into one loss function. Minimizing it adjusts the model parameters so that the tokens which actually follow receive increasingly high probabilities.

From a stochastic point of view, evaluating many such conditional predictions in parallel can be understood as learning from a large number of local samples of the language distribution. The individual training segments supply samples for optimizing one shared global parameter set. Under suitable statistical conditions (near-ergodicity), the model thereby approaches a probability distribution that captures the regularities in the training data increasingly well.

By optimizing these conditional probabilities, the model therefore learns far more than the isolated prediction of single tokens. It builds a high-dimensional statistical representation of linguistic relations, and of the structural relations contained in the texts, which every new prediction can draw on depending on the context.

How such a context-dependent representation is built inside the transformer, and what the attention mechanism contributes to it, is the subject of the next section.

\subsubsection{2.3.3 Self-attention}

The key to processing a token sequence lies in effectively learning and using the relations that hold among its elements. The crucial concept for this is self-attention, as mentioned several times already.

\begin{center}

\emph{It acts as a context-dependent relevance filter, determining which information from other tokens matters most for the processing of a given token.}

\end{center}

Its implementation is one of the essential prerequisites for the performance of modern transformers.

\textbf{Projection into different representation spaces:} Let the current token sequence be represented by the matrix

\[
\mathbf{X}\in\mathbb{R}^{L\times d_{\text{model}}}
\]
Each of its \(L\) rows holds the current representation of one token position.

For self-attention, these representations are projected into new spaces by trainable linear transformations. Multi-head attention uses \(h\) such projections in parallel, often with

\[
d_{\text{head}}=\frac{d_{\text{model}}}{h}.
\]
Each head can thus view the token sequence through a different learned projection. Figuratively speaking, the heads provide different geometric ``viewpoints'' of the same token sequence.

These views are not assigned to particular linguistic relations in advance. Projections that prove useful for minimizing the loss function form only during training. Individual heads may then come to capture, say, syntactic, semantic, or position-related relations preferentially. Such functions, however, need not be cleanly separated from one another nor confined to a single relation each.

For clarity, we first consider a single head and omit the head index. Each current token representation \(\mathbf{x}_{i}\), yields three vectors:

\begin{equation}\tag{2.3.3}
\mathbf{q}_i=\mathbf{x}_i\mathbf{W}_Q,
\qquad
\mathbf{k}_i=\mathbf{x}_i\mathbf{W}_K,
\qquad
\mathbf{v}_i=\mathbf{x}_i\mathbf{W}_V.
\end{equation}
For the whole token sequence these equations then become

\[
\mathbf{Q}=\mathbf{X}\mathbf{W}_Q,
\qquad
\mathbf{K}=\mathbf{X}\mathbf{W}_K,
\qquad
\mathbf{V}=\mathbf{X}\mathbf{W}_V.
\]
where

\[
\mathbf{W}_Q,\mathbf{W}_K,\mathbf{W}_V
\in
\mathbb{R}^{d_{\text{model}}\times d_{\text{head}}}
\]
are trainable projection matrices.

The three projections serve different functions. Informally, they can be interpreted as follows:

\begin{itemize}

\item

The query vector \(\mathbf{q}_{i}\) describes what information position \(i\) is ``looking for'' in this head.

\item

The key vector \(\mathbf{k}_{j}\) describes what kind of information position \(j\) ``offers'' for such a comparison.

\item

The value vector \(\mathbf{v}_{j}\) holds the informational contribution that, given sufficient relevance, is passed on from position \(j\) to position \(i\).

\end{itemize}

These labels are merely intuitive interpretations. Mathematically, they are learned linear projections of the current token representations.

\textbf{Determining relevance:} The relation between the querying position \(i\) and a potential information source \(j\) is scored by the dot product

\begin{equation}\tag{2.3.4}
\mathbf{q}_i\cdot\mathbf{k}_j
\end{equation}

The vector \(\mathbf{q}_{i}\) contains the \(d_{\text{head}}\) projections of the current representation of token \(t_{i}\) (which in the first block is its position-encoded embedding vector); \(\mathbf{k}_{i}\) and \(\mathbf{v}_{i}\) arise in the same way from the same representation. The dot product measures how well the representations of the two tokens \(t_{i}\) and \(t_{j}\) match (their similarity) once projected into the query and key spaces, and thus provides the quantitative basis of attention: a graded score of how strongly two tokens are related.

Recall that, geometrically,

\[
\mathbf{q}_i\cdot\mathbf{k}_j
=
\lVert\mathbf{q}_i\rVert
\lVert\mathbf{k}_j\rVert
\cos\Theta_{ij}.
\]
The value therefore depends both on how closely the directions agree and on the lengths of the two vectors. Training the matrices \(\mathbf{W}_{Q}\) and \(\mathbf{W}_{K}\) transforms the original representation space so that relations useful for the task at hand can produce high or low dot products.

The scaled attention scores are

\begin{equation}\tag{2.3.5}
s_{ij}
=
\frac{\mathbf{q}_i\cdot\mathbf{k}_j}
{\sqrt{d_{\text{head}}}},
\qquad
\mathbf{S}
=
\frac{\mathbf{Q}\mathbf{K}^{\top}}
{\sqrt{d_{\text{head}}}}.
\end{equation}
The factor \(\sqrt{d_{\text{head}}}\) prevents the dot products from growing too large, as they typically would as the dimension increases, which would push the subsequent softmax function into regions of very small gradients.

The score \(s_{ij}\) can be read as a measure of how relevant the information offered at position \(j\) is for the processing of position \(i\).

The attention relation is inherently directed. In general,

\[
s_{ij}\neq s_{ji},
\]
since query and key are produced by different learned transformations. A token \(t_{j}\) can therefore matter greatly for the processing of \(t_{i}\) without \(t_{i}\) having the same importance for \(t_{j}\).

The scores are not permanently stored model parameters. They are computed afresh for every specific token sequence. What is learned permanently are the transformation matrices \(\mathbf{W}_{Q}\) and \(\mathbf{W}_{K}\), which determine what relations are sought and which relations are emphasized.

Attention thus combines two properties: the learned projection matrices represent generalizable mechanisms for detecting relevant relations, while the scores computed from them remain entirely dependent on the context at hand.

\textbf{From scores to attention weights:} The scores of a row are normalized with the softmax function:

\begin{equation}\tag{2.3.6}
a_{ij}
=
\frac{\exp(s_{ij})}
{\sum_{k=1}^{L}\exp(s_{ik})}.
\end{equation}
For every position \(i\), then,

\[
a_{ij}\geq0,
\qquad
\sum_{j=1}^{L}a_{ij}=1.
\]
In matrix form,

\begin{equation}\tag{2.3.7}
\mathbf{A}
=
\operatorname{softmax}
\left(
\frac{\mathbf{Q}\mathbf{K}^{\top}}
{\sqrt{d_{\text{head}}}}
\right),
\end{equation}

with the softmax function applied row by row.

The values \(a_{ij}\) form a normalized \emph{attention distribution}. They must not be confused, at this point, with the later probability distribution over possible next tokens.

The exponential transformation of the softmax function amplifies differences between the scores and turns them into relative weights. At the same time, the softmax function is invariant to a common shift of all scores in a row:

\[
\operatorname{softmax}(s_{i1}+c,\ldots,s_{iL}+c)
=
\operatorname{softmax}(s_{i1},\ldots,s_{iL}).
\]
What matters, then, are the relative differences among the scores of a row.

Formally, this normalized exponential form is closely related mathematically to a Gibbs-Boltzmann distribution. Within the attention mechanism, however, it serves first of all as an adaptive distribution of weights over the available information sources.

\textbf{Context-dependent centers of gravity:} The attention weights take effect in the weighted sum of the value vectors:

\begin{equation}\tag{2.3.8}
\mathbf{z}_i
=
\sum_{j=1}^{L}
a_{ij}\mathbf{v}_j.
\end{equation}
In matrix form,

\begin{equation}\tag{2.3.9}
\mathbf{Z}
=
\mathbf{A}\mathbf{V}.
\end{equation}
Since the weights \(a_{ij}\) are nonnegative and sum to one for each row, \(\mathbf{z}_{i}\) can be interpreted geometrically as a context-dependent weighted center of gravity, a barycenter, of the value vectors.

Every token position produces its own such center. Strongly weighted value vectors contribute correspondingly more to the result than weakly weighted ones.

Here it becomes especially clear how learned relevance relations give rise to specific mixtures of information:

\begin{center}

\emph{Query and key determine where information should come from;\\ the values determine what information is actually transferred.}

\end{center}

The matrix \(\mathbf{W}_{V}\) is not a mapping back into the original representation space. It merely produces the value representations of one head.

\textbf{Multi-head attention:} A single head provides only one learned view of the relations in the token sequence. For this reason, \(h\) heads are used in parallel. Each head \(r\) has its own projection matrices:

\[
\mathbf{Q}^{(r)}
=
\mathbf{X}\mathbf{W}_Q^{(r)},
\qquad
\mathbf{K}^{(r)}
=
\mathbf{X}\mathbf{W}_K^{(r)},
\qquad
\mathbf{V}^{(r)}
=
\mathbf{X}\mathbf{W}_V^{(r)}.
\]
From these, each head produces its own context matrix

\begin{equation}\tag{2.3.10}
\mathbf{Z}^{(r)}
=
\mathbf{A}^{(r)}\mathbf{V}^{(r)}.
\end{equation}
The results of all heads are concatenated for each token position:

\begin{equation}\tag{2.3.11}
\mathbf{Z}_{\text{cat}}
=
\operatorname{Concat}
\left(
\mathbf{Z}^{(1)},\ldots,\mathbf{Z}^{(h)}
\right).
\end{equation}
With \(h \cdot d_{\text{head}} = d_{\text{model}}\), the concatenated matrix again has width \(d_{\text{model}}\).

The contributions of all heads are then mixed by the trainable output matrix \(\mathbf{W}_{O}\):

\begin{equation}\tag{2.3.12}
\mathbf{F}
=
\mathbf{Z}_{\text{cat}}\mathbf{W}_O.
\end{equation}
The matrix \(\mathbf{W}_{O}\) lets the model learn how the different views obtained by the heads are to be used jointly in further processing.

The permanently learned matrices \(\mathbf{W}_{Q}^{(r)}\), \(\mathbf{W}_{K}^{(r)}\), \(\mathbf{W}_{V}^{(r)}\), and \(\mathbf{W}_{O}\) thus encode general mechanisms of information selection and processing. The specific attention matrices \(\mathbf{A}^{(r)}\), by contrast, arise afresh for every token sequence and represent the context-dependent activation of these learned mechanisms.

\textbf{Respecting temporal order:} For an autoregressive language model, the representation at position \(i\) may use information only from positions \(j \leq i\). Future tokens must remain invisible to that position.

A causal mask achieves this. For a token sequence with five positions it may look like this:

\[
\mathbf{M}
=
\begin{pmatrix}
0 & -\infty & -\infty & -\infty & -\infty\\
0 & 0       & -\infty & -\infty & -\infty\\
0 & 0       & 0       & -\infty & -\infty\\
0 & 0       & 0       & 0       & -\infty\\
0 & 0       & 0       & 0       & 0
\end{pmatrix}.
\]
The mask is added to the scores before the softmax function:

\begin{equation}\tag{2.3.13}
\widetilde{\mathbf{S}}^{(r)}
=
\mathbf{S}^{(r)}+\mathbf{M}.
\end{equation}
This gives
\[
\mathbf{A}^{(r)}
=
\operatorname{softmax}
\left(
\mathbf{S}^{(r)}+\mathbf{M}
\right).
\]
and consequently, for all positions \(j > i\),
\[
a_{ij}^{(r)}=0.
\]
In computing its representation, the token at position \(i\) thus cannot ``see'' any of the tokens that follow it.

What makes the transformer efficient is that no separate truncated context window has to be computed for each position. A single masked matrix operation produces the context-dependent representations for all positions simultaneously.

Throughout these row-wise transformations, the subscript \(i\) keeps its meaning as the position marker of the token sequence. The representation itself changes from block to block, but its assignment to position \(i\) is preserved.

\textbf{Residual connection and feedforward network:} The matrix \(\mathbf{F}\) produced by multi-head attention again has dimension \(L \times d_{\text{model}}\). It can now be added to the input representation \(\mathbf{X}\) through a residual connection. In simplified form,

\begin{equation}\tag{2.3.14}
\mathbf{X}_{\text{att}}
=
\operatorname{Norm}
\left(
\mathbf{X}+\mathbf{F}
\right).
\end{equation}
The exact order of normalization and subunit depends on the particular transformer architecture; modern GPT architectures typically use variants of pre-layer normalization (pre-LN). For the functional understanding pursued here, what matters is that \textbf{the residual path preserves the existing representation and adds the contextual information gained by attention.}

The representation enriched in this way is then passed to a feedforward network operating position by position. In simplified form, it can be written as

\[
\operatorname{FFN}(\mathbf{x})
=
\mathbf{W}_2\,
\sigma(\mathbf{W}_1\mathbf{x}+\mathbf{b}_1)
+
\mathbf{b}_2.
\]
The first transformation typically expands the representation from \(d_{\text{model}}\) to a larger inner dimension \(d_{\text{ff}}\). After a nonlinear activation function is applied, the result is projected back to \(d_{\text{model}}\). Attention and FFN together form the transformer block as shown in Fig.2.2.

While attention combines information across different token positions, the feedforward network processes each position independently, with the same trained parameters. It enables nonlinear transformations of the previously contextualized features and thus contributes substantially to the further processing of the representations.

Here, too, the result is combined with the subunit's input through a residual connection. In simplified form,

\[
\mathbf{H}
=
\operatorname{Norm}
\left(
\mathbf{X}_{\text{att}}
+
\operatorname{FFN}(\mathbf{X}_{\text{att}})
\right).
\]
The matrix \(\mathbf{H}\) then holds, for every token position, a further processed, context-dependent representation. 
In a transformer with several blocks, \(\mathbf{H}\) is at the same time the input of the next block. Only the output of the final block forms the model's final hidden-state representation.

At this point, a distinction drawn earlier becomes especially important:

\begin{center}

\emph{The current token representations are not written back into the embedding matrix.}

\end{center}

The matrices \(\mathbf{X}\), \(\mathbf{F}\), and \(\mathbf{H}\) are activations of the particular token sequence being processed and exist only for that computational pass. The embedding matrix \(\mathbf{E}\), by contrast, belongs to the central, permanently trainable parameters of the model.

Only during backpropagation are gradients computed from the loss function that reach back to the rows of the embedding matrix actually used. The optimization algorithm then changes their parameters. What is stored is thus not the current context-dependent token representation; rather, training gradually changes the global parameters from which future representations are produced.

\textbf{Parallel prediction during training:} A key advantage of causal masking is that a single forward pass simultaneously produces training signals for many positions of a sequence.

From the final hidden state \(\mathbf{h}_{i}\) at position \(i\), the logits for the following token are computed:
\[
\boldsymbol{z}_i
=
\mathbf{h}_i\mathbf{W}_U+\mathbf{b}_U.
\]
Applying the softmax function turns them into a probability distribution over the entire vocabulary:
\[
P_{\theta}
\left(
t_{i+1}
\mid
t_1,\ldots,t_i
\right).
\]
A sequence of length \(L\) thus yields predictions for all admissible positions at once:
\[
t_1\rightarrow t_2,\qquad
(t_1,t_2)\rightarrow t_3,\qquad
\ldots,\qquad
(t_1,\ldots,t_{L-1})\rightarrow t_L.
\]
\begin{wrapfigure}{r}{0.46\textwidth}
\vspace{-25pt}
\centering
\includegraphics[width=\linewidth]{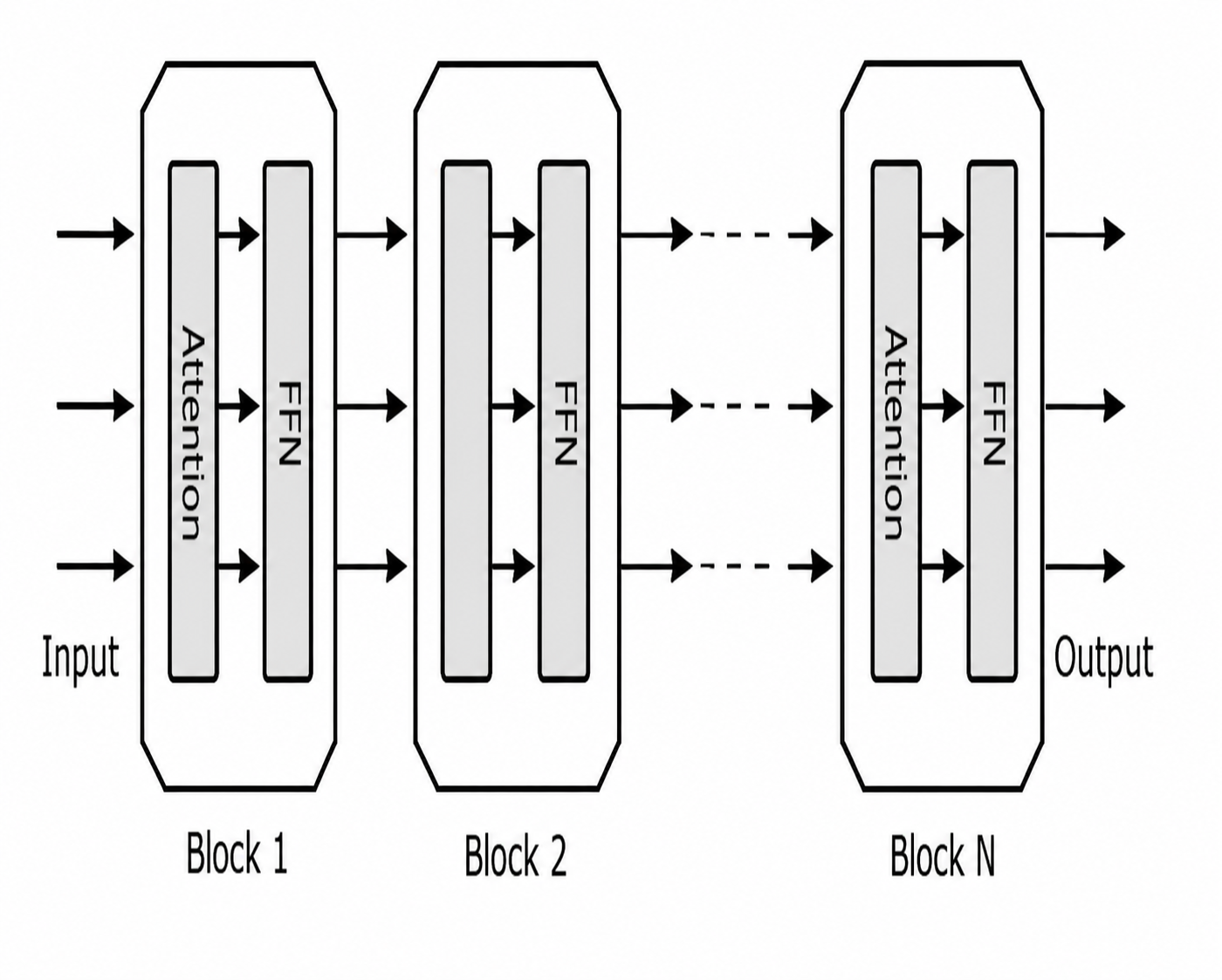}
\vspace{-25pt}
\caption*{Fig. 2.2 Transformer network}
\vspace{-20pt}
\end{wrapfigure}
The matrix operations for attention, multi-head processing, feedforward networks, and output projections can be executed in highly parallel fashion for all positions. The causal mask nevertheless guarantees that each individual prediction can access only the past context admissible for it.

The transformer thus combines two seemingly opposed principles: it models a strictly directed, autoregressive sequence of conditional probabilities, yet computes the representations this requires largely in parallel during training.

\subsubsection{2.3.4 Training}

The system is trained step by step, in an interplay of computing the activations, measuring the error, and correcting the parameters. Its effectiveness depends crucially on how well these steps are coordinated.

Self-attention alone does not yet produce a complete representation of language. In every transformer block, its results are processed further by a feedforward network operating position by position and are then handed on to the next block. There is no rigid sequence in which each block detects exactly one linguistic property. Even so, analyses of neural networks frequently show that increasingly complex and abstract representations can form with growing depth. Local patterns, syntactic relations, and longer-range semantic structures interact to a growing degree.

\textbf{Mapping back to the vocabulary space:} What, then, must happen at the output of the last transformer block?

The last transformer block delivers, for every position \(i\), a context-dependent hidden state \(\mathbf{h}_{i}\). This still lies in the continuous representation space of dimension \(d_{\text{model}}\). The decision as to which token to predict next, however, must be made in the discrete vocabulary space.

To that end, each hidden state is mapped by an output transformation onto a vector with \(V\) components. For the whole sequence, this gives
\begin{equation}\tag{2.3.15}
\mathbf{Z}=\mathbf{H}\mathbf{W}_U+\mathbf{b},
\end{equation}
where each row \(\mathbf{z}_{i}\) holds the logits for the \(V\) possible tokens of the vocabulary. The bias \(\mathbf{b}\) is not essential for basic understanding and is disregarded from here on.

In many language models, the output matrix is coupled with the embedding matrix, known as \emph{weight tying} \cite{press2017tying}. Then \(\mathbf{W}_{U} = \mathbf{E}^{\top}\), so that for a position \(i\) the logits arise as \(z_{ij} = \mathbf{h}_{i} \cdot \mathbf{e}_{j}\). The hidden state is thereby compared with all token representations of the vocabulary.

These dot products are not yet probabilities. It is the softmax function that then produces, for every position, a normalized distribution
\begin{equation}\tag{2.3.16}
p_{ij}
=
\frac{\exp(z_{ij})}
{\sum_{k=1}^{V}\exp(z_{ik})},
\qquad
\sum_{j=1}^{V}p_{ij}=1.
\end{equation}
Here \(p_{ij}\) is the model's estimated probability that the token with vocabulary ID \(j\) follows the context available up to position \(i\). Semantically, this corresponds to
\[
p_{ij}
=
P_{\theta}
\left(
t_{i+1}=j
\mid
t_1,\ldots,t_i
\right).
\]
The separation between position space and vocabulary space is especially important here. The index \(i\) continues to denote the position in the current token sequence, while the index \(j\) denotes a possible entry of the entire vocabulary.

\textbf{The learning objective:} For every prediction position, the token that actually follows is unambiguously identified through its token ID. The target distribution is therefore a one-hot vector \(\mathbf{w}_{i}\): for the ID of the correct next token, \(w_{ij} = 1\); for all other tokens, \(w_{ij} = 0\).

The loss for the prediction at position \(i\) is then the cross-entropy
\begin{equation}\tag{2.3.17}
L_i
=
-\sum_{j=1}^{V}
w_{ij}\log p_{ij}
=
-\log
P_{\theta}
\left(
t_{i+1}
\mid
t_1,\ldots,t_i
\right).
\end{equation}
\textbf{\emph{This is one of the expressions central to the functioning of the whole system.}}

Since many positions are evaluated simultaneously during a training pass, the total loss for a sequence is the sum, or the mean, of the individual prediction losses. In simplified notation,
\begin{equation}\tag{2.3.18}
L
=
-\sum_{i=1}^{L-1}
\log
P_{\theta}
\left(
t_{i+1}
\mid
t_1,\ldots,t_i
\right).
\end{equation}
This equation is the negative log-likelihood of the observed token sequence. Training thus tries to assign the highest possible probability to the tokens that actually follow, given each token's context.

The target distribution itself never changes: a token keeps the same ID throughout training. What changes are the parameters of the model and, insofar as the embedding matrix is trained, also the geometric representations of the tokens.

\textbf{How is the error propagated backward?} It can be shown that for the combination of softmax and cross-entropy the derivative takes a particularly simple form:
\begin{equation}\tag{2.3.19}
\frac{\partial L_i}{\partial z_{ij}}
=
p_{ij}-w_{ij}.
\end{equation}
For the correct token, this value is \(p_{ij} - 1\) and hence negative. For all wrong tokens, it equals \(p_{ij}\) and hence is positive.

This expression is the starting point of the backward computation. For an uncoupled output matrix \(\mathbf{W}_{U}\), for example, a contribution to the gradient of its \(j\)-th token vector is proportional to \(\left( p_{ij} - w_{ij} \right)\mathbf{h}_{i}\).

With weight tying, a finer distinction is needed. The same embedding matrix \(\mathbf{E}\) is then used both to form the input representations and to produce the output logits. The expression \(\left( p_{ij} - w_{ij} \right)\mathbf{h}_{i}\) therefore describes only the contribution to the change of \(\mathbf{E}\) that arises through the output projection. In addition, further gradients propagate back through the entire transformer chain on those rows of the embedding matrix whose tokens served as input.

This is also the crucial point announced earlier:

\begin{center}

\emph{The context-dependent token representations produced during a forward pass are not written back into the embedding matrix.}

\end{center}

The current hidden states vanish after each computational pass. Only backpropagation computes, from the loss, gradients for the permanently stored model parameters. The optimization algorithm then changes these parameters, and with them the embedding matrix.

In geometric terms, the output contribution of the gradient strengthens the logit relation between the hidden state and the correct token and, correspondingly, weakens the competing wrong tokens in relative terms. Under simple gradient descent, the corresponding output contribution to the update of a token vector is
\[
\mathbf{e}_j
\leftarrow
\mathbf{e}_j
-
\eta\,
(p_{ij}-w_{ij})\mathbf{h}_i.
\]
For the correct token, this contribution acts in the direction of \(\mathbf{h}_{i}\); for competing tokens, in the opposite direction.

One may picture this geometric effect as suitable representations moving toward one another. It is, however, not a Hebbian learning rule. The crucial difference is that the change here is determined by the global prediction error and its backpropagated gradient.

The hidden state itself also receives a gradient. In simplified form, the output contribution is

\begin{equation}\tag{2.3.20}
\frac{\partial L_i}{\partial\mathbf{h}_i}
=
\sum_{j=1}^{V}
(p_{ij}-w_{ij})\,\mathbf{e}_j.
\end{equation}
From here, the error is carried back step by step through all preceding transformer blocks. In this way, \(\mathbf{W}_{Q}\), \(\mathbf{W}_{K}\), \(\mathbf{W}_{V}\), \(\mathbf{W}_{O}\), the feedforward weights, the normalization parameters, and finally the embeddings are changed jointly so that future predictions improve.

The knowledge that arises during training is therefore not stored in any single place. Neither the embedding matrix nor the attention matrices nor any individual weight matrix contains the acquired knowledge by itself. It is distributed over the totality of the trained model parameters.

One might therefore call this \emph{world-pattern knowledge} distributed across the network: a system of statistical, semantic, syntactic, and structural relations, condensed into billions of parameters, that is activated anew each time a specific context is processed.

It is this difference between permanently stored world-pattern knowledge and the merely transient context-dependent activations that is of fundamental importance for understanding the system.

\textbf{Forward and backward pass:}

Determining the gradients first requires a complete forward pass, in which the activations and the loss are computed. In the subsequent backward pass, the derivatives of the loss function are carried back through the entire network.

The backward pass is typically considerably more expensive than a pure forward pass, because the matrix operations must additionally produce gradients with respect to the inputs and the weight matrices. As a rough order of magnitude, a complete training step of forward plus backward computation is often put at about three times the cost of pure inference. The actual ratios, however, depend strongly on architecture, hardware, and implementation.

\textbf{From the probability vector to the next token:}

When the trained system is used, the end result is again a probability distribution over possible next tokens. One possibility, with weight tying and without bias, would simply be to choose the token with the highest probability:
\begin{equation}\tag{2.3.21}
\hat{t}_{i+1}
=
\arg\max_j p_{ij}
=
\arg\max_j
\left(
\mathbf{h}_i\cdot\mathbf{e}_j
\right).
\end{equation}
Such a pure maximum decision, however, is only one possible decoding strategy. Modern language models frequently also use stochastic selection procedures, for example temperature-controlled sampling or restricting the choice to the most probable tokens. Text generation can thereby be made less deterministic and linguistically more varied.

It should be added that a capable pretrained language model alone does not yet make a complete chatbot. The actual pre-training is followed by further phases of what is known as post-training. These can include supervised fine-tuning, learning from human or synthetically generated preference signals, and reinforcement learning or other optimization procedures.

To these are added system components for dialogue management, safety rules, tool use, information retrieval, and the integration of external results. Only this interplay turns a pretrained language model into a practically usable chat system.

At this point, the basic functional description of the text generation module is complete enough to allow for functional comparisons with biological neural systems. First, however, a closer look at the training cost is needed, as it plays an important part in the arguments that follow.

\subsubsection{2.3.5 Illustration and summary}

An intuitive overall picture may help in visualizing the learning process. In simplified terms, the model can be viewed as a high-dimensional, malleable web of relations. The token representations form its points, while the attention mechanisms establish varying relations between these points for each specific context. The gradient produced by the loss function is propagated backward through the whole network and changes the parameters so that future passes yield a geometry more favorable to prediction.

The frequently used image of an elastic net with springs is genuinely helpful here, but must not be taken literally: the attention weights are not permanently existing connections. They are computed afresh for every context. What is permanently changed are the parameters of the model, which govern how these context-dependent relations form.

Visualization techniques support an intuitive picture of the learning process. High-dimensional representations can be projected into two- or three-dimensional spaces with dimension-reduction methods. In many instances, structures such as groupings, clusters, or preferred directions can then be visualized in this way.

\begin{figure}[htbp]
\centering
\begin{minipage}[t]{0.48\textwidth}
\centering
\includegraphics[width=\linewidth]{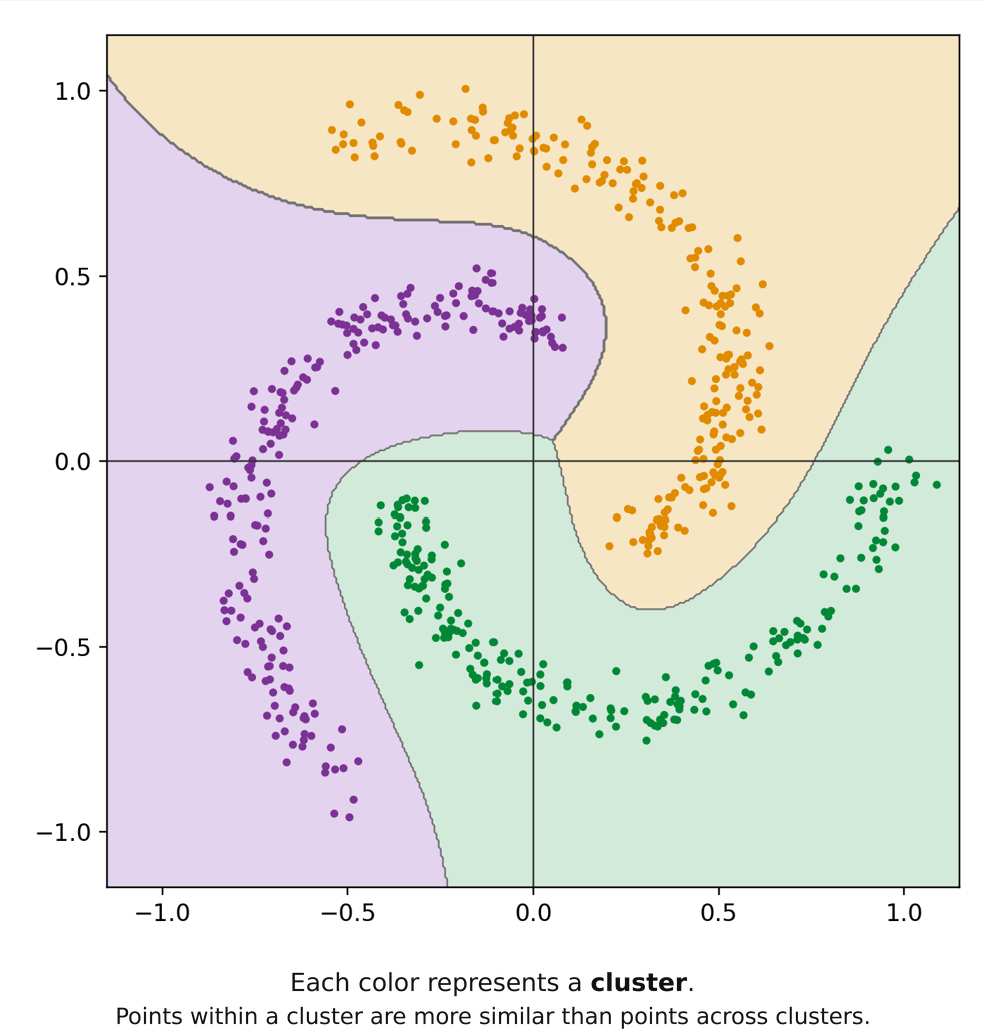}
\captionof*{figure}{Fig. 2.3 Schematic representation of cluster formation in the representation space}
\end{minipage}\hfill
\begin{minipage}[t]{0.48\textwidth}
\centering
\includegraphics[width=\linewidth]{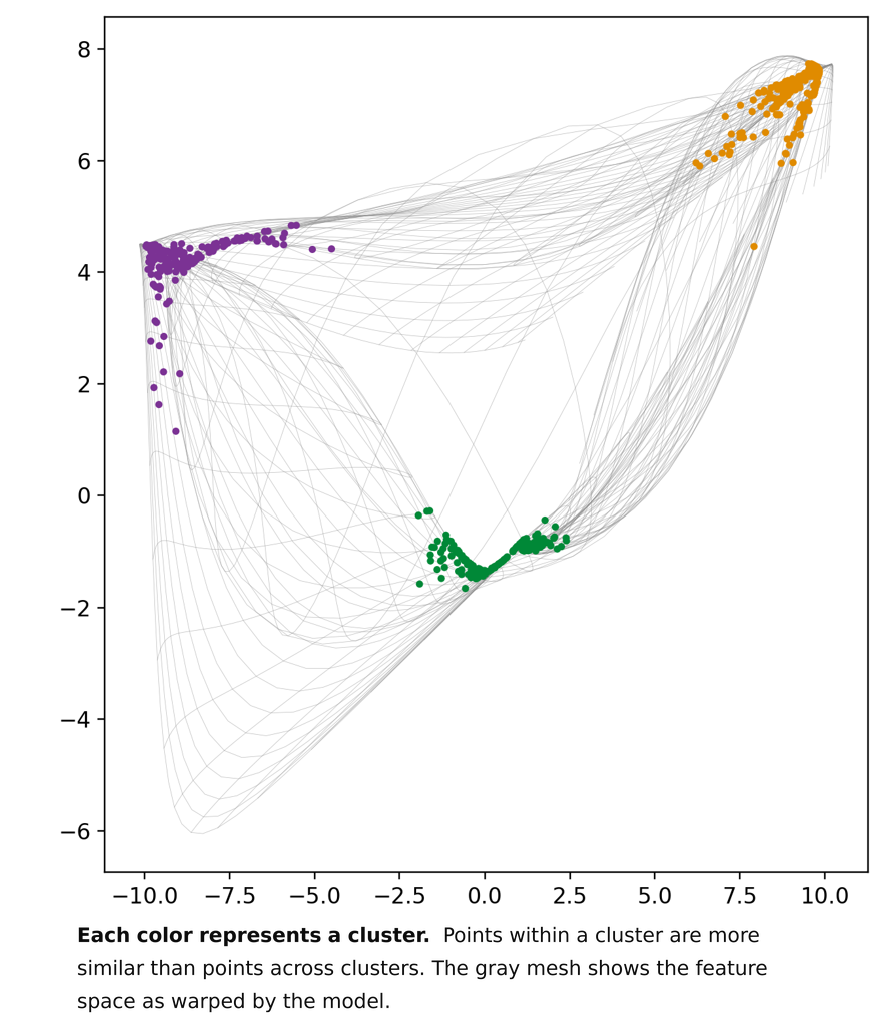}
\captionof*{figure}{Fig. 2.4 Schematic cluster structure with separating regions}
\end{minipage}
\end{figure}

Figures 2.3 and 2.4 are to be understood as schematic illustrations of such structure formation. In real language models, geometric structures can correspond to syntactic, semantic, or other contextual relations.

Representations change along the blocks of a transformer sequence. Increasingly complex and more context-dependent features can emerge. This can be interpreted as a progressive process of \textbf{contextual abstraction}.

Nowhere are these emergent structures explicitly stored as finished rules, concepts, or clusters. They result from the interplay of the trained parameters and are created anew each time a specific context is processed. This is a remarkable property of modern neural networks: knowledge need not appear as an explicitly stored rule; it can appear as the capacity of the whole system to produce suitable representations under specific conditions.

The model's \emph{world-pattern knowledge} introduced above is therefore not comparable to a lexicon or a knowledge database. It is better conceptualized as the potential --- distributed across the entire network of parameters --- to generate the appropriate relations and representations for a given context.

Remarkably, in high-dimensional representation spaces, one can sometimes discern directions and structures that correlate with particular properties of the processed information, although these properties were never provided to the model explicitly as rules. Such emerging structures indicate that large neural networks can capture regularities in data whose explicit formulation need not even be known to humans.

In autoregressive language models, the attention mechanism is always used under the optimization pressure of a prediction task. Therefore, the relations that are learned preferentially are precisely those useful for predicting a next token. If the same basic mechanism were trained with a different learning objective, different weights and representations could form.

This highlights the universality of the attention mechanism, but also the crucial importance of the learning objective. Attention determines which pieces of information are put in relation to one another in a given context; the loss function ultimately determines which of these relations prove useful in training and are therefore reinforced through changes in the model parameters.

\textbf{Summary:}

The text processing module of ChatGPT served to illustrate some characteristic procedures of modern AI built on large artificial neural networks.

The transformer combines a mechanism for the context-dependent selection and combination of information with nonlinear feedforward networks. This leads to the emergence of increasingly complex representations across many successive blocks. Training directs the entire process toward a goal that is simple to state: to predict --- as well as possible --- the probability distribution of the following token from the context available at each point.

From a probabilistic point of view, the model thus learns an approximation of the conditional distributions of language. Figuratively speaking, the autoregressive prediction mechanism serves as the conductor of the learning process: it sets the goal that guides the formation of the internal representations.

Much of the transformer's technical elegance stems from not having to enforce the temporal causality of prediction through strictly sequential computation of the training steps. Rather, causal masking makes it possible to compute the conditional predictions for numerous positions of a token sequence in parallel.

This concept thus reflects an attempt to make structures emerging during learning immediately usable for further information processing. In this respect, it shows an interesting parallel to natural learning processes, in that structures once acquired influence the processing of new information.

The difference remains, however, fundamental. Today's gradient-based training requires a globally defined optimization process whose error signal is propagated back through large parts of the network. Biological nervous systems, by contrast, organize learning to a considerable extent through local plasticity mechanisms, modulatory signals, and hierarchically coupled control loops.

It is precisely this contrast that leads to one of the central questions of this exposition: how much of the enormous global optimization effort of modern AI is truly unavoidable, and how much could be replaced by systems that derive a greater part of their organization from local, self-organizing learning mechanisms?

\subsubsection{2.3.6 The cost}

Reported data on the computational and energy costs of large language models carry considerable uncertainty, since from GPT-3 onward the developers have published only part of the technical data required. GPT-3 therefore offers a particularly suitable reference point: its key model specifications were published and allow at least an order-of-magnitude estimate of the training cost at the time \cite{Brown2020FewShot,patterson2021carbon,T.Jonas2023}.

\begin{longtable}[]{@{}
  >{\raggedright\arraybackslash}p{0.50\linewidth}
  >{\raggedright\arraybackslash}p{0.45\linewidth}@{}}
\toprule
Parameter & Values for GPT-3 \\
\midrule
\endhead

Dimension of the model space $d_{\mathrm{model}}$
    & 12,288 \\

Dimension of the FFN $d_{\mathrm{ff}}$
    & $4\,d_{\mathrm{model}} = 49{,}152$ \\

Number of transformer blocks (layers)
    & 96 \\

Model type
    & dense autoregressive transformer \\

Total number of trainable parameters
    & 175 billion ($1.75\cdot10^{11}$) \\

Number of training tokens
    & about 300 billion \\

Total training compute
    & about $3.14\cdot10^{23}$ FLOP \\

Rough split of the training cost
    & about 1/3 forward and 2/3 backward computation \\

Training hardware
    & cluster of NVIDIA V100 GPUs on Microsoft infrastructure \\

Training energy
    & on the order of 1 GWh, depending on system boundary and
      method of calculation \\

\bottomrule
\end{longtable}

Roughly calculated, the training of GPT-3 processed about
\[
\frac{300\ \mathrm{billion\ tokens}}
     {175\ \mathrm{billion\ parameters}}
\approx 1.7
\]

training tokens per parameter. The Chinchilla scaling results published a few years later showed that, for a given compute budget, considerably smaller models trained on considerably more tokens can often be trained more economically \cite{J.Hoffmann2022Training}. The order of magnitude found there, about 20 tokens per parameter, is not to be understood as a general law of nature or a minimum. Still, the comparison shows that GPT-3, by today's standards, was a heavily parameterized and, by later scaling results, undertrained model.

Even at this stage, the state of the art required extraordinary resources. The pure training compute of GPT-3 was about \(3.14 \cdot 10^{23}\) FLOP. Estimates of the associated energy consumption depend strongly on whether only the computing hardware proper is counted, or also memory, communication, cooling, and further data-center losses. As an order of magnitude for the training run itself, an energy demand of somewhat over 1 GWh can be assumed \cite{patterson2021carbon}. Even then, training a single model equaled the annual electricity consumption of hundreds of European households.

For GPT-4 and GPT-5, no comparably complete official figures on parameter count, training data, compute, and energy consumption exist. Extrapolating from GPT-3's training to later model generations is speculative. The assumption sometimes made that GPT-5, simply because it belongs to a later generation, could have required a training cost about 100 or 200 times that of GPT-3, is not tenable given the efficiency gains achieved in the meantime. Recent independent analyses \cite{cottier2024rising,emberson2026compute} even conclude that the training of GPT-5 probably used less compute than that of GPT-4.5, because a larger share of the performance gain could be achieved through improved post-training procedures.

This by no means implies that the resource problem has disappeared. Rather, there are currently two competing developments. On the one hand, new GPU generations, reduced numerical precision, optimized algorithms, better training data, specialized model architectures, and new post-training procedures are improving efficiency considerably.

On the other hand, the efficiency gains are reinvested in larger models, more extensive training and post-training procedures, and a rapidly growing number of experiments and applications. The cost of developing a frontier model is therefore no longer just the cost of a single final training run. On top of that come numerous preliminary experiments, parameter studies, data preparation, synthetic data generation, and post-training. Then there is the ongoing post-development inference cost for what are now millions of users.

The historical development from GPT-3 to today's systems thus shows a notable tension: the individual computational step, and the model performance obtainable from a given amount of computing power, keep becoming more efficient, while at the same time the total computing power devoted to artificial intelligence keeps growing at an extraordinary rate.

The energy demand of modern AI is therefore less and less captured by the training consumption of a single model alone. Rather, it is reflected in the construction of an extensive physical infrastructure of data centers, power grids, energy generation, and cooling systems.

A current example from Austria shows how directly this development now affects real infrastructure. Google's first large data center in the country is currently being built in Kronstorf in Upper Austria. For the first construction stage, which is already under way, the stated maximum power draw is 150 MW; for a possible later full build-out, it is up to 500 MW. For later stages, an expected consumption of several terawatt-hours of electricity per year has been mentioned. A single data-center site would thus consume roughly as much electricity as hundreds of thousands of households.

The example also makes clear that the effect of AI is not confined to the visible function of a language model, an image generator, or an AI assistant. Behind the seemingly immaterial world of algorithms, a very material infrastructure of data centers, power grids, power plants, cooling systems, and land use is taking shape. In Kronstorf, the debate therefore now covers not only the economic opportunities of the project but also its electricity and water demand, land sealing, local temperature increases, and possible environmental effects.

Illustrative comparisons help bring home these orders of magnitude. A continuous electrical power demand of 500 MW would correspond, at full utilization, to an annual energy consumption of 4.38 TWh. For comparison: the human brain runs on only about 20 W. This juxtaposition is of course no scientific efficiency comparison, since biological and artificial systems have entirely different tasks, structures, and capabilities. But it conveys a sense of the amounts of energy now involved.

None of this is meant to diminish the extraordinary capabilities of modern language models and other AI systems, which have by now proven themselves in countless applications. The enormous development cost is, moreover, spread over a very large number of uses, and the economic calculation behind this development has, at least so far, evidently paid off.

The development does, however, make the question of the efficiency of the underlying computing principles ever more pressing. The larger the infrastructure required for advancing and applying AI becomes, the stronger the incentive becomes not to seek performance gains solely through the further concentration of conventional computing power.

Efficiency gains in hardware and methods remain indispensable means of limiting resource demand. More capable processors, improved algorithms, and more efficient model architectures show that energy demand need not grow in proportion to the performance of AI. However, whether these advances alone will suffice remains an open question, given the simultaneously surging demand for AI computing power. Efficiency gains frequently enable new and even larger applications and therefore do not necessarily reduce absolute resource consumption.

The very magnitude of the investments in AI factories and data centers should therefore itself be a strong reason to step up research into fundamentally different, more energy-efficient computing principles. The question is thus not whether today's development of large AI systems is successful, which it undoubtedly is, but whether the ever-stronger concentration of computing power in AI gigafactories can remain the only path to further performance gains in the long run.

The search for such alternatives leads directly to the approaches discussed in the following chapters, which make the relevant insights of recent decades into the structure, information processing, and learning of natural neural systems usable for artificial ones.

%% file: chapters/03_neuronale_netze.tex
\section{3. Neural networks}

Since the inception of AI, understanding the human nervous system has been an important source of new ideas, despite the extraordinary difficulties associated with coaxing this highly complex system into revealing its hidden mechanisms and the precise quantitative laws that govern them.

Our knowledge of the brain and nervous system has expanded enormously in recent decades. Research is focusing increasingly on the time course and strength of the interactions between neurons, between different types of synapses, between patterns of neural activity, and between whole brain areas.

Even a few of the basic mechanisms of neural learning give a sense of the options that neuromorphic approaches offer for artificial emulation. Neural information processing has always evolved under the joint demands of performance, efficiency and economy; it certainly holds great potential for substantially expanding connectionist AI and for lowering some of the hurdles that current learning methods face. The following sections are devoted to such mechanisms and to other processes that can be gleaned from biology.

\subsection{3.1 A paradigm shift in AI?}

Before we immerse ourselves in the world of biological nervous systems, it is worth asking first what the consequences of emulating them in artificial systems might be.

The text generation module of ChatGPT is an impressive demonstration of the basic ideas of modern ANN-based AI. Objects, tokens, and other units of information are represented by high-dimensional vectors and repeatedly related to one another in new ways through learned transformations. Different architectures and learning objectives permit different forms of representation and abstraction.

In the course of this development, artificial neurons evolved from comparatively simple processors of scalar input signals into the building blocks of vast mathematical transformation systems. High-dimensional vector sequences are transformed linearly, combined in a context-dependent way and then undergo further nonlinear processing.

Accordingly, large artificial neural networks have become high-performance mathematical optimization machines. A given learning objective is formalized in a loss function, and billions of parameters are modified so that this loss function is minimized step by step. At its mathematical core, backpropagation is in fact nothing more than the systematic application of the chain rule to an enormous system of nested functions.

As astonishing and elegant as the solutions achieved in this way are, the engineering cost grows with model size and task complexity. As emphasized in the previous chapter, the expansion of the required computing infrastructure has itself become a major factor in the development of modern AI.

Yet over several decades, a different approach was developed in parallel: \emph{neuromorphic computing}.

The concept of neuromorphic systems goes back largely to the work of Carver Mead \cite{mead1989analogvlsi}, and in particular to his 1990 publication ``Neuromorphic Electronic Systems'' \cite{mead1990neuromorphic}. Early neuromorphic systems were modeled directly on biological information processing. Electronic models of the retina, the cochlea, and neural circuits already revealed the potential of this approach.

As semiconductor technology grew more powerful, large digital and mixed-signal systems followed, within which spiking neural networks could be implemented at a much larger scale. Their development did not proceed in clearly distinguishable generations. Nevertheless, a clear transition can be recognized: from the emulation of individual biological functions, through large-scale neuromorphic architectures, to today's systems, in which event-driven operation, local states, and learning mechanisms are increasingly handled jointly.

This development was substantially influenced by the discovery and precise description of activity-dependent synaptic plasticity. Time-dependent learning rules such as spike-timing-dependent plasticity (STDP) in particular showed that the temporal structure of neural events can itself contribute directly to the modification of synaptic connections.

Neuromorphic processors from Intel, IBM, and other research groups have since demonstrated that event-driven neural processing can be implemented in large-scale integrated circuits. So far, their particular strengths have emerged chiefly where information arrives distributed in time and space and must be processed on the fly with the lowest possible energy consumption, for instance, in sensing, robotics, and edge applications.

Neuromorphic systems continue to spread to ever more complex applications; they are now not only beginning to reach domains so far dominated by transformer-based ANNs but also offer interesting ways of contributing to central undertakings of modern AI that go beyond them (see 4.2.3). The first spiking transformers \cite{zhou2023spikformer} and spiking or spike-inspired language models \cite{zhu2023spikegpt} show that essential elements of these architectures can in principle be combined with event-driven, highly sparse processing. Although genuinely spiking systems are not yet a fully equivalent replacement for today's large language models, the results so far show that the boundary between classical and neuromorphic AI is becoming increasingly permeable.

A fundamental difference from classical ANNs is linked to the role of time.
In biological neural networks, both the number of spikes within a time interval and their exact timing, latencies, intervals, and collective activity patterns can carry information. In the neuronal models of classical ANNs, by contrast, time is generally not an information-carrying variable in its own right. There, the temporal sequence of computational operations serves primarily to carry out a mathematical computation.

For spiking neural systems, on the other hand, the following holds:

\begin{center}

\emph{The dynamics of events can themselves become carriers of information.}

\end{center}

This adds a further dimension to the network's information processing. What can become crucial for processing and learning is not just the strength of an activation, but also \emph{when} a neuron fires, and in what temporal relation to other neural events.

In addition, the temporal dynamics renders active pathways, and the neurons involved, directly observable. Local learning mechanisms can act precisely where the relevant events occur. With STDP, for example, the temporal relation between pre- and postsynaptic spikes can directly trigger a change at that synapse.

This, then, offers a fundamentally different way of learning:

\begin{center}

\emph{The primary ordering of the learning process can begin locally at the synapse.}

\end{center}

A network can thereby modify some of its connections during the ongoing stream of events, without every individual adjustment having to be computed directly from a global error signal. Local plasticity does not, however, automatically replace every global learning signal. Rather, modern learning methods for spiking networks combine local plasticity rules, modulatory signals, credit-assignment mechanisms, and partly also gradient-based methods in various ways.

This, though, is where an essential difference from classical backpropagation lies: the question of which element of the network should be changed, and when, can at least partly be answered from the locally available events and states themselves.

\textbf{Spiking neural networks (SNNs):}

Spiking neural networks adopt the temporal, event-driven mode of operation of biological nervous systems. Their artificial neurons transmit information not exclusively through continuous activation values but through discrete spike events and their temporal structure.

SNNs are therefore not simply another variant of classical ANNs. They open up an additional design space in which event-driven operation, temporal coding, local states, and local plasticity can be direct constituents of information processing.

An important goal here is to perform computations only where and when relevant events actually occur. Combined with suitable neuromorphic hardware, which makes co-location of memory and computation at the synapses, massive parallelism, and event-driven temporal sparsity the starting point of a design freed from the bottleneck of the von Neumann architecture, this opens up the potential for substantially more energy-efficient processing.

Whether, and to what extent, SNNs can actually reduce the high energy cost of today's large AI systems on a broad scale, or even take over some of their present tasks, remains an open question. Their significance at present is precisely that they open up a fundamentally different path.

The preconditions for this approach to neural information processing and learning are in place, and current developments show that it is not confined to simple, specialized applications.

\subsection{3.2 A look at the human nervous system}

The human nervous system is by no means the only information-processing and adaptive system in the human body. In the finely branched network of the bloodstream, for example, B and T cells circulate as the principal carriers of immunological memory. Non-neuronal cell systems, too, show adaptive changes and forms of cellular memory. Neurotransmitters and their receptor systems, moreover, act far beyond the central nervous system and influence the exchange of information throughout the organism. Despite the interactions that undoubtedly exist between these systems, sufficiently general models that would allow their combined operation to be carried over directly into artificial learning systems are still lacking.

Even the nervous system itself is a functionally highly differentiated structure. Sensory and motor neurons, interneurons, neuroendocrine cells, and numerous other specialized cell types are only some of its components. Within the neural communication network itself, different mechanisms act together in the generation, propagation, and processing of neural activity, at the level of individual cells and at the level of cell populations.

In what follows, this diversity forces us to restrict ourselves to a few basic mechanisms relevant to our topic, and to accept the simplifications that this necessarily entails.

Let us first consider the structural and functional organization of the human nervous system. For our purposes, the processes under consideration can be roughly assigned to three levels:

\textbf{The micro level} is the level of individual neurons, their synapses, and the molecular and bioelectrical mechanisms of neuronal activity. If the membrane potential of a neuron exceeds a threshold under suitable conditions, an action potential can be triggered. Successive action potentials form spike trains, whose temporal structure and frequency are essential carriers of neural information. Through the interplay of many neurons, they give rise to complex activity patterns.

\textbf{The meso level} forms the bridge between the individual neuron and the large-scale functions of the brain. It ranges from local neural circuits through larger functional assemblies to interconnected structures and areas. Thalamic nuclei, the hippocampus, the amygdala, and cortical networks are examples of components of such highly parallel systems with many feedback loops.

\textbf{The macro level}, finally, comprises the interplay of large-scale neural systems that together support functions such as sensory information processing, homeostasis, motor control, memory, decision making, and thought. The brain regions involved interact through numerous feedforward and feedback connections. Their activity is coordinated by various rhythmic processes.

The neural network forms the physical basis of information processing at all of these levels. Spatially and temporally structured activity patterns arise in it continually. Some fade again after a short time; others leave longer-lasting changes in the networks in question through processes of plasticity. The hippocampus plays a central role particularly in the formation and consolidation of new declarative and episodic memories, and supports their gradual integration into widely distributed cortical representations.

In classical ANNs, a trained network state is fixed by storing the trainable parameters, in particular the connection weights. The brain, by contrast, implements a multitude of mechanisms of neuronal plasticity through which the properties of neural networks can change on very different spatial and temporal scales. They range from short-term changes in synaptic efficacy, through longer-term changes in individual synapses, to structural remodeling of neural networks. The synaptic and neuronal level is at present one of the most important interfaces with artificial learning models. In the future, however, organizational principles of higher levels, and in particular the joint processing of distributed, meaningful activity patterns, could increasingly become a focus of AI research.

\textbf{Activity patterns:}

Phenomenologically, much of brain activity can be understood as the dynamics of spatially and temporally organized activity patterns. Patterns that recur, or that remain stable over longer periods, can be read as the expression of learned network structures. They should not, however, be equated with spatially well-defined clusters; many neural representations are distributed, overlapping, and dependent on the situation.

Nevertheless, the notion of clusters is a helpful abstraction. Activity patterns can form functional ensembles whose members, through learning, tend to be activated together. A very large number of such partially overlapping ensembles can take part in representing perceptions, experiences, concepts, and abstract relations. New experiences activate existing representations, modify their links, and can produce new combinations. From this perspective, acquiring knowledge is not a matter of storing new, isolated clusters but of continually reorganizing an existing and highly interconnected system.

In artificial neural networks, some aspects of such processes can be described as the formation of geometric structures in high-dimensional representation spaces. The attention mechanism of transformers is a particularly clear example: it weights the relations between representations according to context and forms new, more abstract representations from them. This points to an interesting prospect: \emph{learning and prediction could increasingly build on representations that are already abstracted and partly stripped of details irrelevant to the task at hand.} In this respect there is at least a functional analogy to essential properties of human perception, response, and learning.

Powerful models of how neural activity patterns are encoded and decoded, and of the structural effects of their hierarchical and dynamic interaction, are therefore important topics of basic research. Transferring them to artificial systems could in the long run take us a step beyond the mere emulation of individual neurons, toward exploiting the higher-order organizational principles of biological information processing.

\subsubsection{3.2.1 The biological neuron}

The neuron is one of the most complicated microscopic building blocks of life on Earth. It is a highly specialized cell whose function rests on a complex interplay of electrical, chemical, and metabolic processes. Through selectively permeable membranes, through voltage- and ligand-gated ion channels, in particular for sodium, potassium, and calcium ions, and through a multitude of molecular signaling pathways, it takes in information, processes it, and passes it on to other cells.

The underlying mechanisms are far too complex to be covered here, so our discussion is confined to a greatly simplified functional model. This model leaves out many specialized cell types, such as pyramidal cells with orientation-selective dendritic subregions, direction-selective amacrine cells of the retina, or heavily branched cells such as Purkinje cells, often with several tens of thousands of convergent inputs each; these become relevant mainly when the aim is to model complex, highly convergent signal processing within individual neurons. Yet it is precisely these simplifications that make it easiest to connect the model to possible neuromorphic implementations.

The human brain contains about 86 billion neurons \cite{herculano2009human}, which form a highly interconnected system through an extraordinarily large number of synaptic connections. Their distribution is extremely uneven: the comparatively small cerebellum, for instance, contains more than half of all the neurons in the human brain.
This alone shows that the performance of a nervous system cannot be inferred directly from the number or packing density of its neurons. What matters are the cell types, the wiring, the network architecture, and the dynamics of their interaction. Spatial proximity can favor communication within local networks, but it is only one of many organizational principles of neural systems.

Even a seemingly ``resting'' neuron is by no means biologically inactive. Maintaining the membrane potential, the ion gradients, metabolism, and numerous molecular processes consumes energy continuously. Electrically, a neuron can remain in a subthreshold state without firing action potentials.

Biological neurons thus combine two closely coupled functional domains: electrochemical information processing and the maintenance of their metabolic and cellular viability. Signal processing takes place via dendrites, synapses, cell body, and axon, while long-term viability depends on a continuous supply of energy and nutrients provided by the neuron itself, by glial cells, and by the vascular system.

Beyond immediate signal processing, neurons and their connections are capable of adaptive change. These neuronal and synaptic plasticity mechanisms are an essential basis for learning and memory.

\begin{figure}[htbp]
\centering
\includegraphics[width=0.90\linewidth]{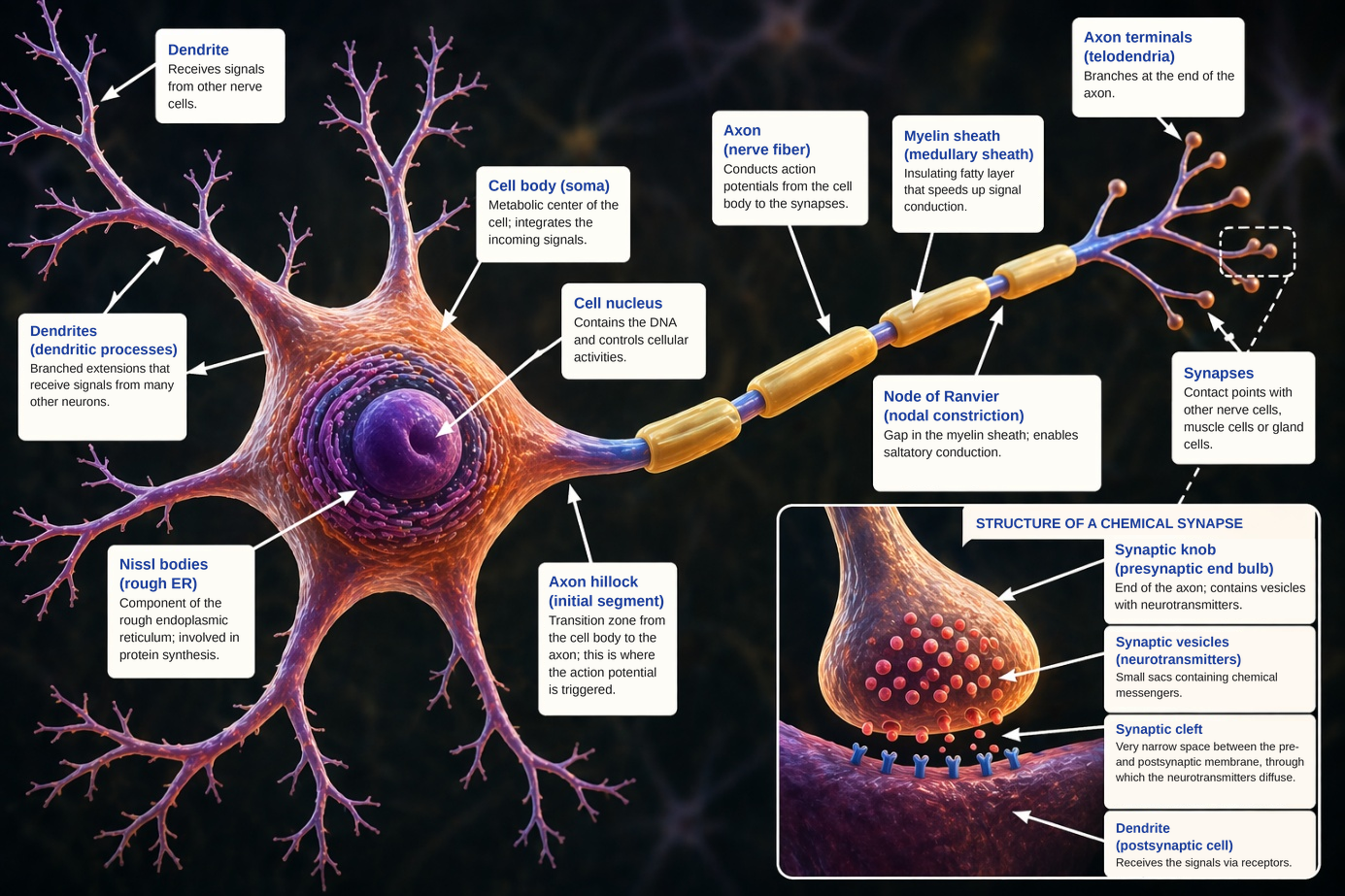}
\caption*{Fig. 3.1 The neuron}
\end{figure}

\textbf{Function and components of a neuron:}

As a first approximation, a neuron can be described as a membrane-enclosed capsule whose interior has a different ionic composition from its surroundings: above all, many positive K\textsuperscript{+} ions inside and many positive Na\textsuperscript{+} ions outside. A sodium-potassium pump (per pump cycle, 3 Na\textsuperscript{+} ions out and 2 K\textsuperscript{+} ions in) maintains these concentration gradients at an energy cost. Because the resting membrane is permeable mainly to K\textsuperscript{+} ions, a charge separation develops across it, and with it a negative membrane potential. At rest, the concentration gradient and the electrostatic forces it produces balance each other, resulting in a membrane potential of about \(- 70\, mV\).

The controlled opening and closing of selective ion channels, controlled by voltage or by neurotransmitters, changes this potential for short periods and thus forms the basis of neuronal signal processing. Synapses are the junctions at which this control is passed from cell to cell.

Fig. 3.1 shows schematically the components of a neuron and their functions that matter most for our discussion.

\begin{enumerate}

\item

The \textbf{cell body (soma)} contains the nucleus and most of the organelles the cell needs for its metabolism. Next to it lies the axon hillock, which leads into the axon initial segment. Excitatory and inhibitory inputs arriving at the dendrites and soma are integrated across space and time. The axon initial segment is of particular importance for triggering an action potential, because it has a high density of voltage-gated sodium channels.

As mentioned, a typical neuron has a resting potential on the order of \(- 70\, mV\). It results from the interaction of the different ion gradients, selective membrane conductances, and active transport processes, in particular the sodium-potassium pump.

\begin{figure}[htbp]
\centering
\includegraphics[width=0.90\linewidth]{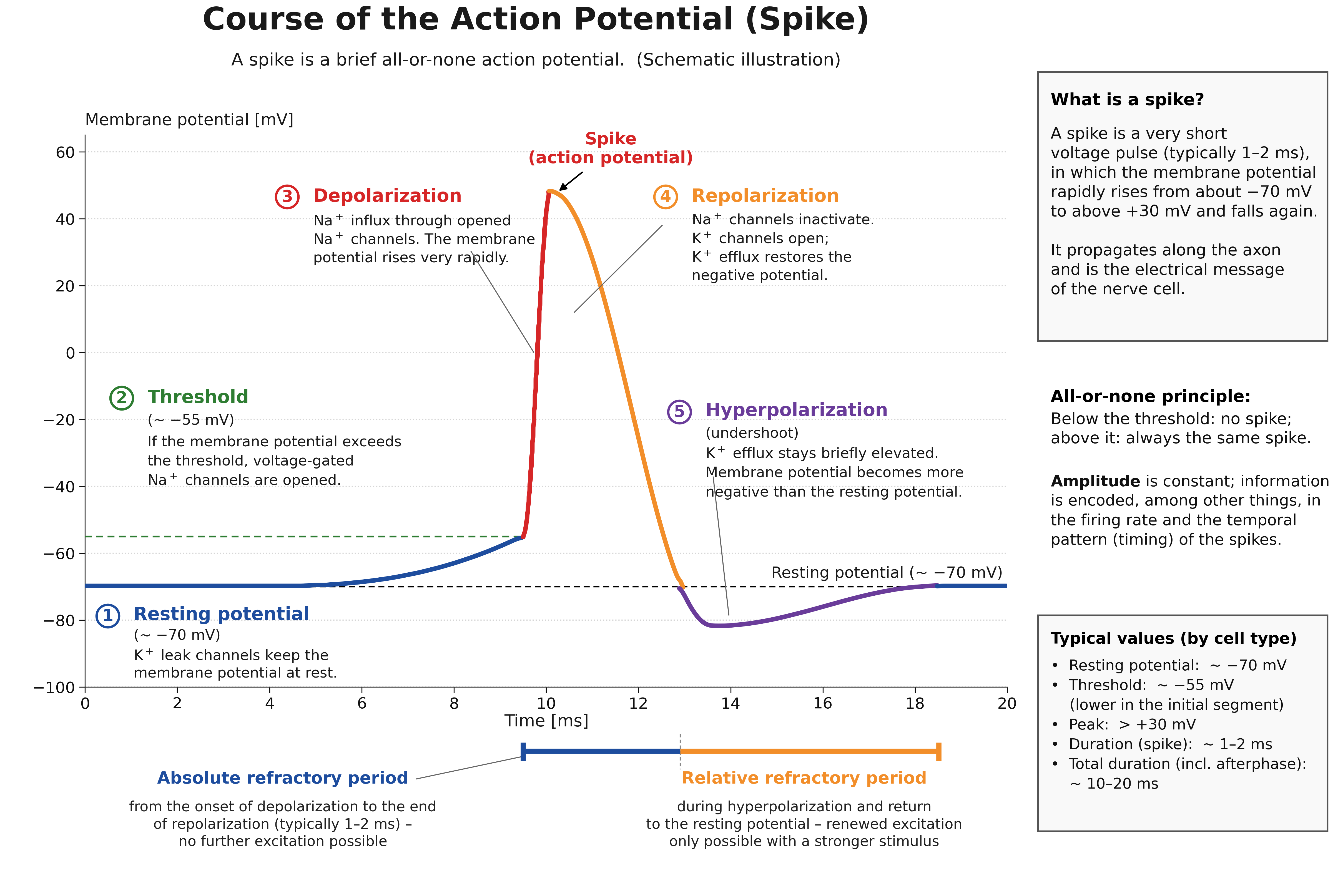}
\caption*{Fig. 3.2 Time course of an action potential with refractory periods}
\end{figure}

If incoming signals depolarize the cell enough to reach threshold at the axon initial segment, a steeply rising action potential is triggered in an avalanche-like process: the initial influx of Na\textsuperscript{+} opens more and more Na\textsuperscript{+} channels, so that even more Na\textsuperscript{+} rushes in. The subsequent efflux of K\textsuperscript{+} brings the potential back down and ends the spike (repolarization).

The ion movements involved, however, amount to only fractions of a thousandth of the neuron's internal ion content, and the sodium-potassium pump easily restores them afterwards. Each spike, so to speak, takes only a small sip from the neuron's large lake of ions.

Typical thresholds, depending on the neuron and its current state, often fall in the range of about \(- 50\) to \(- 55\, mV\).

The action potential is a brief, regenerative electrical impulse that travels along the axon. Once it has been initiated, it is followed first by an absolute refractory period, during which no further action potential can be generated, and then a relative refractory period, during which a stronger stimulus is needed to trigger a new one. The afterhyperpolarization that is often observed arises, among other things, from potassium conductances that remain active for longer.

Fig. 3.2 shows a typical time course.

The cell body also houses the nucleus and organelles such as the mitochondria, which supply energy for many processes, including the maintenance of ion gradients, protein synthesis, and intracellular transport. Neuronal activity can also alter gene expression through intracellular signaling pathways and in this way support long-term structural and functional adaptation.

\begin{figure}[htbp]
\centering
\includegraphics[width=0.9\linewidth]{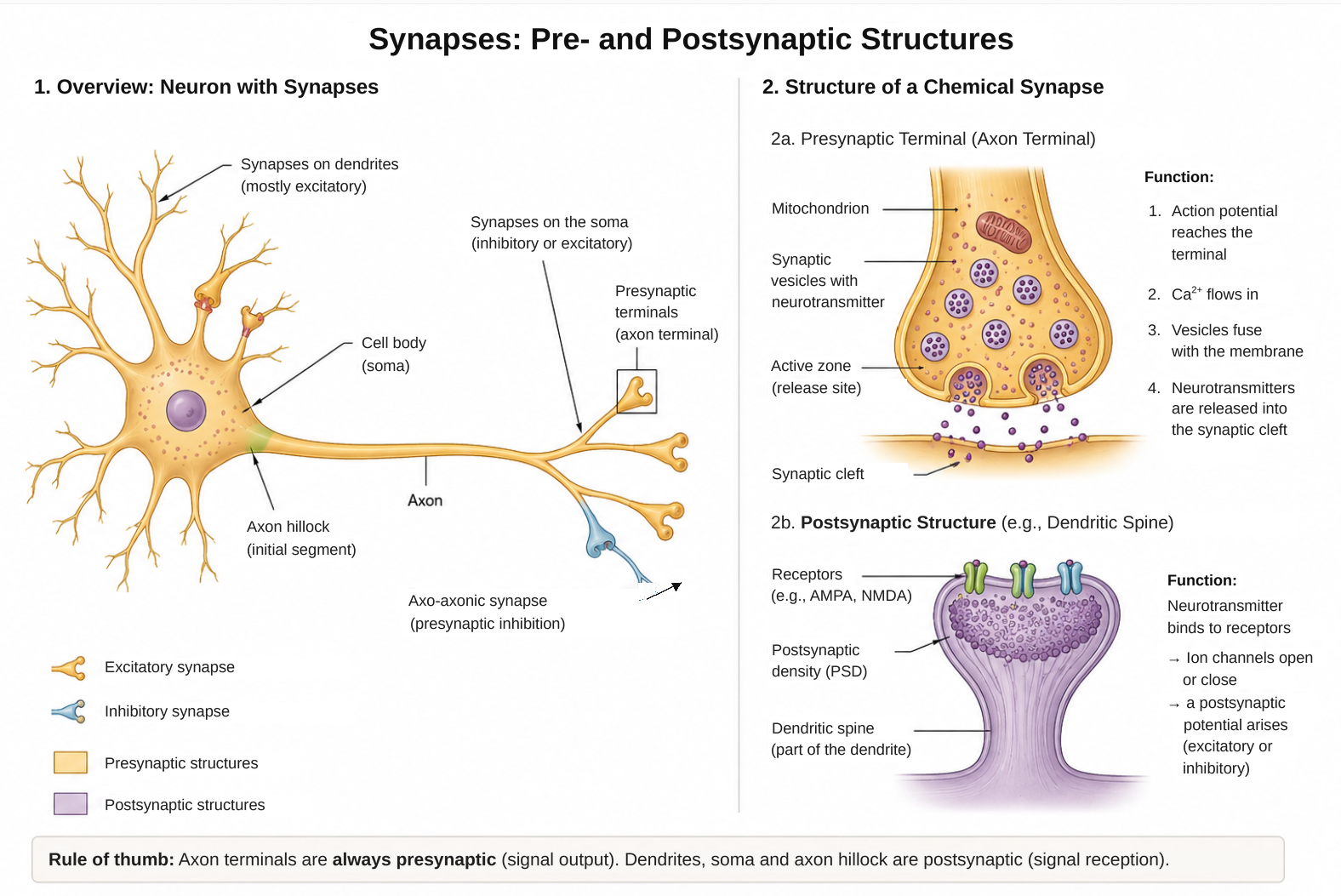}
\caption*{Fig. 3.3 Structure and function of chemical synapses}
\end{figure}

\item

The \textbf{axon} conducts action potentials to downstream target cells and, in humans, can be up to a meter long. Its conduction velocity depends, among other things, on the axon diameter and on myelination. Myelinated axons reach particularly high conduction velocities through saltatory conduction.

The properties of axons and their myelination can also change in an activity-dependent
way. Such processes are increasingly studied as part of neuronal plasticity, but they
operate on different time scales and through different mechanisms than fast synaptic weight
adjustment.
\item

A \textbf{dendrite} is a usually highly branched process of a nerve cell and forms a major receiving surface for synaptic input. Dendrites do not simply relay these signals passively to the soma; their own voltage-dependent properties allow them to carry out complex local processing.

Dendritic structures, and dendritic spines in particular, can be remodeled, newly formed, or eliminated during learning and development. They are therefore an important component of structural neuronal plasticity.

\item

The \textbf{synapses} form the functional connections between neurons. A chemical synapse consists of the presynaptic terminal, the synaptic cleft, which is only a few tens of nanometers wide, and the postsynaptic membrane. Synapses are often located on dendrites but can also attach to the soma and to other regions of a neuron.

Fig. 3.3 shows schematically how they are arranged and how a presynaptic terminal interacts with the postsynaptic membrane of a downstream neuron.

\begin{figure}[htbp]
\centering
\includegraphics[width=0.80\linewidth]{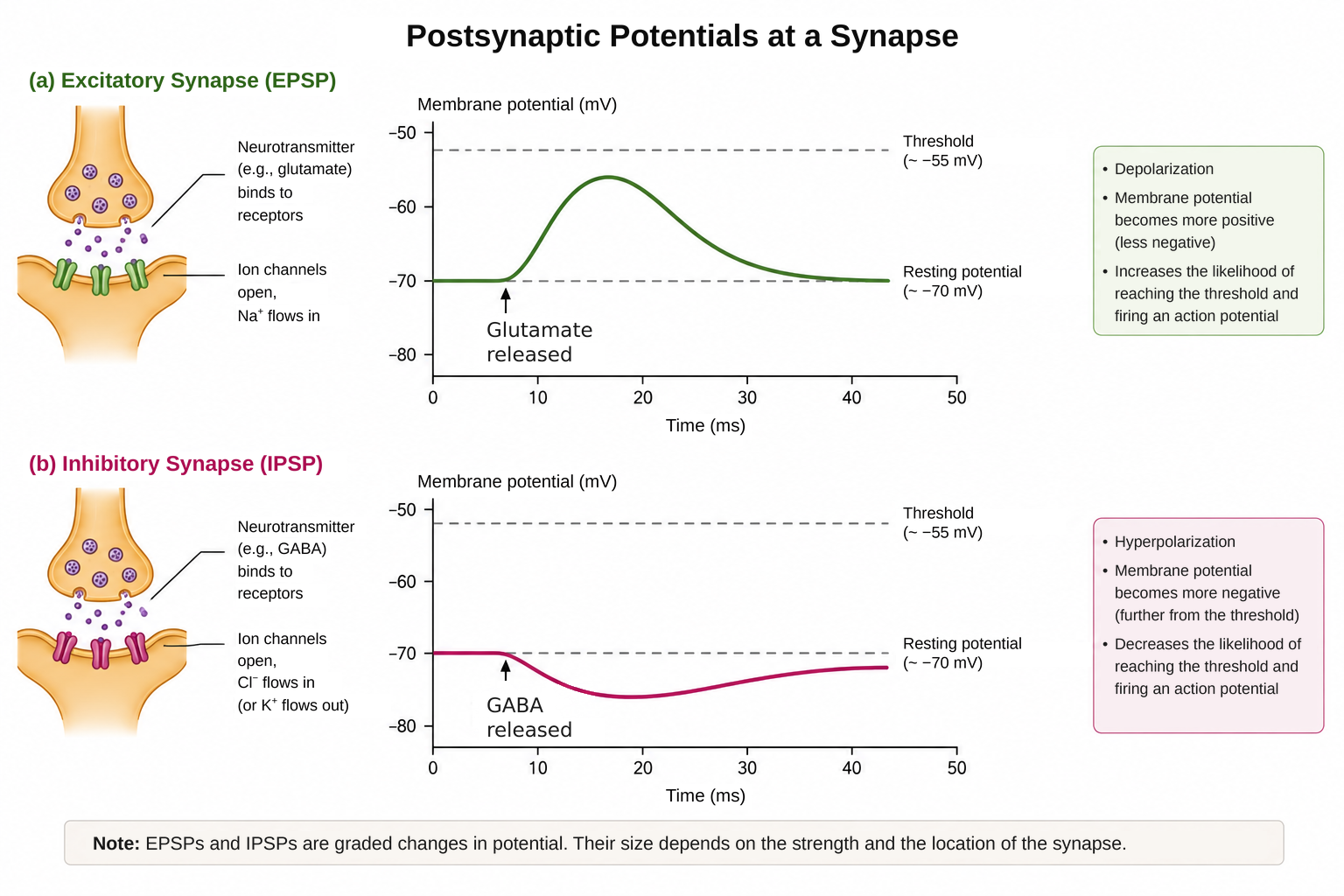}
\caption*{Fig. 3.4 Interaction of excitatory and inhibitory potentials}
\end{figure}

The presynaptic side is the transmitting side, the end of the arriving axon. When an action potential reaches the synapse, voltage-gated Ca\textsuperscript{2+} channels open there; the calcium influx causes small transmitter-filled vesicles to fuse with the membrane and release their contents into the narrow synaptic cleft. The postsynaptic side is the receiving side, the membrane of the target cell on the opposite side of the cleft. It carries receptors that match this particular transmitter, like a lock and key. The two sides are coupled as follows: the transmitter diffuses across the cleft and binds to the receptors, and this binding opens ion channels on the postsynaptic side. The channels let ions pass through the membrane; for an excitatory signal, mainly Na\textsuperscript{+} ions flow inward, which raises the membrane potential locally. These postsynaptic potentials are illustrated in Fig. 3.4.

A single synapse is usually not enough to make the cell fire. Many small potentials add up, both spatially (many synapses at once) and temporally (in rapid succession), and converge at the axon hillock, as shown in Fig. 3.5. If their sum exceeds the threshold there, the voltage-gated Na\textsuperscript{+} channels open all at once and a new action potential arises: the cell ``fires'' and in turn becomes the presynaptic side of the next synapse.

Excitatory inputs push the potential toward the threshold, inhibitory inputs push it away; the balance between them decides whether the cell fires.

\begin{figure}[htbp]
\centering
\includegraphics[width=0.80\linewidth]{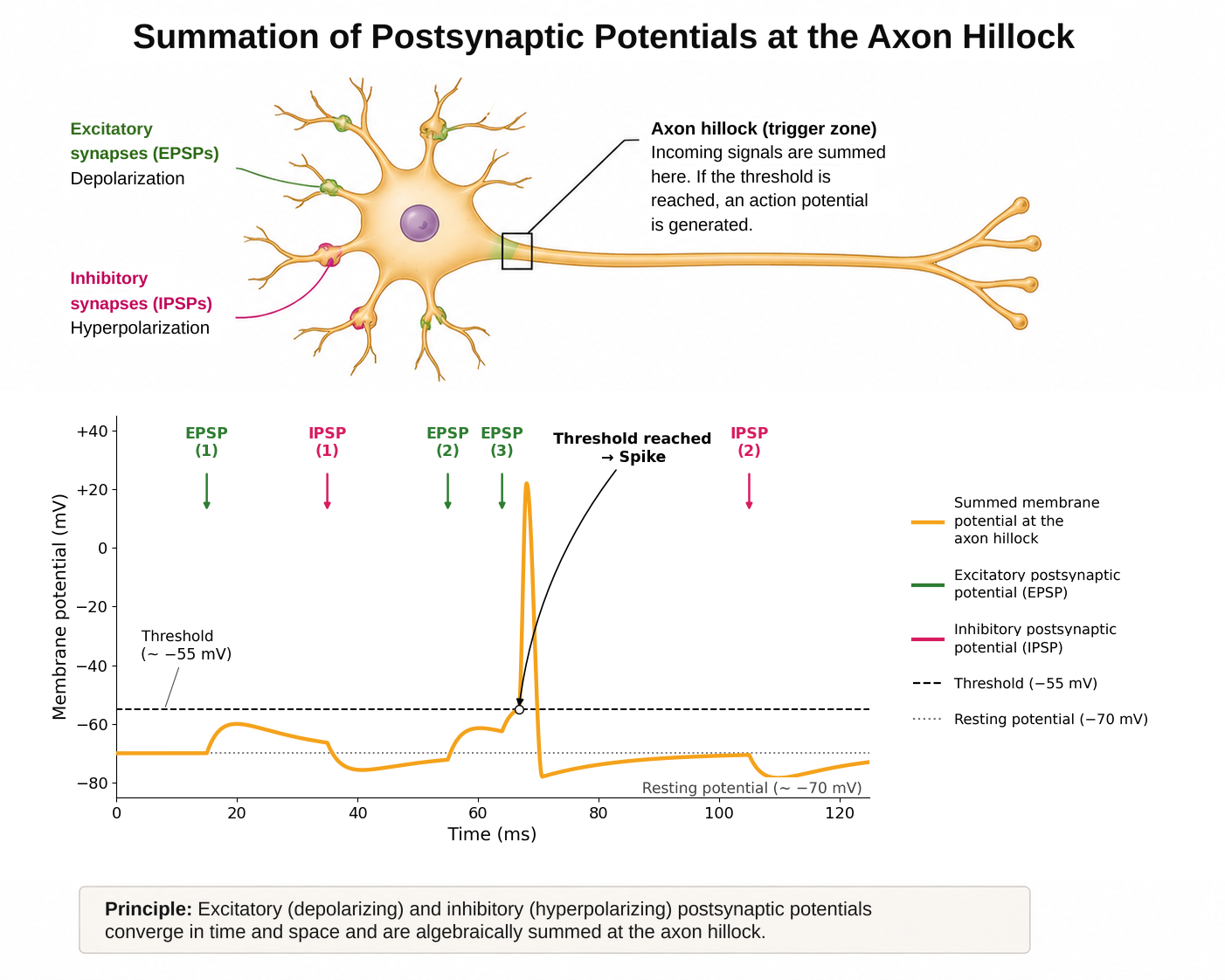}
\caption*{Fig. 3.5 Spatial and temporal integration of synaptic inputs}
\end{figure}

In summary: in response to an action potential, the presynaptic terminal releases transmitter; the transmitter opens channels in the postsynaptic membrane and shifts its potential. Once enough of these events coincide, the axon hillock triggers the next firing.

\end{enumerate}

A few remarks on the \textbf{excitatory and inhibitory synapses} mentioned above will complete the picture. The human brain has an estimated \(10^{14}\) to \(10^{15}\) synapses, and at most of them signal transmission is chemical. Based on their predominant effect, these chemical synapses can be divided, in simplified terms, into excitatory and inhibitory connections. In the cortex, excitatory synapses are the clear majority. The numerically smaller group of inhibitory synapses can nevertheless have an extraordinarily strong influence on network dynamics.

What matters is not only their number but also their location and wiring. Inhibitory synapses near the soma or the axon initial segment can control the initiation of an action potential particularly effectively and in this way, figuratively speaking, exercise a kind of ``veto''. Figs. 3.4 and 3.5 illustrate schematically how excitatory and inhibitory inputs interact. The important point is that the combined effect of EPSPs and IPSPs determines whether an action potential is initiated.

It should be noted that the human nervous system contains many different kinds of nerve cells which, apart from their specializations, all work largely according to the basic principles just described. Their particular characteristics may point to functions that neuromorphic designs could exploit.

\textbf{Electrical synapses:}

Besides chemical synapses there are electrical synapses, which couple neighboring cells directly through gap junctions. Each of the two cells contributes a hemichannel (connexon), and together these form a direct connection between the interiors of the cells.

Electrical currents and small molecules can therefore pass directly between coupled cells. The coupling is very fast, although not entirely free of delay, and depending on the type of synapse it can be bidirectional or functionally asymmetric (rectifying).

In a simplified electrical description, the coupling current between two cells can be approximated by
\[
I_{\mathrm{gap}} = g_{\mathrm{gap}}(V_1-V_2)
\]
Such coupling can reduce differences between membrane potentials and, under suitable conditions, promote synchronization of neuronal populations. Depending on network structure and cell dynamics, however, it can also produce more complex effects.

Electrical coupling between cells also plays a role outside the nervous system; gap junctions, for example, enable the coordinated electrical activity of cardiac muscle cells. In neural networks they are involved, among other things, in synchronizing certain interneuron populations and in rhythmic activity.

For artificial neural systems, this mechanism is conceptually interesting. Virtual electrical couplings could be used to link the dynamics of neighboring or functionally related neurons directly to one another. Whether, and under what conditions, such mechanisms can actually speed up learning, support cluster formation, or improve credit assignment remains an open research question.

This open question seems particularly relevant for neuromorphic systems: in addition to modifying individual synaptic weights, dynamically coupling whole groups of neurons could provide a further mechanism of self-organization.

\textbf{Glial cells:} The classical picture of the nervous system as an information-processing system in which only neurons take part is increasingly being broadened by findings on the active role of glial cells, which contribute to metabolic support, homeostasis, myelination, immune defense, and repair, among other things.

Astrocytes in particular interact closely with synapses and can modulate their function and plasticity. Glial cells are therefore not merely passive supporting tissue but part of the complex regulation of neural information processing.

There is as yet no generally accepted computational abstraction of their diverse functions. For future neuromorphic models, however, including higher-order regulatory mechanisms of this kind could be what enables a further step beyond the mere emulation of neurons and synapses.

\subsubsection{3.2.2 The dynamics of signal propagation}

To emulate biological neural networks in artificial systems, we need mathematical models of how neural activity arises and propagates. Many details are still not fully understood: how different neuron types interact, how synchronizing and modulating mechanisms work, and how stochastic influences come into play. Even so, the essential processes can be captured with comparatively simple models.

As a first approximation, it is sufficient to treat the activity of a neuron as a sequence of brief spike events.

\begin{figure}[htbp]
\centering
\includegraphics[width=0.85\linewidth]{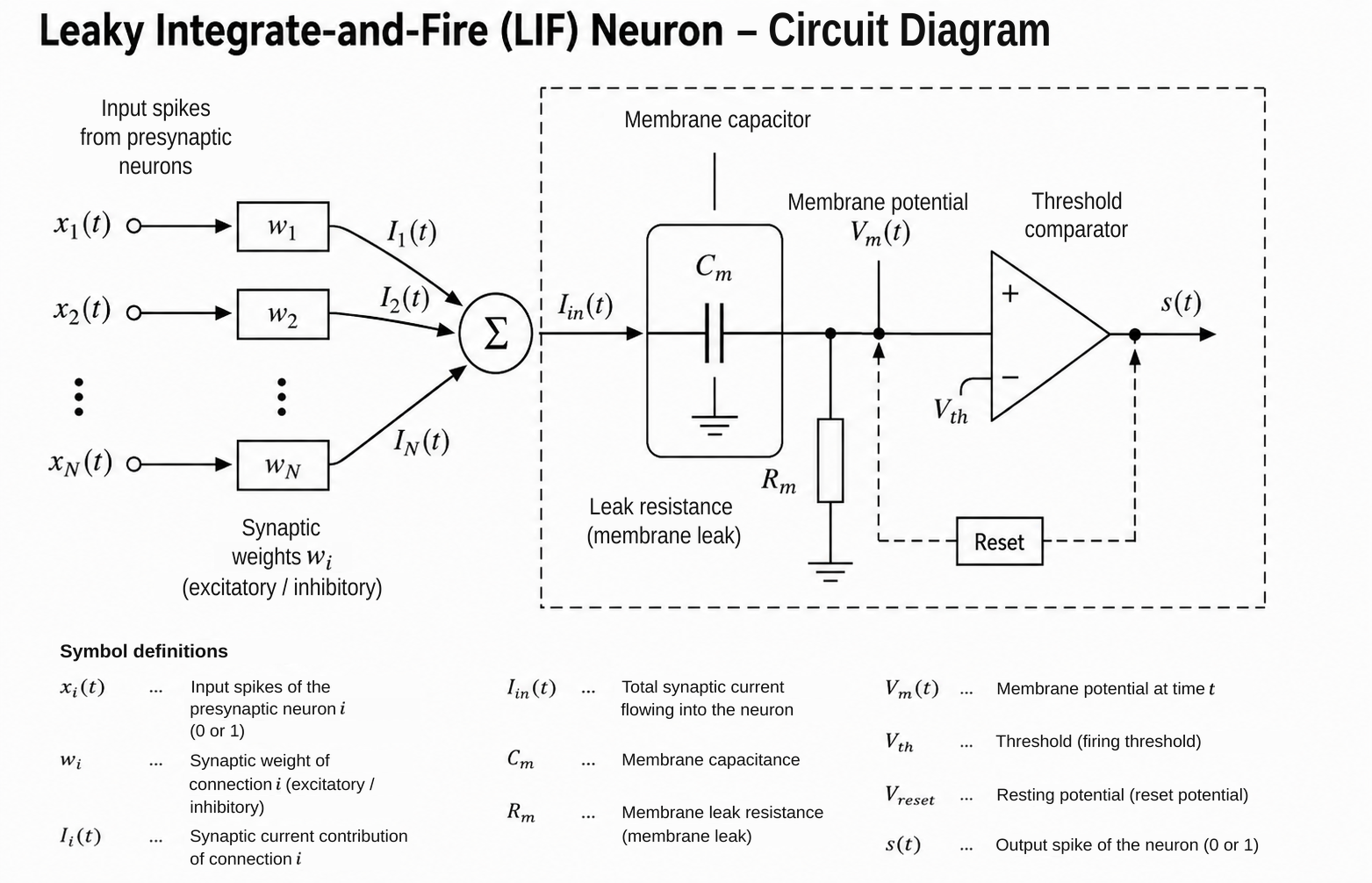}
\caption*{Fig. 3.6 The LIF model}
\end{figure}

Remarkably, one of the oldest models is still among the most widely used. It goes back to Louis Lapicque, who in 1907 formulated a simple equivalent circuit for the excitability of nerve tissue \cite{lapicque1907}, long before Hodgkin and Huxley gave a quantitative description of the biophysical mechanisms of the action potential \cite{hodgkin1952quantitative}. A more accurate model of the important information-processing operations, and of how they interact with structural changes in the neuron, must include at least the actual mechanisms by which potentials are generated and propagated. These mechanisms are complicated, and their detailed effects are often still poorly understood, but as the field develops they will naturally become central to accurate models.

In the leaky integrate-and-fire (LIF) model, following Lapicque, the cell membrane is treated in simplified form as a parallel electrical circuit consisting of a capacitance \(C_{m}\) and a leak resistance \(R_{m}\). Current conservation gives

\begin{equation}\tag{3.1}
C_m\frac{dV_m(t)}{dt}
=
-\frac{V_m(t)-V_{\mathrm{rest}}}{R_m}
+I_{\mathrm{in}}(t).
\end{equation}

The capacitor represents the membrane's ability to store electrical charge, while the leak resistance ensures that, in the absence of further input, the membrane potential returns toward the resting potential \(V_{rest}\).

Incoming synaptic signals change the membrane potential. When it reaches a threshold \(V_{th}\), a spike is triggered. The LIF model does not reproduce the actual waveform of the action potential; it simply registers its occurrence as an event. The membrane potential is then set to a reset value, and a refractory period can also be specified during which no further spike can be triggered.
In a greatly simplified current-based formulation, the input current can be written as

\begin{equation}\tag{3.2}
I_{\mathrm{in}}(t)
=
\sum_j w_j S_j(t)+I_{\mathrm{bias}},
\end{equation}
where \(S_{j}(t)\) describes the spike events arriving over the \(j\)-th connection and \(w_{j}\) is their synaptic efficacy. In a model this simple, excitatory and inhibitory contributions can be represented by positive and negative weights, respectively.

The bias term can represent a constant background drive. Stochastic influences, by contrast, are better described by separate noise terms.

The effect of successive input signals depends strongly on their timing. Because of the leak term, the influence of earlier events fades, so inputs that arrive in close succession summate more effectively than inputs that are far apart.

\begin{wrapfigure}[17]{r}{0.5\textwidth}
\centering
\includegraphics[width=\linewidth]{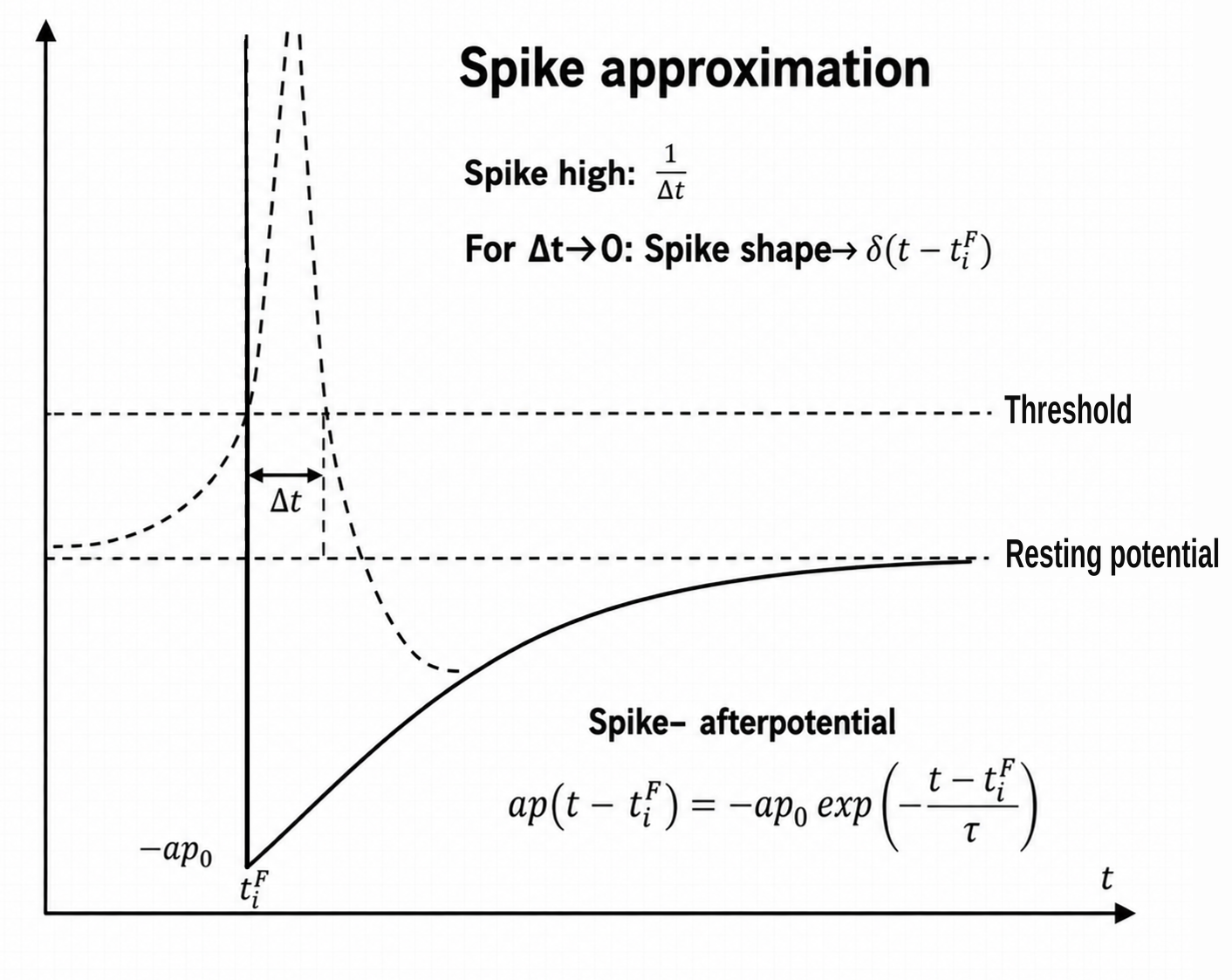}
\caption*{Fig. 3.7 Idealization of a spike event}
\end{wrapfigure}

For the mathematical description of spike trains, individual action potentials are often idealized as point-like events in time \cite{gerstner2002spiking}. Fig. 3.7 illustrates the step from the real time course of an action potential to this idealized representation, in which the solid black curve shows the spike as a (non-differentiable) delta function followed by the decay of the membrane potential.

If \(t_{j}^{k}\) denotes the time of the \(k\)-th spike of neuron \(j\), its spike train can be written as

\begin{equation}\tag{3.3}
S_j(t)=\sum_k\delta(t-t_j^k)
\end{equation}

\par
For impulse-shaped input signals, the LIF equation then becomes

\begin{equation}\tag{3.4}
C_m\frac{dV_m(t)}{dt}
=
-\frac{V_m(t)-V_{\mathrm{rest}}}{R_m}
+
\sum_j w_j\sum_k\delta(t-t_j^k)
+
I_{\mathrm{bias}}.
\end{equation}

Between reset and threshold events, the effect of a single input spike can be described by an exponentially decaying kernel. With the membrane time constant \(\tau_{m} = R_{m}C_{m}\), an idealized current impulse arriving at \(t_{j}^{k}\) contributes a term proportional to
\begin{equation}\tag{3.5}
\kappa(t-t_j^k)
=
\frac{1}{C_m}
e^{-(t-t_j^k)/\tau_m}
\Theta(t-t_j^k),
\end{equation}
where the Heaviside function \(\Theta\) ensures that a spike has an effect only after it arrives.

If we disregard the reset and threshold operations, the membrane potential can then be written formally as
\begin{equation}\tag{3.6}
V_m(t)
=
V_{\mathrm{rest}}
+
\sum_j w_j\sum_k
\kappa(t-t_j^k)
\end{equation}

A general initial state and a constant bias would each add a corresponding term.

The delta idealization therefore allows an elegant step from the temporal integration of the input signals to a sum of decaying contributions from past spike events.

At this point an important difference from classical ANNs becomes directly apparent: information takes the form of a pattern of events distributed in space and time. When activity is sufficiently sparse, only some of the neurons are active in any short time interval, so an event-driven engineered implementation only has to respond to the events that actually occur.

If one pictures the activity of such a network as a thunderstorm, one sees many individual lightning flashes at different places and different times, never merging into a steady glow across the whole sky.

Accordingly, an enormous structure of spatiotemporal activity patterns extends across the network. But this raises a fundamental question: what information is transmitted, by whom, and to whom? How is it encoded in the spike trains, and how is it decoded by downstream neural structures?

It would be presumptuous to assume that evolution produced only a single coding scheme for the many different tasks of neural communication. In fact, firing rates, spike times, latencies, relative intervals, synchrony, and population patterns can all carry information to varying degrees. Their exact functional significance remains one of the central questions of neuroscience.

Beyond the dynamics of the membrane potential, there are further temporal processes, in particular the generation of excitatory and inhibitory postsynaptic currents through transmitter release, receptor activation, and channel opening. Biologically, therefore, an arriving spike usually produces not an instantaneous delta current but a synaptic current that extends over time.

Such time courses can be approximated, for example, by exponential kernels or alpha functions. Their time constants are not in general smaller than the membrane time constant; they depend strongly on the type of synapse and receptor. Depending on the desired level of abstraction, SNN models either include these dynamics explicitly or absorb them into simplified synapse models.

The temporal dynamics of neural activity give rise to different coding schemes. Two are particularly intuitive: coding by the mean activity of a neuron or a population, that is, rate coding, and coding by spike times or latencies, for example the time-to-first-spike (TTFS) scheme. Both are considered in more detail in Section 4.1.2.

\textbf{Noise in neural networks:}

Biological and artificial neural systems alike are subject to stochastic influences from various sources. In biological neurons, for example, stochastic ion channel gating, fluctuations in synaptic transmitter release, and the activity of many other neurons can all contribute to fluctuations of the membrane potential. In electronic implementations, device and circuit noise adds to this.

In mathematical models, such influences can be described in different ways depending on their cause. Continuous fluctuations, for instance, can be modeled by an additive stochastic term, while randomly arriving spike events can be described by stochastic point processes.

A particularly simple model is the homogeneous Poisson process. If spike events are assumed to occur independently of one another at a constant mean rate \(\lambda\), the probability of exactly \(k\) events within a time interval \(\Delta t\) is
\begin{equation}\tag{3.7}
P\{N(\Delta t)=k\}
=
\frac{(\lambda\Delta t)^k}{k!}
e^{-\lambda\Delta t}.
\end{equation}
For this idealized process, the mean \(\langle \cdot \rangle\) and the variance of the event count are equal.

This does not mean, however, that biological spike trains are in general Poisson processes. The Poisson distribution serves instead as an important reference and noise model. Real neuronal spike trains can deviate from it considerably because of refractory periods, oscillations, time-dependent firing rates, and network interactions.

If we add a continuous stochastic current \(\xi(t)\) and randomly arriving spike events \(S_{j}^{r}(t) = \sum_{r}^{}\delta\left( t - t_{j}^{r} \right)\), a simplified stochastic LIF equation can be written, for example, as

\begin{equation}\tag{3.8}
C_m\frac{dV_m(t)}{dt}
=
-\frac{V_m(t)-V_{\mathrm{rest}}}{R_m}
+ I_{\mathrm{bias}}
+ \xi(t)
+ \sum_j w_j
  \left[
    \sum_k \delta(t-t_j^k)
    + \sum_r \delta(t-\tilde{t}_j^r)
  \right],
\end{equation}
where \(\xi(t)\) denotes a Gaussian noise process for the continuous fluctuations and \(S_{j}^{r}(t) = \sum_{r}^{}\delta\left( t - t_{j}^{r} \right)\) denotes stochastic point processes of the arriving spike events.

This representation shows two qualitatively different influences. The continuous noise component produces random fluctuations of the membrane potential, and the leak attenuates the effect of past fluctuations over time. Randomly arriving spikes, by contrast, produce discrete synaptic contributions whose effect also decays afterward.

The LIF neuron therefore has an elementary short-term memory: the past continues to act, but its influence decreases exponentially with increasing temporal distance.

Noise is not merely a nuisance here. Under certain conditions, a suitable degree of stochastic fluctuation can lift weak signals over a threshold, support exploratory behavior, or improve the robustness of a learning method. This effect is known, for example, from stochastic resonance \cite{douglass1993noise}. Excessive noise, on the other hand, destroys the temporal and spatial structure of the transmitted information and can become a problem, particularly for precise temporal coding schemes.

This raises a question that is crucial for learning: how can a neuron or a neural network pick out, from a multitude of stochastic and structured events, those temporal relations that recur and are relevant to its function?

One possible answer comes from time-dependent plasticity mechanisms such as spike-timing-dependent plasticity (STDP).

Under suitable conditions, STDP can bring out recurring temporal correlations against an unstructured background. In this way, a neuron can become increasingly sensitive to particular spatiotemporal input patterns.

This makes a remarkable transition possible: through local, time-dependent plasticity, what starts as a simple integrator of arriving spike events can become a selective detector of recurring causal structures.

How this mechanism works is the subject of the next section.

\subsubsection{3.2.3 Spike-timing-dependent plasticity (STDP)}

Biological networks have their own temporal dynamics, and because of this not every past event matters equally for what a neuron is doing now.

Earlier influences fade, while events that occur close together in time are linked especially effectively. The network therefore processes not only which neurons are active but also, within short time windows, the order in which they become active.

Experiments in the 1990s showed just how much this temporal order matters. Markram et al.~\cite{markram1997regulation} and, in particular, Bi and Poo \cite{Bi1998Synaptic} demonstrated that the change in strength of certain synapses depends on the temporal order of presynaptic and postsynaptic activity, and that a few milliseconds decide whether a connection grows or shrinks. This turned Hebb's postulate from a vague slogan into a measurable, temporally directed rule: learning depends not only on the fact \emph{that} two neurons are active at roughly the same time, but also on the \textbf{order} in which they fire.

\begin{wrapfigure}[16]{r}{0.45\textwidth}
\centering
\includegraphics[width=\linewidth]{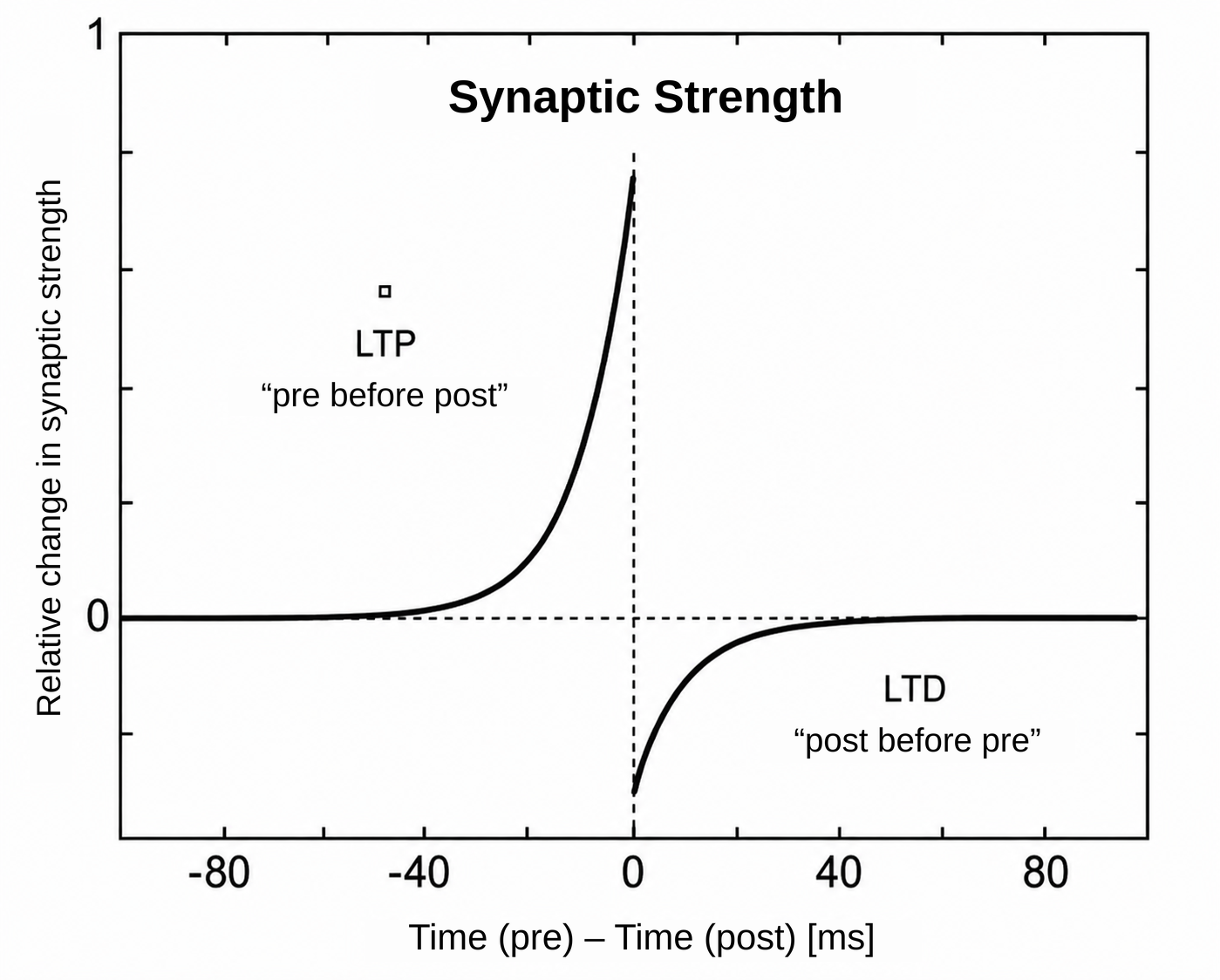}
\caption*{Fig. 3.8\quad Schematic representation of an STDP learning window}
\end{wrapfigure}

In this sense, STDP is a mechanism that detects \emph{causal temporal structure} and translates it into synaptic strength. The pre-before-post order counts as causal: if the presynaptic neuron fires \emph{before} the postsynaptic one, it could have contributed to the latter's firing, and it is exactly this configuration that is rewarded.

Classical pair-based STDP distinguishes two cases:

\textbf{Causal order (pre before post):} If the presynaptic neuron's activity arrives shortly before the postsynaptic spike, synaptic efficacy is strengthened (LTP, long-term potentiation).

\textbf{Reverse order (post before pre):} If the postsynaptic neuron fires before the presynaptic neuron, synaptic efficacy is weakened (LTD, long-term depression).

With \(\Delta t = t_{post} - t_{pre}\), a commonly used idealized STDP window can be written as
\begin{equation}\tag{3.9}
\Delta w(\Delta t) =
\begin{cases}
A_+ e^{-\Delta t/\tau_+},
& \Delta t>0 \qquad (\mathrm{LTP}),\\
-A_- e^{\Delta t/\tau_-},
& \Delta t<0 \qquad (\mathrm{LTD}).
\end{cases}
\end{equation}
Here \(A_{+}\) and \(A_{-}\) set the size of the maximum weight changes, while \(\tau_{+}\) and \(\tau_{-}\) set the temporal extent of the learning windows (see Fig. 3.8). This idealized rule captures the basic behavior; the exact values, and even the shape of the learning window, depend on the type of neuron and synapse and on the experimental conditions. The double exponential shown here is therefore not a universal biological law but a particularly important and widely used model of pair-based STDP \cite{gerstner1996neuronal,song2000competitive}.

The strongest changes typically occur at short intervals and decay as the interval grows. In a sense, then, the synapse has a local temporal memory: the longer ago a relevant event occurred, the less it influences the current synaptic change. In popular, if only approximate, form, this can be summarized as

\begin{center}

\textbf{``Fire together, wire together.} \textbf{Out of sync, weaken the link.''}

\end{center}

What matters, then, is not the ``together'' alone but the \textbf{temporal order} of the events.

But why does plasticity depend on order at all? For one important class of glutamatergic synapses, the underlying mechanism can be pictured in an intuitive way.

The presynaptic neuron fires first and releases glutamate, which binds to NMDA receptors on the postsynaptic membrane. This primes the synapse, and the bound glutamate leaves, as it were, a lingering trace. At a negative membrane potential, however, the ion channel of these receptors remains blocked by magnesium ions.

Shortly afterwards, the postsynaptic neuron fires. Its action potential travels not only along the axon but also back from the axon initial segment into the soma and dendrites; this is called a \emph{backpropagating action potential}. In effect, this backward propagation is the message ``I have fired,'' sent to the neuron's own inputs, and it depolarizes the synapse from the postsynaptic side.

The two signals now meet exactly at the synapse: the backpropagating depolarization relieves the magnesium block of the NMDA channel that is still occupied by glutamate. Only this coincidence opens the channel; calcium flows into the postsynaptic cell and sets in motion intracellular signaling processes that strengthen the connection over the longer term (LTP).

If the order is reversed, that is, if the postsynaptic neuron fires \emph{before} the presynaptic one, the calcium influx is smaller and differently timed. According to the calcium hypothesis, a large, steep rise in calcium leads to LTP, a moderate one to LTD. The same substance therefore decides, through its amount and its time course, whether the synapse is strengthened or weakened. This is the molecular basis of the temporal asymmetry.

The synapse therefore holds, locally, two pieces of information essential for learning: \emph{there was just now a presynaptic event here}, and \emph{shortly afterwards the postsynaptic neuron was strongly activated}. Their temporal coincidence is converted directly, at the site of the connection, into a change in synaptic efficacy.

This is perhaps the most remarkable property of the mechanism from the point of view of artificial learning systems. Consider the contrast with artificial neural networks: these learn by means of a global backpropagation algorithm that computes the error backward from the output to every single connection. STDP needs none of this:

\emph{Local changes in the connection structure arise during ongoing neural activity, solely from pre- and postsynaptic timing, without any globally computed error information having to be propagated back to this synapse.}

Recurring sequences therefore preferentially strengthen particular connections and leave traces of temporally structured processes in the network.

At this point we need to distinguish between the functional change of a synapse and the site of its molecular implementation. The weight \(w_{ij}\) used in SNN models describes, in the first instance, only the effective strength of the connection from presynaptic neuron \(i\) to postsynaptic neuron \(j\); its update is formally described by
\begin{equation}\tag{3.10}
w_{ij}^{\mathrm{new}}
=
w_{ij}^{\mathrm{old}}+\Delta w_{ij}.
\end{equation}
Biologically, such a change can be implemented by different presynaptic and postsynaptic mechanisms: postsynaptically, for example, through the number and properties of the receptors; presynaptically, among other things, through the probability of transmitter release. Which mechanisms dominate depends largely on the type of synapse and the form of plasticity involved.

Short-term plasticity, with mechanisms such as synaptic depression and facilitation, is a separate matter. These mechanisms alter the instantaneous transmission efficacy of a synapse depending on its immediately preceding activity. Models such as the Tsodyks-Markram model \cite{tsodyks1997neural} describe such dynamic processes, but they are not simply an extension of classical STDP by an additional presynaptic weight adjustment.

\textbf{STDP and the refractory period:}

The refractory period constrains the possible temporal sequences of spikes of a single neuron, but it must not be confused with the STDP learning window. The time interval \(\Delta t = t_{post} - t_{pre}\) refers to spikes \emph{of two different neurons}. A presynaptic spike can therefore occur just a few milliseconds before a postsynaptic spike without being prevented by the postsynaptic neuron's refractory period. What the refractory period limits is how quickly \emph{the same} neuron can fire again, and in longer spike trains it also indirectly shapes the statistics of the spike pairs available for STDP. In SNN models, neuronal refractory dynamics and the STDP learning rule are therefore best treated as two interacting but fundamentally different mechanisms.

The simple pair-based STDP model is itself only a first approximation, however. In real neurons, plasticity also depends on preceding spike sequences, firing rates, membrane potential, neuromodulators, and further state variables \cite{sjostrom2001rate}; extended STDP models attempt to include such influences. The crucial basic idea nevertheless remains remarkably simple: \textbf{the temporal structure of neural activity itself becomes a local learning signal.} Recurring temporal relations directly modify the connections over which they are transmitted.

\subsubsection{3.2.4 Self-organization of neural networks}

The concept of self-organization (see also \cite{prigogine1981dialog}) is sometimes mystified more than it needs to be. In its simplest sense, it refers to the emergence of ordered structures from local interactions, without a central authority prescribing every step. STDP provides a particularly clear example. If a presynaptic neuron repeatedly becomes active shortly before a postsynaptic spike, the connection between the two is strengthened. As a result, that connection is more likely to act again when similar input patterns arrive later. A small statistical preference can grow step by step into a more stable functional pathway. Figuratively speaking, what begins as a barely visible trail becomes a well-trodden road. Temporal sequences that occur frequently leave traces in the network, and these traces in turn influence which activity patterns will occur preferentially in the future. The result is a positive feedback loop:

\begin{center}

\emph{Activity changes the connection;\\ the changed connection influences future activity.}

\end{center}

This mechanism has no knowledge of any global goal. No neuron needs to ``know'' which complete pathway is to form in the network. Order can instead emerge step by step from many local changes. On its own, however, STDP does not guarantee a meaningful global organization of the network. The weakening half of the rule (LTD) does counteract unrestricted strengthening and establishes a first form of competition between connections, but is not enough on its own. Without further limiting and stabilizing mechanisms, a positive feedback loop could strengthen individual connections without bound. Biological networks therefore combine local plasticity with additional processes such as inhibition, homeostasis, weight normalization, structural plasticity, and neuromodulatory influences. Stable activity patterns and preferred functional pathways can arise only from the interplay of such mechanisms.

The role of STDP in shaping the human brain during infancy is particularly striking. A child is born with a neural network that is still largely immature and largely unorganized. The first contact between two nerve endings is made through axon guidance followed by synaptogenesis. The filopodia on the growth cones of growing axons probe their surroundings using surface molecules and in this way reach a cell in the target region. For sensory inputs this may be a particular cortical area, for example, but it may also be the target region of corticocortical connections. If the surface proteins find matching partners on the dendrites of a suitable downstream neuron, a functioning synapse begins to form immediately (the birth of a synapse). A vast labyrinth of wiring emerges.

Spontaneous, internally generated activity patterns such as retinal waves contribute to an initial refinement even before birth, but after birth it is the incoming stimuli from the environment that determine which connections are useful. As signals propagate across many thousands of synaptic connections, STDP strengthens the connection between each pair of neurons if the presynaptic neuron fires and its postsynaptic neighbor follows within the learning window, and weakens it if the order is reversed. Whether a connection is retained depends not only on how often signals arrive simultaneously but, above all, on their causal order within the STDP learning window; the postsynaptic spike against which this timing is measured only occurs at all if many inputs are summed. Ultimately, the STDP rule is what the brain uses to remove the many unsuitable pathways from a huge neural labyrinth (pruning).

Of course, the activity of a single neuron carries little meaning on its own. Only populations, that is, pairs, triplets, neuronal groups, and so on, support meaningful processing. Hundreds or thousands of neurons must send their signals simultaneously to a group of target neurons. In the cortex these are usually densely and reciprocally interconnected neurons. If neuron X fires, it activates neuron Y, which fires back and activates neuron X again. STDP binds the two neurons together. Such a reciprocal loop remains stable only if the limiting mechanisms mentioned earlier, above all inhibition and homeostasis, do their part; only their interplay with STDP prevents runaway excitation and turns the loop into a truly stable, self-sustaining network. Synchronizing oscillations, among other factors, play an important role in maintaining it. This is the route by which the foundations of short-term memory, representations, abstractions, and other faculties of the human mind arise.

Clearly, not every intelligent system needs an STDP-like mechanism of its own; artificial systems, such as those based on backpropagation, demonstrate this clearly. Yet none of them would exist without a preceding natural, self-organized intelligence, which in turn rests on STDP (or on a functionally identical, time-dependent mechanism). To put it somewhat provocatively, with respect to their origins there would be no natural intelligence without STDP, and therefore no artificial intelligence either. STDP makes it possible to detect causal relations and, by encoding time intervals in synaptic pathways, to detect coincidences. Temporally recurring relations are recognized locally and translated into changes in the connection structure. Temporal patterns thereby give rise to more permanent functional structures.

For artificial spiking neural networks, this is exactly the idea that matters. Learning does not have to rely exclusively on a central optimizer that corrects all parameters after a complete computation pass. Local states and local activity history can directly determine which connections are candidates for change. This also provides the bridge to later learning mechanisms such as eligibility traces: a synapse can temporarily record that it took part in a possibly relevant event and later link this local trace to a delayed evaluative signal. Combining local self-organization with more global learning and evaluative signals is one of the particularly interesting research directions in modern neuromorphic systems.

\subsubsection{3.2.5 Neural learning}

Biological learning is not a single uniform process. It is the combined result of adaptive processes that operate on different time scales and at different structural levels. For our purposes, a rough division of these processes into two levels is useful:

\textbf{Short-term synaptic adaptation:}

Short-term plasticity refers to fast, reversible changes in synaptic strength. These changes typically last from milliseconds to seconds, sometimes longer, and require no permanent structural change to the synapse. Two important forms are short-term facilitation and short-term depression. In \textbf{facilitation}, presynaptic action potentials that arrive in close succession can briefly make the synaptic connection more effective. In \textbf{depression}, by contrast, repeated strong activity can temporarily reduce the strength of transmission (vesicle depletion).

The synapse therefore carries an elementary memory of its own recent activity. Its effect at any given moment depends not only on its long-term weight but also on what happened in the immediately preceding milliseconds or seconds. This is an interesting difference from many classical ANNs: the connection is not simply a fixed parameter that stays constant during a forward pass; it also has a dynamic internal state. Short-term plasticity can therefore act as a filter on temporal patterns, favoring some sequences of activity and attenuating others.

\textbf{Longer-lasting and long-term synaptic plasticity:}

Repeated or particularly effective activity patterns can trigger longer-lasting changes in synaptic connections. At the level of neurotransmitter receptors, synaptic strength is changed by the Ca\textsuperscript{2+}-triggered insertion or removal of AMPA receptors. (Calcium is the central signaling molecule that converts activity into a lasting change at the synapse.) A well-known form is long-term potentiation (LTP), in which the efficacy of a synapse is increased for an extended period. Its counterpart, long-term depression (LTD), can produce a longer-lasting weakening.

In many forms of glutamatergic LTP, increased postsynaptic calcium influx plays a central role. What matters is not a simple, universal ``calcium threshold'' but the magnitude and time course of the calcium signal and the intracellular signaling cascades it activates.

In the early phases of LTP, proteins and receptors that are already present can be redistributed or functionally modified. For example, the number of effective AMPA receptors in the postsynaptic membrane can increase, which strengthens synaptic transmission.

These early changes can remain stable for minutes to hours without necessarily requiring extensive new protein synthesis.

\textbf{Eligibility traces and delayed evaluation:}

Learning systems face a particular problem when the significance of neural activity becomes clear only after a delay. A behavior may, for example, be rewarded or punished only seconds later. The system must still be able to identify which previously active synapses might have been responsible for the outcome.

Theoretical models often address this with the concept of an \emph{eligibility trace}. A recently active synapse temporarily leaves behind a local state that decays over time. If a more global modulatory signal arrives while this state persists, the synaptic change can be applied specifically to the connections that are still marked.

A common schematic summary is:

\begin{center}

\emph{Local activity creates the eligibility; a later modulatory signal decides on the change.}

\end{center}

This principle is particularly interesting for machine learning because it links local learning to a global evaluative signal that arrives later. Not every synapse needs to receive a global error signal at all times. It only needs to record temporarily that it took part in a process that might turn out to be relevant.

\textbf{Synaptic tagging and late long-term plasticity:}

Such short-lived eligibility traces should be distinguished from \emph{synaptic tagging} (see Section 4.1.4). In synaptic tagging, particularly active synapses can be marked molecularly and can later capture plasticity-related proteins that are synthesized in response to strong neural activity.

This idea helps to explain how synapse-specific changes can be stabilized over long periods even though the proteins are produced in the cell body or in larger compartments of the cell.

In the late form of LTP, longer-term changes can develop in synapse structure, dendritic spines, and synaptic contact sites. A change that was initially purely functional is progressively consolidated into a structural one.

The transition from early to late LTP is not, however, a simple command to ``freeze'' a network state once it has been reached. Biological memory traces remain dynamic and can be modified further by renewed activation, reconsolidation, or competing learning processes.

\textbf{In summary:}

Short-term synaptic states retain information about immediately preceding activity. Long-term plasticity changes the efficacy of individual connections. Neuromodulatory signals can link local activity to later evaluations. Protein synthesis and structural remodeling stabilize some of these changes over long periods.

The biological nervous system therefore has something that exists in classical ANNs only in a very restricted form: \textbf{several memory and learning time scales that act simultaneously within the network itself.}

On this model, an artificial network no longer needs to treat every adaptation exclusively as an immediate change to a permanent weight. It can also make use of short-lived synaptic states, delayed eligibility traces, modulatory evaluative signals, and structural changes that are stabilized only later.

Learning then no longer appears as a single optimization step but, as in biology, as a staged process of provisional adaptation, evaluation, selection, and consolidation.

\subsubsection{3.2.6 Systems consolidation}

In the human brain, the memory traces that learning produces are not stored permanently in a single step. They are stabilized and integrated into existing knowledge by processes that operate at different temporal and spatial scales. In simplified terms, two forms of consolidation can be distinguished:

\begin{itemize}

\item

\emph{Synaptic consolidation} occurs within minutes to hours at the neuronal connections involved. It provides the initial stabilization of local plastic changes such as those described above in connection with LTP and LTD.

\item

\emph{Systems consolidation} refers to longer-term reorganization over days, months, or even longer periods, during which memories are integrated into distributed neural representations and linked with existing knowledge.

\end{itemize}

When new declarative and episodic memories are formed, the \emph{hippocampus} initially plays a central role, which is why it is often described figuratively as the \textbf{``gateway to memory''}. It is best understood not as a store but as an organizer and a fast-working connecting link: it binds together the simultaneously activated components of an experience, which are distributed across different cortical areas, and strongly supports their further consolidation.

During longer-term storage, the neural representations involved are repeatedly reactivated and, in the process, gradually modified and linked in new ways. New experiences are thereby integrated more and more fully into existing representations and knowledge structures.

Throughout this process, the hippocampus acts, in a sense, as a fast-working organizer that works together with cortical networks, which change much more slowly. This division of labor makes it possible to capture new experiences rapidly and relate them to one another without requiring every new piece of information to fundamentally alter stable knowledge structures that are already in place. McClelland, McNaughton, and O'Reilly formalized this interaction between a fast-learning hippocampal system and a slowly integrating cortical system as the theory of complementary learning systems \cite{mcclelland1995complementary}.

Sleep plays an important role here. During various sleep stages, patterns of neural activity are reactivated and coordinated between the hippocampus and cortical networks. \emph{Replay} is particularly well studied: sequences of activity that occurred during an experience can be replayed in time-compressed form during subsequent rest and sleep \cite{wilson1994reactivation}.

In non-REM sleep, slow cortical oscillations, thalamocortical sleep spindles, and hippocampal sharp-wave ripples \cite{buzsaki2015ripples} work together in a temporally coordinated manner. This interplay is regarded as an important mechanism for stabilizing new memories and gradually integrating them.

REM sleep also contributes to memory processing, although its exact role is harder to attribute to any single type of memory. The various sleep stages probably work together in a complementary way, supporting different aspects of the stabilization, reorganization, and integration of memories.

Systems consolidation is therefore a process in which new memories are fitted into existing knowledge structures through linking, transformation, and partial abstraction.

This organizational principle is of particular interest for artificial intelligence. A fast-learning system must be able to take in new information without overwriting knowledge it has already acquired. A slower consolidation process, by contrast, can repeatedly process selected experiences and gradually integrate them into more stable internal representations.

This basic idea recurs in various approaches to continual learning and in architectures that combine fast and slow learning components. Methods based on replay, whether of stored examples, internal representations, or artificially generated training patterns, likewise aim to stabilize what has already been learned while new learning is under way, and so to counteract \emph{catastrophic forgetting}.

DualNet \cite{pham2021dualnet} is one example of an explicit combination of fast and slow learning processes. Sleep and replay in biological systems have also inspired a number of studies on artificial consolidation methods. So far, however, artificial systems reproduce biological systems consolidation only to a very limited extent.

For neuromorphic systems in particular, this suggests a further possibility. Local plasticity could first rapidly produce provisional structures, while higher-level consolidation processes repeat, evaluate, and gradually integrate successful activity patterns into more stable knowledge structures.

Biological learning thus appears as the interplay of two complementary principles: \emph{fast capture and organization of new experiences, and slow integration into existing knowledge.} In this way, nature solves a problem that is also of fundamental importance for artificial learning systems: learning something new without forgetting what has already been learned.

\textbf{Predicting rather than representing: predictive coding in brain and machine}

The consolidation described so far, the conversion of labile memory traces into a stable but continually updated store, can be interpreted at a more general level as the maintenance of an internal model: the nervous system holds a representation of what to expect and continually compares it with what actually arrives. Our brain does not depict the world like a camera. It bets. At every moment, the cerebral cortex predicts what will be seen, heard, and felt next, and then attends almost entirely to the difference between its prediction and what really arrives: in other words, to the surprise. Whatever confirms the expectation costs hardly any attention; whatever contradicts it becomes a signal. Specialists call this principle predictive coding, and it is astonishingly economical: instead of constantly processing everything anew, the brain essentially processes only its own errors.

This is the core of predictive coding. The idea is old and well supported: Barlow's principle of efficient coding \cite{barlow1961efficient} and the retinal interpretation by Srinivasan, Laughlin, and Dubs \cite{srinivasan1982predictive} describe the nervous system as one that transmits not the signal itself but only its unexpected part. Building on earlier work by Mumford \cite{mumford1992computational}, Rao and Ballard \cite{rao1999predictive} and Friston \cite{friston2005theory} developed the principle into a specific theory of the cerebral cortex. Its grounding in known cortical mechanisms, the layer-specific wiring of prediction and prediction error, has been worked out in the canonical microcircuits of Bastos et al.~\cite{bastos2012canonical} and in the review by Keller and Mrsic-Flogel \cite{keller2018predictive}.

The evolutionary advantage of the principle lies chiefly in its energetic consequence, which has an information-theoretic basis. According to Shannon, the capacity of a communication channel of given bandwidth grows with its signal-to-noise ratio; transmitting more information reliably therefore requires a higher signal-to-noise ratio, and in neural systems every transmitted bit has a directly measurable metabolic cost \cite{laughlin1998metabolic}. A system that transmits only the unpredictable part of a message, the prediction error, reduces the amount of information to be transmitted, and with it the energy demand, without losing information. The high metabolic cost of neural signal transmission \cite{attwell2001energy} exerts considerable selection pressure toward this economy. At a more general level, the free-energy principle \cite{friston2010free} describes the same relationship as a behavioral principle: an evolved organism continually minimizes its ``surprise'' and in doing so keeps itself within a limited set of expected states that are compatible with survival. Put simply, it maintains a set point and continually corrects deviations from it, both internally (interoception or allostasis \cite{barrett2015interoceptive}) and in its interaction with the environment.

It should be emphasized that the strong form of this thesis, the free-energy principle as a comprehensive principle of life, is theoretically elegant but empirically contested. The weak form (hierarchical prediction and error coding), by contrast, is considered well supported \cite{walsh2020evaluating}. Predictive coding is therefore not a peripheral idea but one of the defining frameworks of systems and cognitive neuroscience \cite{clark2013whatever}.

Remarkably, modern artificial intelligence has arrived at the same basic principle independently. The most capable systems build internal world models, that is, internal representations of their environment, for instance within LeCun's JEPA framework \cite{lecun2022path}. These models do not merely represent the current state; they predict how it will evolve and learn from the prediction error. Predicting rather than representing: this is the point at which natural and artificial memory meet. The conceptual link between the two worlds is predictive coding itself, which is now being explicitly combined with spiking networks \cite{ndri2025survey}. The connection is therefore not merely metaphorical but structural: both kinds of architecture hold a predictive model of what to expect and update it on the basis of its errors.

Predictive coding mechanisms in the human nervous system are diverse, specialized, and far from fully understood. But this convergence of natural and artificial memory has a natural, though not inevitable, implication for hardware. If brain and machine compute essentially the same thing at their core, it is reasonable to consider, for the artificial side too, the economical, brain-like style of computation that nature evolved for this principle: event-driven, spiking, or neuromorphic architectures. The first spiking world models show that this is more than a hope. On standard benchmarks they match comparable classical systems \cite{sun2025spikingwm}. The finding should not be overstated, though: these are benchmarks, and the spiking variant is not cleverer than the conventional one; it reaches the same level of performance. The gain lies in the computation behind it: potentially a fraction of the energy. And that is exactly what could tip the balance in a world whose demand for electricity to power AI is growing rapidly.

So far, the approach has proved most effective in sensory applications where energy and latency are critical \cite{chowdhury2025robotic}. The current state of this development is discussed in Section 4.2.3.

The aim here was only to emphasize that consolidation in the brain and learning in the machine rest on the same principle: maintaining a model of what is expected, and learning from its errors.

%% file: chapters/04_spiking_neural_networks.tex
\section{4. Spiking neural networks (SNNs)}

\subsection{4.1 SNNs: the third generation of neural networks}

Hodgkin and Huxley \cite{hodgkin1952quantitative} won the Nobel Prize for their mathematical description of how neuronal action potentials (spikes) arise. Almost half a century passed before computer science and AI began to treat spiking neural networks as computing architectures in their own right.

Wolfgang Maass, a computer scientist and mathematician at Graz University of Technology, played a key role in this shift. In his foundational 1997 paper \emph{Networks of Spiking Neurons: The Third Generation of Neural Network Models} \cite{maass1997networks}, he classified spiking neural networks as the third generation of neural network models. Maass proved mathematically that temporally coded SNNs have greater computational power than the classical neural network models he considered: certain functions can be computed with markedly fewer neurons. He attributed this gain mainly to the use of precise spike timing as an additional dimension of information processing.

The subsequent development of artificial SNNs was driven by a large body of research on coding, on new learning methods, and on scaling networks up, and above all by the development of suitable neuromorphic hardware. Leading companies and research institutions such as IBM and Intel made major contributions to its implementation in engineering and to practical applications. Roy, Jaiswal, and Panda \cite{roy2019towards} give a comprehensive overview of this path toward spike-based machine intelligence.

The further path of SNNs will be shaped by close interaction between neuroscience research and the implementation of its findings in information technology. This is not only a matter of finding more efficient solutions to problems that are already understood and solved in principle. The diverse \textbf{functional} toolbox of biological nervous systems remains a source of fundamentally new forms of information processing and learning, and so possibly also of further advances in AI as a whole.

\subsubsection{4.1.1 Spiking neural nets: the basis}

The fundamental difference between SNNs and classical ANNs is that information is no longer transmitted as continuous activation values but as brief spike events and their timing. These events drive the processing. Computation therefore takes place mainly where and when spikes occur. On event-driven neuromorphic hardware in particular, inactive regions can largely remain at rest, and this is the main basis for the potentially high energy efficiency of SNNs.

Models of neuronal dynamics are the first requirement for understanding and designing SNNs. The LIF model described in Section 3.2.2 captures, in greatly simplified form, how a neuron integrates incoming signals and generates a spike when its firing threshold is exceeded. Depending on the task, spike generation can be modeled deterministically or stochastically.

This, however, describes only the dynamics of a single spiking neuron. In its basic form, the LIF model has no memory of the timing of its past spikes and no mechanism for modifying its synaptic weights through learning. If the spike-timing-dependent plasticity described in Chapter 3 is to operate in an artificial SNN, the LIF core must therefore be supplemented with local memory and learning mechanisms.

\textbf{Integrating STDP:}

STDP needs to know whether a presynaptic spike preceded or followed a postsynaptic spike, and how large the interval between the two events was. This information must therefore be retained in the system for a short time.

A simple and widely used solution is to introduce two decaying memory traces, a pre-trace and a post-trace. The pre-trace \(x_{pre}(t)\) records the spikes arriving at the presynaptic side, and the post-trace \(x_{post}(t)\) records the spikes generated by the postsynaptic neuron. Each trace is incremented when its event occurs and then decays with its own time constant:
\begin{equation}\tag{4.1}
\begin{aligned}
\tau_{\mathrm{pre}}\frac{dx_{\mathrm{pre}}}{dt}
&=-x_{\mathrm{pre}}+\sum_k\delta(t-t_k^{\mathrm{pre}}),\\
\tau_{\mathrm{post}}\frac{dx_{\mathrm{post}}}{dt}
&=-x_{\mathrm{post}}+\sum_l\delta(t-t_l^{\mathrm{post}}).
\end{aligned}
\end{equation}

When a new spike occurs, the system can immediately check which of the two complementary traces is still present and how far it has decayed. A postsynaptic spike that occurs while a pre-trace remains can trigger strengthening of the synapse concerned, in accordance with the STDP rule. Conversely, a presynaptic spike that arrives while a post-trace remains can lead to weakening of the synapse.

In this way the originally non-learning LIF core is supplemented with a local temporal memory. The neuronal dynamics generate the spikes, the pre- and post-traces preserve their history for a short time, and the STDP rule converts the timing relation between the events into a local change in synaptic weight.

The past that is relevant to a weight change therefore does not have to be reconstructed from the activity of the whole network. For a short time it is directly available at the site where the change is to take place. This local memory mechanism can be extended with further traces. In particular, the \emph{eligibility traces} introduced later can preserve a synapse's eligibility for a weight change for longer, until, for example, a delayed reward or modulation signal arrives.

\textbf{Balancing the spiking network:}

For both coding and learning, the activity of an SNN must be kept within a suitable operating range. Without stabilizing mechanisms, learning networks in particular tend either toward excessive activity or toward falling silent. STDP can reinforce this tendency: a connection that is already strong increases the probability of postsynaptic firing and, under suitable conditions, can therefore be strengthened further.

Homeostatic mechanisms provide an important counterbalance. For instance, the total efficacy of the synapses converging on a neuron can be adjusted slowly without entirely destroying the relative weight structure that learning has produced. The excitability of the neuron itself can also be regulated. If, for example, its firing threshold is raised temporarily after a spike and then slowly decays again, neurons that would otherwise dominate permanently are held back, and others gain a better chance of responding to similar input patterns.

Another effective strategy is lateral inhibition, often in the form of a \emph{winner-take-all} mechanism. Strongly activated neurons suppress competing neurons in their neighborhood through inhibitory connections. This prevents large regions of the network from responding simultaneously to the same input pattern. Competing activity patterns emerge that can support the specialization of individual neurons or groups of neurons.

Depending on the task, such mechanisms can be supplemented with weight bounds, normalization, adjusted thresholds, and pre-trained network structures. Methods with separate phases of activity and rest are also being investigated as a way to stabilize learned structures and limit undesirable changes in the weights.

Which of these mechanisms are used depends strongly on the network architecture, the learning method, and the hardware. Modern neuromorphic platforms provide programmable mechanisms for this purpose, allowing neuronal dynamics, synaptic plasticity, and stabilization methods to be combined.

\subsubsection{4.1.2 Coding in SNNs}

\begin{figure}[htbp]
\centering
\includegraphics[width=1.0\linewidth]{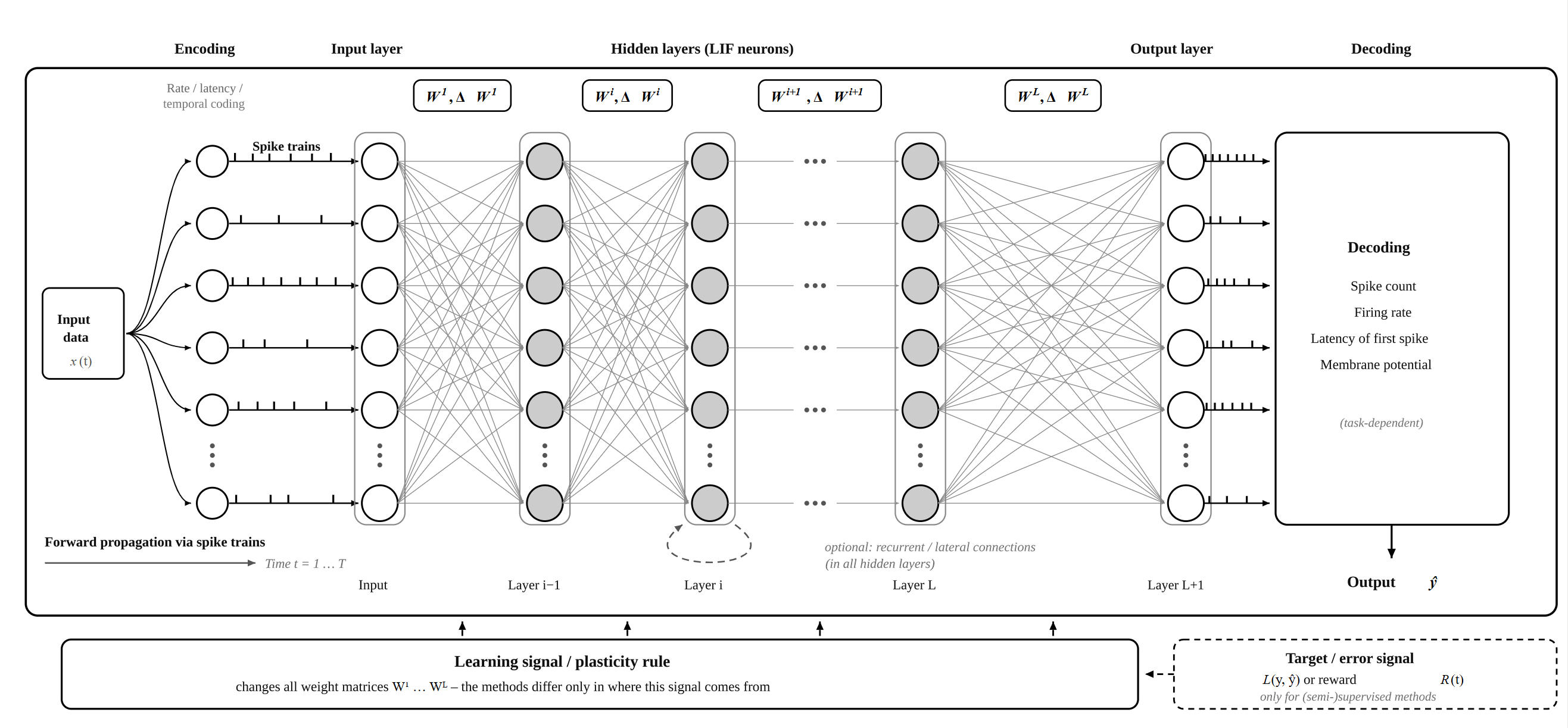}
\caption*{Fig. 4.1 Principle of a learning SNN}
\end{figure}

As a guide to what follows, Fig. 4.1 shows the basic structure and functional organization of an SNN: an input layer, hidden layers, an output layer, and the optional generation of an error or teaching signal.

Artificial SNNs use a remarkably wide range of coding and learning methods \cite{Auge2021ASO}. Classical ANNs process continuous numerical values, whereas the elementary activity of an SNN consists of discrete spike events. Information can be carried by the activity of individual neurons, by firing rates, by the distribution of activity across groups of neurons, and above all by the temporal structure of the spikes. Using time as an additional carrier of information is what substantially extends the information-processing capabilities of SNNs.

The choice of coding scheme depends on the task, the required accuracy, the robustness to noise, and the acceptable computational cost. In many control and sensing applications, a relatively coarse representation of the input variables is sufficient. Processing precise continuous quantities or high-dimensional vectors, on the other hand, requires more powerful coding schemes.

An encoder therefore maps the input data onto a spatiotemporal spike pattern that the SNN can process. Two basic approaches are examined more closely below. In rate coding, information is transmitted mainly through the frequency of the spikes; in temporal coding, through their timing. Both basic forms can be supplemented by population coding and other methods, and they can be combined with one another.

\textbf{Rate coding:}

Rate coding rests on the assumption that the essential information in a signal lies in the frequency of the spikes within a certain period, not in their exact timing. A stronger stimulus produces a higher firing rate, a weaker stimulus a lower one.

In practice, a time window \(T\) can be divided into small intervals \(\Delta t\) and filled with spikes according to a prescribed rate. The more precisely the rate is to be determined from a finite spike train, the more events, or the longer an observation time, are generally needed.

The firing rate of a LIF neuron depends directly on its input current. If the interspike interval at a constant input current \(I_{i}\) is computed from equation 3.1, the sum of the charging time up to the threshold and a refractory period \(\tau_{Refr}\) gives the characteristic current-rate relation
\begin{equation}\tag{4.2}
f_i(I_i)=
\left[
\tau_{\mathrm{Refr}}
-\tau_m
\ln\left(1-
\frac{I_\text{th}}
{I_i}
\right)
\right]^{-1}.
\end{equation}

Here \(I_{\text{th}} = \frac{V_{th} - V_{rest}}{R_{m}}\) with \(I_{i} \geq I_{\text{th}}\) is the current needed to reach the threshold. Below this threshold current, the neuron does not fire. Above it, the firing rate rises with the input current.

\begin{wrapfigure}[20]{r}{0.65\textwidth}
\centering
\includegraphics[width=\linewidth]{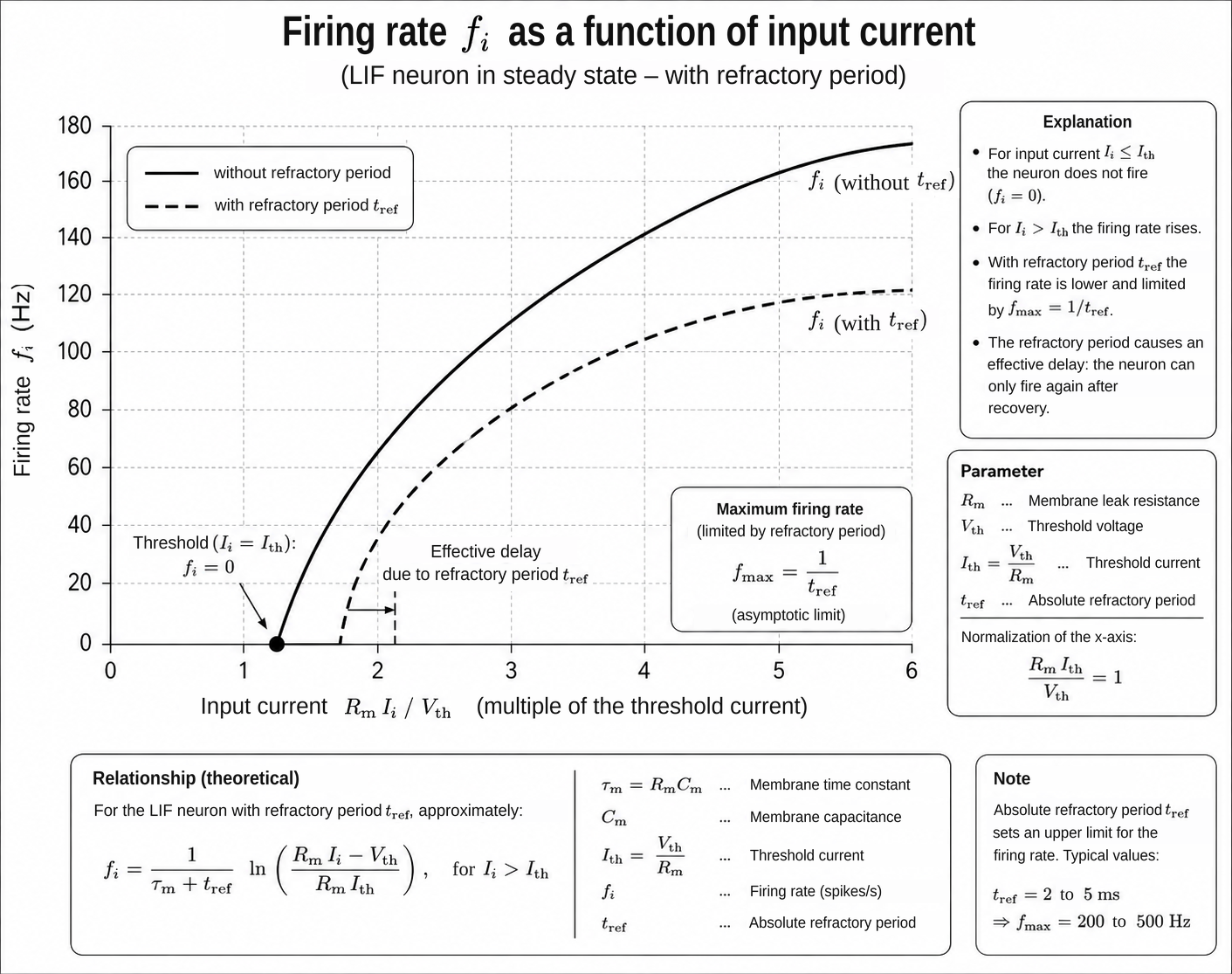}
\caption*{Fig. 4.2 The firing rate}
\end{wrapfigure}

Fig. 4.2 shows a typical curve.

One way to generate rate patterns is Poisson coding. In each small time interval \(\Delta t\), a spike is generated with a probability approximately equal to \(f_{i}\Delta t\). The individual spike times are therefore random, while their mean frequency represents the target firing rate. Compared with strictly periodic spike trains, this avoids unwanted artificial synchronization.

At the output, the received spike trains must be converted back into a value that is meaningful for the task at hand. The simplest option is to count the spikes in a fixed time window. Sliding time windows allow a faster response to changes. Another option is to filter the arriving spikes with an exponentially decaying trace, so that the current state automatically represents a time-weighted measure of recent activity.

Rate coding is comparatively robust to small shifts in individual spike times. Its accuracy, however, is limited by the available observation time and the number of spikes that can be evaluated. One way to increase both accuracy and robustness is to distribute the information across a whole population rather than a single neuron.


\textbf{Population coding:}

This principle has a biological counterpart. The direction of a movement, for example, is not represented in the motor cortex by a single neuron alone. Different neurons have different preferred directions and respond with different strengths.

It is the joint activity pattern of the population that represents the actual direction of movement precisely.

\begin{wrapfigure}[14]{r}{0.55\textwidth}
\centering
\includegraphics[width=\linewidth]{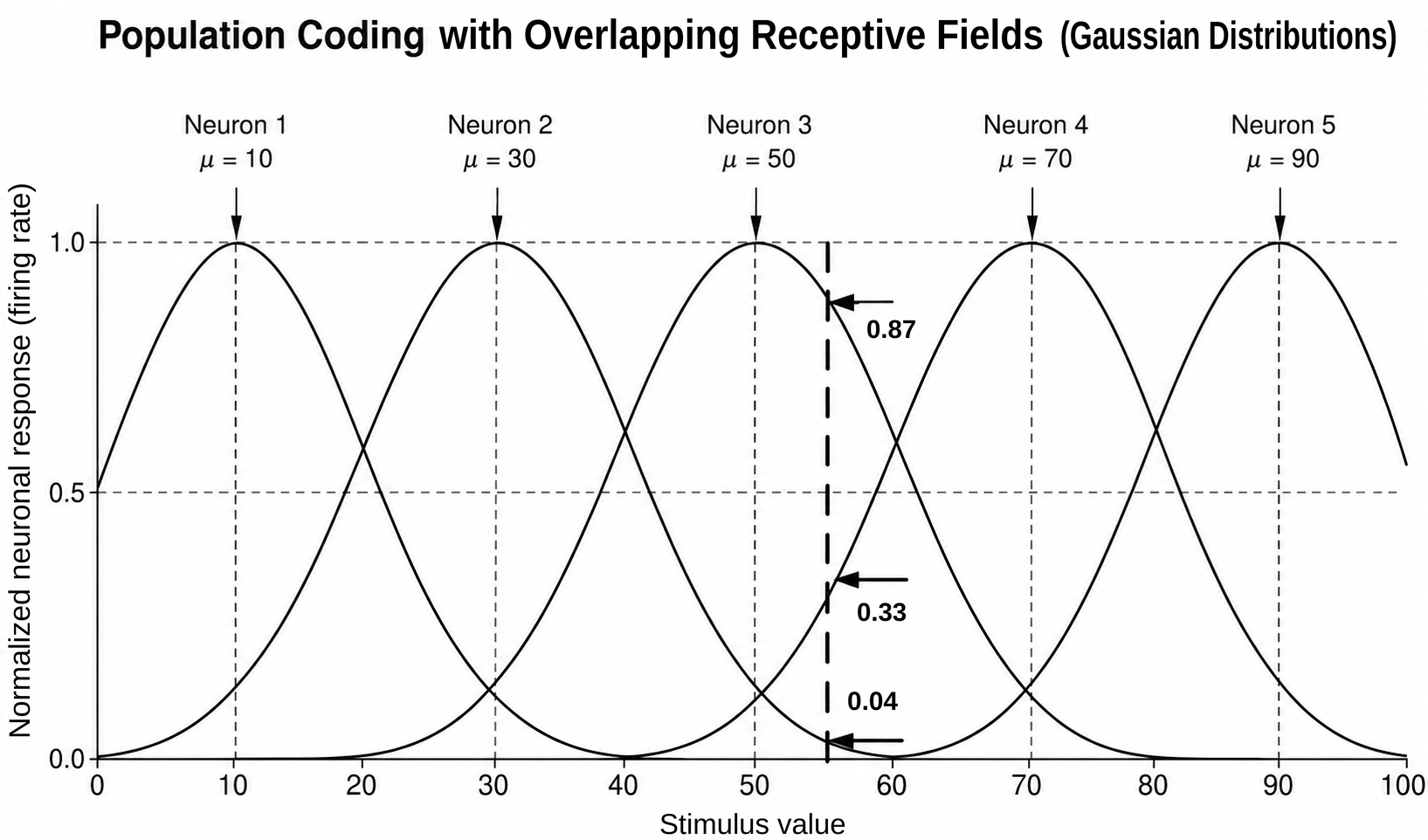}
\caption*{Fig. 4.3 Population coding}
\end{wrapfigure}

As a simple example, consider how a number \(x\) is coded by a population of five neurons. Each neuron \(i\) has a preferred value \(x_{i}\) and responds around this value with a characteristic tuning curve, for example
\begin{equation}\tag{4.3}
f_i(x)=f_{\max}
\exp\left(
-\frac{(x-x_i)^2}{2\sigma^2}
\right).
\end{equation}

A given input value therefore activates several neurons to different degrees. In Fig. 4.3, for instance, the number 55 would drive neurons 3, 4, and 2 at 87\%, 33\%, and 4\% of the maximum firing rate. The information no longer lies in the activity of a single neuron but in the pattern across the whole population. A simple decoding scheme weights the preferred values of the neurons by their measured activities \(n_{i}\):
\begin{equation}\tag{4.4}
\hat{x}
=
\frac{\sum_{i=1}^{N}n_i x_i}
{\sum_{i=1}^{N}n_i}.
\end{equation}
This simple example already shows the basic principle: a continuous value is mapped onto a distributed activity pattern and then reconstructed from it.

The same principle can be extended to \textbf{vectors}, which connects it to the forms of representation used in classical artificial neural networks. The classical population vector code assigns each neuron \(i\) a preferred direction vector \(\mathbf{c}_{i}\). The activity of the neuron depends on how well the current stimulus vector \(\mathbf{s}\) aligns with this preferred direction. For normalized vectors \(\parallel \mathbf{c}_{i} \parallel = 1\), one can write, for example,
\begin{equation}\tag{4.5}
\cos\theta_i=
\frac{\mathbf{c}_i\cdot\mathbf{s}}
{\|\mathbf{s}\|}.
\end{equation}
A simple linear tuning function is
\begin{equation}\tag{4.6}
f_i(\mathbf{s})
=
b_i+a_i(\mathbf{c}_i\cdot\mathbf{s}),
\end{equation}
where \(b_{i}\) is a baseline activity and \(a_{i}\) a gain factor. The more closely the direction of \(\mathbf{s}\) aligns with \(\mathbf{c}_{i}\), the more strongly the neuron responds.

A population of \(N\) neurons with differently oriented vectors \(\mathbf{c}_{i}\) therefore samples the \(D\)-dimensional stimulus space from different directions. The original vector is converted into a distributed activity pattern from which it can be recovered. In circuit terms, this coding can be implemented directly in the input weights. With \(w_{di} = a_{i}c_{i,d}\), neuron \(i\) receives the input
\begin{equation}\tag{4.7}
I_i=
\sum_{d=1}^{D}w_{di}s_d+b_i.
\end{equation}
Coding a vector therefore reduces to a weighted summation of its components by a population of differently tuned neurons.

The original vector must then be recovered from this distributed activity pattern. Methods of varying complexity are available for this, and we consider them in order from the simple to the more precise: first two direct, non-statistical decoders, then the statistically optimal Bayesian estimate.

\textbf{Vector decoding without statistics: the population vector:}

The center-of-mass decoding (4.4) described for a single number \(x\) extends directly to vectors. The preferred \emph{value} \(x_{i}\) is replaced by the preferred \emph{direction vector} \(\mathbf{c}_{i}\) from (4.5) to (4.7): a neuron fires more strongly the better the stimulus vector \(\mathbf{s}\) aligns with its preferred direction. The vector we want is the sum of the preferred directions weighted by the activities \(n_{i}\),
\begin{equation}\tag{4.8}
\hat{\mathbf{s}}=\sum_{i=1}^{N} n_i\,\mathbf{c}_i
\qquad(\text{normalized if required}).
\end{equation}
This method goes back to Georgopoulos et al.~\cite{Georgopoulos1986}, who used it to reconstruct the direction of movement from the activity of neurons in the motor cortex. It is still used today in brain-machine interfaces, and it requires neither a likelihood model nor a prior, only a weighted vector sum.

\textbf{Component-wise (factorized) coding:}

A simpler approach still is to give \emph{each} component \(s_{d}\) its own population. Each subpopulation codes its component as if it were a single number (tuning curves as in 4.3) and is decoded separately by a center of mass (4.4); the overall vector is then the collection of these separately estimated components, a \emph{population of populations}, as it were. The advantage is the simple structure, which grows only linearly with the number of dimensions; the price is that relations \emph{between} the components are not represented directly (the binding problem). Salinas and Abbott \cite{salinas1994} early on compared systematically the basic methods for reconstructing coded vectors from firing rates; Pouget et al.~\cite{Pouget2000} give an overview of such population codes.

\textbf{Vector decoding with statistics: Bayesian estimation:}

Both methods so far assume that the observed activity pattern reflects the underlying stimulus clearly enough. In reality, the measured spike counts fluctuate because of biological or technical disturbances. This leads to a classical estimation problem: \emph{which input vector most probably produced the observed activity pattern?}

Let \(n_{1},\ldots,n_{N}\) be the spike counts observed within a time window \(T\). For any assumed input vector \(\mathbf{s}\), we can compute the probability of this pattern. Under the simplifying assumption of independent Poisson processes, the likelihood is
\begin{equation}\tag{4.9}
L(\mathbf{s})
=
P(n_1,\ldots,n_N\mid\mathbf{s})
=
\prod_{i=1}^{N}
\frac{
\big(f_i(\mathbf{s})T\big)^{n_i}
e^{-f_i(\mathbf{s})T}
}
{n_i!}.
\end{equation}
The maximum-likelihood estimate is the vector that could most plausibly have produced the observed activity pattern:
\begin{equation}\tag{4.10}
\hat{\mathbf{s}}_{\mathrm{ML}}
=
\arg\max_{\mathbf{s}} L(\mathbf{s}).
\end{equation}
If we also have prior knowledge about which input values are more probable than others, it can be included as a prior \(P\left( \mathbf{s} \right)\). The maximum a posteriori estimate is then
\begin{equation}\tag{4.11}
\hat{\mathbf{s}}_{\mathrm{MAP}}
=
\arg\max_{\mathbf{s}}
\left[
\ln L(\mathbf{s})+\ln P(\mathbf{s})
\right].
\end{equation}

The reasoning is simple: the observed spike pattern is known, its cause is not. So we treat the possible causes as \textbf{hypotheses} and ask, for each one, how probable it is that this particular cause would have produced the observed pattern.

To do this, we specify a set of candidate vectors \(\mathbf{s}_{1},\mathbf{s}_{2},\ldots\). Each candidate \(\mathbf{s}_{j}\) is a hypothesis about the possible cause of the observed activity pattern. For each hypothesis, the expected firing rates \(f_{i}\left( \mathbf{s}_{j} \right)\) are read off the tuning curves and compared with the spike counts \(n_{i}\) actually observed.

Each hypothesis now needs a \emph{score} that indicates how well it fits the observation. This number is nothing new; it is the probability from (4.10), merely put into a form that is convenient for comparison. We take the logarithm of the Poisson likelihood (4.9). The natural logarithm turns the product of the individual probabilities into a sum:
\[
\ln L(\mathbf{s}_j)
=
\sum_{i=1}^{N} n_i\,\ln f_i(\mathbf{s}_j)
\;-\;
T\sum_{i=1}^{N} f_i(\mathbf{s}_j)
\;+\;
\underbrace{\sum_{i=1}^{N}\big(n_i\ln T-\ln n_i!\big)}
_{\text{the same for every hypothesis}}.
\]
The term under the brace depends only on the fixed observed values \(n_{i}\) and the window length \(T\), not on the hypothesis \(\mathbf{s}_{j}\). It has the same value for every candidate and therefore cannot affect the ranking. If we add the prior knowledge \(\ln P\left( \mathbf{s}_{j} \right)\) (posterior instead of pure likelihood) and drop every term that is irrelevant to the comparison, we are left with exactly the score we want:
\begin{equation}\tag{4.12}
Q_j
=
\sum_{i=1}^{N}
n_i\,\ln f_i(\mathbf{s}_j)
\;-\;
T\sum_{i=1}^{N}f_i(\mathbf{s}_j)
\;+\;
\ln P(\mathbf{s}_j).
\end{equation}
\(Q_{j}\) is therefore simply the \textbf{log-likelihood of the hypothesis} \(\mathbf{s}_{j}\), that is, the logarithm of the probability that this particular cause would have produced the observed pattern, plus the log prior \(\ln P\left( \mathbf{s}_{j} \right)\). In short:
\[
Q_j
=
\ln\big(\text{likelihood of hypothesis }\mathbf{s}_j\big)
+\ln\big(\text{prior}\big)
=
\ln P(\mathbf{s}_j\mid \text{observation})+\text{const}.
\]
We can therefore safely read \(Q_{j}\) as the (log) likelihood, or score, of hypothesis \(j\). The hypothesis with the largest \(Q_{j}\) is the same as the one with the largest probability, the MAP estimate; without the prior we obtain the maximum-likelihood decision. The apparently new expression (4.12) is not an additional postulate, only the convenient computational form of (4.11).

Remarkably, what began as a purely statistical hypothesis test can be translated directly into a neural operation. The reason is the logarithm: it has turned the product into a \emph{sum of weighted inputs}, and such a sum is precisely the operation a neuron performs in any case.

If we set the weights
\[
W_{ji}=\ln f_i(\mathbf{s}_j)
\]
and collect the terms that do not depend on the current observation in a bias \(b_{j}\), a \textbf{hypothesis neuron} \(j\) receives the input
\begin{equation}\tag{4.13}
I_j
=
\sum_{i=1}^{N}W_{ji}\,n_i+b_j,
\qquad
b_j=
-T\sum_{i=1}^{N}f_i(\mathbf{s}_j)
+\ln P(\mathbf{s}_j).
\end{equation}
Comparison with (4.11) shows immediately that \(I_{j} = Q_{j}\). Each hypothesis neuron therefore receives its score directly as its membrane input. The stored tuning curves form the weights, the prior sits in the bias, and the observed activities \(n_{i}\) are the inputs.

The important point here is that the incoming spike counts \(n_{i}\) take \textbf{two paths}. On one path, they continue through ordinary forward processing, as described earlier. On the other, a copy is branched off into a separate \emph{decoding layer}, in which each hypothesis neuron builds up its input \(I_{j}\) in parallel according to (4.12). The actual signal processing is not interrupted; the hypothesis layer listens in, so to speak, and continually evaluates which cause best explains the current pattern.

The statistical question \emph{``Which cause best explains the observed pattern?''} is thereby turned into a \textbf{neural competition}. It unfolds in a few clearly distinguishable steps and can be pictured as a race:

\begin{enumerate}

\item

After the common start signal, each hypothesis neuron begins to integrate its input current \(I_{j} = Q_{j}\). The larger the score \(Q_{j}\), the more steeply its membrane voltage rises.

\item

The neuron with the highest score reaches the firing threshold first and fires. At this point it is only a \emph{contender}.

\item

Through a shared inhibitory interneuron, its spike triggers an abrupt inhibition that pushes \emph{all} the other hypothesis neurons below threshold at once.

\item

No second neuron gets to fire. The contender becomes the \emph{winner}.

\end{enumerate}

The crucial point is that the inhibition reaches \emph{all} competitors, not only those in its immediate vicinity. Only then is a single winner left, which is exactly what the search for the \emph{one} most probable cause demands. The four steps are the phases of a single self-resolving process, not several separate rounds of competition.

The \textbf{winner-take-all mechanism} thus selects the best-scoring neuron not by an external computational rule but through the network's own dynamics. The candidate vector \(\mathbf{s}_{j}\) associated with the winner is the estimate we were looking for. The rule \emph{``take the hypothesis with the largest} \(Q_{j}\)\emph{``}, in shorthand \(\widehat{\mathbf{s}} = \arg\max_{j}Q_{j}\), is built directly into the circuit: as the survival of the most strongly excited neuron.

It is this transition that is remarkable: a probabilistic estimation task can be converted back into the dynamics of a neural network through suitable weights, bias values, and competition mechanisms.

\textbf{Temporal coding:}

In spiking neural networks, information can be represented not only by which neurons are active but also by when they spike. Depending on how precise they are, spike times can carry a great deal of information.

In biological networks, spike times are closely tied to a strong biological timing reference (the onset of a stimulus) and to the onset of the large current that the strong stimulus drives into the receiving neuron. In nature, then, a strong external event opens the time window. A strong sensory stimulus immediately drives a large current into the neuron. The membrane voltage rises extremely steeply and crosses the firing threshold very quickly. Global brain rhythms (gamma waves, for example) act as a global clock to which the neurons align their timing.

Artificial SNNs that work with explicit time windows use a global reset or trigger signal (\(t = 0\)) as an artificial timing reference. All neuron membranes are reset simultaneously to their baseline value at the start of a pass. In the commonly used \textbf{time-to-first-spike (TTFS)} scheme, a continuous input value \(x \in \lbrack 0,1\rbrack\) is converted into a firing time \(t_{spike}\) by a decreasing mapping, which may be logarithmic or linear, for example \(t_{spike} = t_{max} \cdot (1 - x)\). A maximum value \(x = 1\) produces an immediate spike at \(t = 0\). To convert the output spike train back into an interpretable value, the system uses the temporal order of the spikes. Starting from the start signal (\(t = 0\)), it waits for the first neuron in the target layer to fire and records that neuron's identity. Depending on the objective, the output is read out either channel by channel or as a learned weighted sum of the responses of all output neurons.

An individual spike carries no address of origin. In the abstract network, its origin is determined solely by the synaptic connection along which it reaches the target neuron; each connection is, in effect, a dedicated wire. Where engineered implementations instead use shared communication paths (a shared bus), this implicit assignment is lost and has to be restored by routing mechanisms, for instance by attaching the address of its source to each event (address-event representation).

In the target neuron, each arriving spike produces a postsynaptic response in proportion to its synaptic weight. The first spike finds the neuron in its initial state; later spikes encounter the traces that earlier events have left behind. The joint effect of the spikes therefore depends not only on which connections are involved but also on their order and the intervals between them.

If the firing threshold is exceeded, a new spike is generated. Its timing is determined by the preceding temporal integration and in turn becomes the input to the next layer. The received spike is therefore not simply passed on through the network; each processing stage forms a new spatiotemporal pattern from the one that arrives.

Nor is the original meaning of an individual input spike preserved on its way through the network. Instead, the information is continually transformed into new \emph{spatiotemporal activity patterns}. The identity of the active neurons forms their spatial component, and the spike timing forms their temporal component. The connectivity determines which events can interact with one another; the temporal dynamics determine how they act on one another. Finally, the output layer interprets the resulting spatiotemporal pattern according to the task at hand. In a classification task, for example, the output neuron that fires first can determine the recognized class. In other tasks, the result can be the firing time itself, relative time intervals, or quantities decoded from them. In supervised training, deviations between the desired and the generated output patterns can likewise be used to form a loss function.

TTFS is only the simplest form of temporal coding. Other methods use, for example, the rank order of arriving spikes (\emph{rank-order coding}), relative time intervals, the phase relative to a reference oscillation, or more complex patterns of several spikes. What they have in common is that they use a dimension of information that pure rate coding largely discards.

\textbf{Rate coding and temporal coding:}

The particular potential of this temporal representation becomes clear in comparison with rate coding. In rate coding, a spike train within an observation window is, in the simplest case, reduced to the number of spikes or their mean rate. The exact timing of the individual events is lost.

Temporal coding, by contrast, exploits exactly this additional information. If a single spike within a time window \(T\) can be distinguished with a usable temporal resolution \(\Delta t\), roughly \(T/\Delta t\) distinguishable time positions are available in principle. A single spike can therefore already carry substantially more information than the mere fact that a spike occurred within this window.

The difference becomes even clearer with several spikes. Whereas rate coding maps different spike trains with the same number of spikes onto the same rate value, in temporal coding the order of the events and the intervals between them can carry additional information.

Rate coding is therefore a robust but information-reducing form of coding. However, because rates have to be transmitted continuously, the network is kept far busier, and at a higher energy cost, than with temporal coding, which makes far more effective use of each individual spike.

In principle, this allows a higher information density per spike, shorter response times, and a lower energy cost.

Temporal coding has a further, particularly impressive capability: the timing information is not only transmitted but also directly shapes the learning process. Whether two spikes arrive almost simultaneously or further apart changes the response of the target neuron and with it the resulting activity pattern. Through time-dependent plasticity mechanisms such as STDP, the same temporal relation can also influence how the connections involved are modified.

The time at which a neuron fires is therefore at once \textbf{a carrier of information, a computational variable, and a learning signal.}

This threefold role is the source of the fundamental potential of time-coded spiking neural networks.

This advantage, however, comes at the cost of increased demands on temporal precision.

Nevertheless, advances in hardware, sensors, temporal synchronization, and event-based signal processing, together with appropriate tolerance margins, can further increase the usable temporal resolution and reliability. What matters is the number of temporal patterns that can be reliably distinguished under real conditions.

The potential of temporal coding is therefore likely to grow further as these temporal dynamics are brought under control. Whereas rate coding discards part of the available timing information in favor of robustness, temporal coding seeks to make exactly this additional dimension usable for information transmission, computation, and learning.

\subsubsection{4.1.3 Learning and training in SNNs}

The early engineering work on SNNs met with considerable skepticism. In particular, it was unclear whether the initially unfamiliar world of temporally distributed spike events could be turned into powerful and controllable learning procedures in the first place.

A demonstration by Bodo Rueckauer and co-workers in 2017 \cite{rueckauer2017conversion}, building on several earlier studies, was therefore of great practical importance: under suitable conditions, pre-trained ReLU networks can be converted into rate-coded SNNs. In such a conversion, the mean firing rate of a spiking neuron approximates the continuous activation of the corresponding ANN neuron. This made it possible to transfer even complex pre-trained networks to event-driven SNNs with only small losses in accuracy: a network is first trained with the established methods of classical ANNs and then converted into an SNN. The real challenge, however, was not solved by this approach but circumvented. For systems that have to learn online, on-chip, or without a pre-trained ANN template, the development of SNN learning methods in their own right remained unavoidable.

Since then, a bewildering variety of such methods has emerged. Driven by demands for training efficiency and low energy consumption, most of this work has followed an implicit guiding principle: to bring learning as close as possible to the individual neuron, not necessarily at the moment of its activity, but with no more delay than the network can tolerate for a corrective update.

How much it matters for learning that a corrective step occurs close in space and time to a neuron's activity naturally depends on the operating conditions, that is, on

\begin{enumerate}

\item

whether the weights can be adjusted separately from later use, or whether the network must continue learning while it is processing (offline or online), and

\item

whether the correction information must be fed back through the whole network from a global error function, or arises largely locally from the immediately available states of neurons and synapses (local or global).

\end{enumerate}

A broad spectrum of learning methods now covers this problem space. It ranges from fully local, self-organized learning without an explicit learning objective, through locally stored eligibility traces combined with additional evaluative signals, to global optimization with surrogate gradients.

\textbf{Self-organized learning:}

The most local form is unsupervised learning based on the STDP described in Sections 3.2.3 and 3.2.4. There is no explicit target or error signal here. With the help of mechanisms such as lateral inhibition, however, neurons or groups of neurons can specialize in recurring input patterns and form structured representations.

\textbf{Local and trace-based learning:}

Learning methods in which the information needed for a later correction is stored locally during ongoing processing are particularly important and attractive for neuromorphic systems. The tool borrowed from nature is the \emph{eligibility trace} already introduced in Chapter 3. Eligibility traces make it possible to assign a reward or punishment, at the right moment, to the synapses that were involved, a problem known as credit assignment.

When a synapse takes part in a processing step, it often only becomes clear later whether its contribution was useful for the result. By then the original spike pattern is long gone. An eligibility trace therefore keeps a decaying record of the synapse's earlier participation at the synapse itself. If an evaluative signal arrives later, it can act preferentially on those synapses whose traces are still present.

The evaluative signals differ considerably between methods. In biologically inspired three-factor learning rules \cite{fremaux2016neuromodulated,gerstner2018eligibility}, for example, a neuromodulator supplements the pre- and postsynaptic activity. A reward or error signal can be broadcast widely across the network, while the local eligibility trace determines which synapses respond to it at all.

In this kind of reward-modulated learning, an eligibility trace is merely a coincidence marker: it records which synapses were recently active together and waits for a single global reward signal that subsequently decides which of the marked synapses are reinforced. The trace itself, however, carries no information about the specific error.

A valuable extension of this idea is therefore the eligibility propagation learning rule (e-prop) developed by Maass's group \cite{bellec2020eprop}. Here, a continuously updated eligibility trace is formed. In contrast to classical backpropagation through time, the information about the participation of individual synapses is propagated forward in time as local traces, where it can meet learning signals that arrive later. If, during training or in online learning, the single global reward signal is replaced by a neuron-specific error signal, the running eligibility trace can produce a partial error correction at each time step: wherever the error signal meets a trace that has not yet decayed, that synapse is corrected. The complete correction is the sum of these momentary partial corrections over the whole run.

Naturally, the quality of learning depends to a large degree on how informative and how specific the additional learning signals are. Between a single global reward signal and a precise, neuron-specific assignment of error there is therefore a broad spectrum of different learning methods.

\textbf{Global learning with surrogate gradients:}

For training complex SNNs directly, learning with \emph{surrogate gradients} \cite{neftci2019surrogate} is among the most powerful methods. It carries the basic idea of gradient-based optimization in classical ANNs over to spiking networks. As mentioned earlier, the non-differentiability of the hard spike threshold is dealt with by substituting a smooth surrogate function during the backward pass. The weights of the temporally unrolled SNN can then be optimized with methods similar to backpropagation through time.

This approach, however, also inherits at least some of the disadvantages of backpropagation through time: intermediate network states across time must be stored for the backward computation, and the correction information is fed back globally through the unrolled network. As a result, key advantages expected from strictly local, event-driven neuromorphic learning are lost.

Surrogate-gradient methods are by no means merely a ``fallback'', however. At present they form a powerful bridge between deep learning and SNNs. From a neuromorphic point of view, though, the goal of replacing as much as possible of the processing needed for learning with local mechanisms that are available online remains attractive.

\textbf{Conversion instead of learning:}

Converting a pre-trained classical network into an SNN, the possibility described at the beginning of this section, remains a common route to putting existing models to practical use. The SNN takes over the learned weights and converts the continuous activations into a suitable spike representation.

Overall, a spectrum of learning methods has emerged. At one end is powerful global optimization, with its high demands on information and memory; at the other are local, event-driven learning mechanisms that make do with the information directly available in the network.

A central direction of development in neuromorphic AI is to move the correction information needed for successful learning as close as possible to the place and time where it originates. The architectural measures described in the next section can support this by structuring the emerging spike patterns as they propagate, which simplifies the task of the learning mechanisms that follow.

\subsubsection{4.1.4 Network design and learning}

How an SNN is designed depends heavily on the class of problem and on the intended range of applications. Current research and development focus both on improving the ability to learn increasingly complex tasks and on usability and the creation of powerful hardware and software platforms. Transferring basic elements of the transformer architecture to spiking networks is itself now an active area of research \cite{zhou2023spikformer}.

The design of the network architecture plays a crucial role. It can be made to carry a substantial part of the work of organizing learning. In SNNs, competition, inhibition, recurrence, and dedicated signal pathways determine during processing which activity patterns arise, which persist, and which neurons are candidates for a weight change in the first place. They therefore influence the learning process to a far greater degree than in ANNs.

The architecture guides \emph{where} and \emph{when} learning can take place; the learning rule then determines \emph{how} the connections concerned are modified.

The architecture is shaped largely by the deliberate use of inhibitory synapses. The following pairings, each matched to a learning method, illustrate the design principle.

\textbf{STDP, lateral inhibition, and winner-take-all:} 

Unsupervised STDP learning produces self-organizing patterns, but these patterns are not ordered. What is missing is guidance that decides which of several similarly excited neurons should become responsible for a particular input pattern. Without additional competition, many neurons can therefore learn similar features.

This ``chaotic'' process can be brought into order by using lateral inhibition deliberately. When a strongly excited neuron fires, it suppresses competing neurons in its neighborhood through inhibitory connections. When this competition is pronounced, winner-take-all behavior arises: the neuron that is activated first, or most strongly, prevails over its competitors and so gets the first chance to adapt its synapses through STDP.

In this way, separate regions of representation, or clusters, emerge. The same principle of competitive self-organization underlies Kohonen's well-known self-organizing maps \cite{kohonen1982som}. A well-known example in the spiking case is the unsupervised classification of handwritten digits using STDP and lateral inhibition alone \cite{diehl2015unsupervised}.

Analog winner-take-all circuits for the neuromorphic implementation of this competition principle, used in Intel's Loihi chips among other systems, were developed by Giacomo Indiveri and co-workers, among others \cite{Indiveri1998}.

\textbf{Local learning, recurrence, and attractors:} 

A frequent problem in SNNs is that a whole activity pattern has to be kept stable for a period long enough to matter for local learning. Recurrent connections can stabilize activity patterns or sustain them for a certain time, in a kind of reverberating loop, and so create a temporal working window for subsequent processing steps. Combined with inhibitory competition mechanisms, this produces dynamic states in which the network preferentially settles on particular activity patterns, which can be understood as attractors of the network dynamics.

Eligibility traces do not strictly require such an attractor; they already carry a local memory of preceding activity on their own. The two mechanisms can, however, complement one another effectively: the eligibility trace at the individual synapse records \emph{who was involved}, while recurrent network dynamics keep the associated activity pattern available for longer at the level of a neuronal ensemble.

The synapse thus preserves a local trace of participation, while the recurrent network can stabilize the functional context of that participation. If a reward or learning signal arrives later, this increases the chance that both the relevant synaptic traces and parts of the associated activity state are still available.

Because recurrent excitation on its own can easily lead to uncontrolled amplification of activity, inhibition plays an important role here as well. Inhibitory feedback limits activity, separates competing states, and so stabilizes the network dynamics. Work on competitive and recurrent neural fields, by Grossberg \cite{grossberg1973contour} among others, and attractor models of neuronal populations \cite{Zhang1996} have contributed substantially to the theoretical description of such stabilization mechanisms.

The architecture thus stabilizes the temporal and functional relation between the eligibility trace and the activity pattern, so that the trace can guide learning more effectively.

\textbf{Gating, broadcast learning signals, and hierarchical networks:} 

For a learning method to be effective, a broadcast learning signal must reach the parts of the network responsible for the error without the complete gradient having to be computed backward through the whole temporally unrolled network.

Here the architecture can take over a substantial part of credit assignment and make large parts of the network accessible to the learning signals in a structured way.

Hierarchical SNNs can be divided functionally into regions or processing levels by gating mechanisms. Inhibitory circuits act here like controllable gates: when closed, they suppress the propagation of an activity pattern; when the inhibition is temporarily lifted, the pattern can pass into another region of the network.

Such gates (which act as access controls for particular regions) can therefore, in principle, not only structure the forward flow of information but also determine which regions are accessible, or particularly receptive, to an additional learning signal. The architecture then provides a spatial preselection, while the eligibility traces within the opened region perform the temporal selection of the synapses actually involved.

The result is a two-stage addressing scheme:

\begin{quote}  
\emph{Gating determines the relevant region of the network; eligibility traces determine the synapses involved.}
\end{quote}

This is not yet a universal replacement for backpropagation. But it does show a fundamentally different approach to credit assignment: instead of an exact chain of gradients running backward, architecture, local memory, and a comparatively simple learning signal can act together.

\begin{quote}
\emph{The more precisely the correction information can be routed in this way by the structure and dynamics of the network itself, the less global backward computation is required. This is an important area of development for SNNs that learn online and on-chip.}
\end{quote}

Instead of reconstructing the necessary correction information globally, as backpropagation does, the aim here is to generate, store, and selectively provide this information locally, while processing is under way.

\textbf{Cluster separation and graded event processing:} 

A further advantage of architecturally structured SNNs is that learning and processing operations can be concentrated on the regions of the network that are actually active. On its own, event-driven processing means only that inactive neurons trigger no spike operation. For this to yield an efficient learning process as well, however, it must be established as unambiguously as possible which neurons and synapses are responsible for a particular activity pattern.

Again, competition, lateral inhibition, and the functional separation of populations help. They concentrate activity in particular regions and prevent any input pattern from provoking a large number of nearly identical responses. The same architectural measure that contributes to feature separation in self-organized learning therefore also reduces the number of elements that are candidates for a learning correction.

Another option is to give spike events additional information beyond their time of occurrence. Neuromorphic systems can use various forms of per-event payload or amplitude-like quantities for this purpose, as implemented in the Loihi 2 project. Such \emph{graded events} extend the strictly binary notion of a spike and make it possible to grade the effect of an event more finely.

For the learning process, this opens up the possibility of not giving every event the same weight. The size of a weight change can, for example, additionally depend on a modulating value, a local state, or a transmitted evaluative signal. Processing remains event-driven but gains an additional means of grading its effect.

The real gain in efficiency does not come from graded spikes alone. It comes from the interplay of several mechanisms: the architecture limits the set of neurons involved, eligibility traces mark the temporally relevant synapses, and additional modulating quantities can then determine how much these synapses are changed.

In this way, a learning task that at first appears global becomes, step by step, an increasingly localized operation.

\textbf{A neuromorphic example:} 

So far we have mostly considered rules that follow neuronal and synaptic functioning. Populations such as an ant colony learn in an entirely different way: in a distributed manner, and astonishingly skillfully at that. When searching for food, individual animals repeatedly probe short trial paths, test their success, lay down pheromone trails, and draw whole columns of workers along behind them. If a path leads nowhere, the next attempt follows, and the old trail fades. No animal has an overview of the whole; the order arises from nothing but local reinforcements.

The global learning objective is thereby dissolved into many sub-objectives, and global correction of the network gives way to local consolidations. Corrections at this level no longer need a global measure of error, only a local signal of success.

That such strategies can be exploited in engineered systems has long attracted attention in both worlds, classical ANNs as well as SNNs. This concerns in particular the local and temporally bounded decomposition of the learning objective (hierarchical sub-objectives or layer-specific loss functions, for instance) and the stage-wise consolidation of the network (elastic weight consolidation, metaplasticity, or the freezing of fully trained parts, for instance). On closer inspection, these two aspects are two sides of a single fundamental question: can learning, which classically presupposes a global authority, (a global objective, a globally propagated error, a correction mechanism overseeing the whole network), be dissolved into purely local processes that decide independently?

The first decomposition localizes the objective: a single loss function spanning the whole network is replaced by locally and temporally bounded sub-objectives that each region can pursue for itself (hierarchical sub-objectives, layer-local losses, running error signals). The second decomposition localizes the time scale: instead of continually adjusting all weights, the network decides section by section what may count as finished and be held fixed, and what must remain plastic (elastic weight consolidation, metaplasticity, the freezing of matured parts). The first thus decomposes what is optimized, the second what of the result is retained.

Taken together, they touch on the two old questions faced by every learning system: the credit assignment problem (how does a local element know its share in a global result, in space and time?) and the stability-plasticity dilemma (how does a network preserve what it has learned without closing itself off from the new?). Biology answers both questions locally and at the molecular level.

In the engineering repertoire, both decompositions are already present, but in scattered form. Decomposition of the objective: in ANNs, layer-local losses and hierarchical reinforcement learning; in SNNs, local STDP and e-prop. Consolidation: in ANNs, elastic weight consolidation (EWC \cite{kirkpatrick2017ewc}) and low-rank adaptation (for fine-tuning); in SNNs, metaplasticity and the fixed reservoir. Much of this has been tried, but these are individual pieces, not yet a principle.

Biology offers a principle, and a molecular one at that. Its clearest building block is the synaptic tagging mentioned in Section 3.2.5 \cite{frey1997synaptic}: an active synapse leaves behind, for a short time, a local, fading marker; only when the cell, prompted by a sufficiently strong event, provides plasticity-related proteins do the tagged synapses ``capture'' them and become permanently consolidated. A purely local marker thus decides which connection endures: the molecular archetype of the eligibility trace.

What the surface and adhesion molecules (integrins, cadherins, neurexins and neuroligins) promise reaches further still. Some of their effects are established; integrins, for instance, mediate the late consolidation phase of LTP \cite{babayan2012integrin}. But their full range of functions, from guiding synaptic growth to stabilization, has so far been mapped only in outline \cite{park2016integrins}. This still open field very probably holds a considerable store of mechanisms for local differentiation and retention that engineering has yet to exploit.

Fully realizing the potential of these cybernetic models seems more readily achievable in spiking networks: with native eligibility traces, STDP, and related local rules, they already implement basic mechanisms to which molecular differentiation could be linked directly.

\newpage

\textbf{Network design in classical ANNs and in SNNs:}

Classical ANNs also use various forms of network structuring (normalization as divisive inhibition, excitatory-inhibitory separation, lateral or competitive layers, and sparsity mechanisms), but they remain confined to a spatial and structural organization and can exploit the fine temporal structure of processing and plasticity, as provided by spiking networks, only to a limited extent.

Both worlds, however, share a basic idea: the architecture pre-shapes the activity patterns and so relieves learning of part of its burden. Yet only in the spiking system does this pre-shaping have consequences for energy: because computation there is event-driven and consumption follows the event rate, every structurally generated sparsity is at once a saving in learning and in energy. In a conventional ANN, which is clock-driven and computes every unit at every step, the same sparsity brings no energy saving. The event-driven character is therefore the real reason why structure and efficiency come together only in the spiking case.

\textbf{The prospects of neuromorphic learning:}

In sum, an impressive variety of methods today exploits the interaction between network geometry and the temporal dynamics of spiking systems. The underlying idea is remarkably consistent across these methods: the architecture pre-shapes the activity patterns as they arise, so that the learning mechanism that follows can manage with less global control.

One of the most important design advantages of spiking neural networks accordingly lies in the way network structure, temporal dynamics, and learning interlock. The network's own dynamics can take over part of the organizational work that in classical ANNs must be performed by global optimization.

This also explains the growing interest in transferring concepts from modern, highly complex AI architectures to spiking systems. Transformers and SNNs do rest on very different processing principles. What they have in common, however, is the central question of how the currently relevant interactions can be selected and reinforced out of a multitude of possible ones. Whereas transformers implement this selection essentially through computed attention weights, SNNs can additionally use their temporal dynamics, competition, and event-dependent gating mechanisms for it.

Whether a new generation of highly complex event-driven AI systems will emerge from this in the long run remains open. The direction, however, is clearly recognizable: the more ordering, selection, and learning decisions can be made locally by the structure and dynamics of the network itself, the smaller the share of global optimization work that must be brought to every point of the system from outside.

\subsubsection{4.1.5 The Neural Engineering Framework}

In recent years, SNN chips have become available to a wide range of users. Interest in using them, above all in mobile, energy-efficient systems, has grown accordingly, and so has the demand for the development tools this requires. The Neural Engineering Framework (NEF), published as early as 2003, fills exactly this gap. Chris Eliasmith and Charles Anderson conceived it as a unified, systematic theory and presented it in their book \emph{Neural Engineering} \cite{eliasmith2003neural}, which is structured like a classical academic textbook. Together with the programming environment Nengo \cite{bekolay2014nengo}, it provides an easy-to-use methodology for programming SNNs.

The approach takes a view of SNNs that is diametrically opposed to that of the learning-based approaches. Instead of learning synaptic weights, it attempts to derive the desired function of a spiking network, at least in part, directly from a mathematical description of the task and then to translate the result into a neural network. The NEF is a particularly advanced version of this approach.

It connects two worlds that look very different at first glance: the continuous world of mathematical quantities, vectors, and functions, and the world of neural activity and discrete spike events. The bridge between them is population coding. Continuous numbers are represented by the activity of many neurons, transformed mathematically, and the result is converted back into the activity of a population of neurons.

A developer therefore uses the NEF like a compiler. With the help of Nengo, the problem is broken down into sub-steps that can be implemented neurally, and the NEF compiler then turns these into rate-coded operations in neural populations.

The NEF combines three basic tasks:

\begin{itemize}

\item

the representation of scalars and vectors by neuronal populations,

\item

the transformation of these represented quantities through connections between populations,

\item

the implementation of dynamical systems through recurrent neural networks.

\end{itemize}

In this way the NEF treats a spiking network as a dynamic computing medium that can be designed. The function of the network does not have to be found through training alone; for known mathematical relationships it can be constructed directly.

The procedure is as follows. First, a vector \textbf{x} is encoded in the neuronal activity of a population. This is done through the tuning curves that correspond to the individual preferred directions of the neurons (see 4.1.2). These produce an activity pattern in the network that the next layer must decode, that is, recognize. For this purpose, an optimization procedure determines the decoders, which together with the encoders determine the decoding connections, that is, the synaptic weights linking to the next population. Generalizing this procedure from identity decoding to the decoding of arbitrary functional relationships between input and output vectors, \textbf{g(x)}, is the essential cornerstone of the NEF, and including the dynamic processes within the neurons makes it possible to design dynamic functions as well.

It is worth looking at this procedure in somewhat more detail.

The first task is to define a \textbf{representation of a vector} suited to a population. The starting point is a \(D\)-dimensional vector
\[
\mathbf{x}\in\mathbb{R}^{D}.
\]
As discussed in Section 4.1.2, it is to be represented by the activity of a population of \(N\) neurons. To this end, each neuron has a preferred direction vector \(\mathbf{e}_{i}\). Its response depends on how strongly the current input vector points in this direction. The input current received by neuron \(i\) is therefore
\begin{equation}\tag{4.14}
J_i(\mathbf{x})
=
\alpha_i\,\mathbf{e}_i\cdot\mathbf{x}
+
J_i^{\mathrm{bias}},
\end{equation}
where \(\alpha_{i}\) denotes the gain of the neuron and \(J_{i}^{bias}\) its bias. The neuronal nonlinearity \(G_{i}\), which the NEF treats in a more general form, converts this input into the activity
\begin{equation}\tag{4.15}
a_i(\mathbf{x})
=
G_i\!\left[
\alpha_i\,\mathbf{e}_i\cdot\mathbf{x}
+
J_i^{\mathrm{bias}}
\right].
\end{equation}
For the LIF neurons considered here, \(G_{i}\) is given, for example, by the mean firing rate that follows from the membrane model. We will assume this case from here on, so that \(a_{i}\left( \mathbf{x} \right)\) corresponds to the firing rate discussed earlier,
\begin{equation}\tag{4.16}
a_i(\mathbf{x})=f_i(\mathbf{x}),
\end{equation}
and in a network that actually spikes, its place is taken by the corresponding spike train. The NEF thus exploits the close relation between spike activity and mean neural response to have spiking neurons represent continuous numbers.

The vector \(\mathbf{x}\) is represented by the joint activity pattern
\[
f_1(\mathbf{x}),f_2(\mathbf{x}),\ldots,f_N(\mathbf{x})
\]
of the whole population. Neurons with different preferred directions each represent the vector from their own point of view. Only collectively do they provide a sufficiently precise representation.

\textbf{Decoding:}

To recover the original vector from this distributed activity pattern, one forms a weighted linear combination of the neuronal activities:
\begin{equation}\tag{4.17}
\hat{\mathbf{x}}
=
\sum_{i=1}^{N}
f_i(\mathbf{x})\,\mathbf{d}_i.
\end{equation}
The vectors \(\mathbf{d}_{i}\) are called \emph{decoders}. They are chosen so that the reconstructed vector \(\widehat{\mathbf{x}}\) approximates the original vector \(\mathbf{x}\) as closely as possible over a specified range of possible input values.

To achieve this, one must require the best possible agreement between \(\mathbf{x}\) and \(\widehat{\mathbf{x}}\) for as many sample values of \textbf{x} as possible.

For this purpose, \(M\) representative sample points
\[
\mathbf{x}^{(1)},\ldots,\mathbf{x}^{(M)}
\]
are selected. For each of these points, the activity of all \(N\) neurons is computed. These activities form the matrix
\begin{equation}\tag{4.18}
A=
\begin{pmatrix}
f_1(\mathbf{x}^{(1)}) & \cdots & f_N(\mathbf{x}^{(1)})\\
\vdots & & \vdots\\
f_1(\mathbf{x}^{(M)}) & \cdots & f_N(\mathbf{x}^{(M)})
\end{pmatrix}.
\end{equation}
The associated target values, in the simplest case the vectors \(\mathbf{x}^{(1)},\ldots,\mathbf{x}^{(M)}\) themselves, are collected row by row in a target matrix \(X\). We are therefore looking for a decoder matrix \(D\) such that
\begin{equation}\tag{4.19}
AD\approx X
\end{equation}
holds. Since far more sample points than neurons are chosen (\(M \gg N\)), the system of equations is overdetermined and in general has no exact solution. The decoders are therefore determined as the solution of the least-squares problem:
\begin{equation}\tag{4.20}
D
=
\arg\min_D
\|AD-X\|^2.
\end{equation}
The minimum is reached where the derivative of the error with respect to \(D\) vanishes. Setting this derivative to zero,
\[
\frac{\partial}{\partial D}\,\|AD-X\|^2
=
2\,A^{\mathrm T}\!\left(AD-X\right)
\;\stackrel{!}{=}\;0
\]
yields the \emph{normal equations}
\[
A^{\mathrm T}A\,D
=
A^{\mathrm T}X ,
\]
and, provided \(A^{T}A\) is invertible, their solution
\[
D
=
\left(A^{\mathrm T}A\right)^{-1}A^{\mathrm T}X .
\]
This matrix \(\left( A^{T}A \right)^{- 1}A^{T}\) is exactly the \emph{pseudoinverse} \(A^{+}\). The compact notation
\begin{equation}\tag{4.21}
D=A^{+}X
\end{equation}
is the condensed form of the normal equations.

This step, however, has a problem. The matrix \(\Gamma = A^{T}A\) appearing here is simply the correlation matrix of the tuning curves. Its entry
\[
\Gamma_{ij}
=
\sum_{m=1}^{M} f_i(\mathbf{x}^{(m)})\,f_j(\mathbf{x}^{(m)})
\]
measures how similarly neurons \(i\) and \(j\) respond across all sample points. If two or more tuning curves resemble one another or are partly redundant, which is typical when there are many neurons, the corresponding rows of \(\Gamma\) become nearly linearly dependent. Some eigenvalues of \(\Gamma\) then approach zero. Their reciprocals in \(\Gamma^{- 1}\) grow correspondingly without bound, and even tiny disturbances, such as spike noise in the measured activities, are amplified into large decoder values of opposite sign. The naive pseudoinverse does not give a stable solution. What we have here is not a harmless least-squares problem but an ill-posed problem in Hadamard's sense, and solving it requires regularization.

This is usually done by \textbf{Tikhonov regularization}, which in intuitive terms means adding a constant to the diagonal. One replaces the correlation matrix by
\[
\Gamma \;\longrightarrow\; \Gamma+\lambda I ,
\]
that is, one raises every diagonal entry by the same amount \(\lambda > 0\). Geometrically, this shifts \emph{all} eigenvalues upward by \(\lambda\): the eigenvalues that were dangerously small are moved away from zero, their reciprocals remain finite, and the nearly redundant directions can no longer blow up the decoder. The regularized solution is
\begin{equation}\tag{4.22}
D=
\left(A^{\mathrm T}A+\lambda I\right)^{-1}
A^{\mathrm T}X .
\end{equation}
The parameter \(\lambda\) has an intuitive meaning here: it corresponds to the expected \emph{noise level} of the activities, and in the NEF, \(\lambda\) is obtained directly from the assumed spike-noise strength \(\sigma\). A small \(\lambda\) gives accurate but noise-sensitive decoders; a larger \(\lambda\) smooths the solution and makes it robust at the cost of a slight systematic bias. \(\lambda\) thus controls the classical trade-off between \emph{accuracy and susceptibility to noise}. In practical computation, one usually replaces explicit inversion with a singular value decomposition of \(A\), which exposes the small singular values directly and allows them to be damped in a controlled way: the numerically stable version of the same idea.

This completes the first essential step: a continuous number or vector can be recovered approximately from the nonlinear activity of many neurons.

\textbf{Computing a function:} The key step in the NEF is that the decoders do not have to reconstruct the original vector \(\mathbf{x}\). They can instead be chosen directly so that they approximate a desired function
\begin{equation*}
\mathbf{y}=g(\mathbf{x})
\end{equation*}
In place of the requirement \(\sum_{i}^{}f_{i}\left( \mathbf{x} \right)\mathbf{d}_{i} \approx \mathbf{x}\), one now requires
\begin{equation}\tag{4.23}
\sum_i
f_i(\mathbf{x})\mathbf{d}^{\,g}_i
\approx
g(\mathbf{x}).
\end{equation}
The first step is to compute the target values \(g\left( \mathbf{x}^{(1)} \right),\ldots,g\left( \mathbf{x}^{(M)} \right)\) at the chosen sample points. Collecting them in the target matrix \(Y_{g}\) gives
\begin{equation}\tag{4.24}
A D_g\approx Y_g
\end{equation}
and, in regularized form,
\begin{equation}\tag{4.25}
D_g=
\left(A^{\mathrm T}A+\lambda I\right)^{-1}
A^{\mathrm T}Y_g .
\end{equation}
A notable feature of this result is that the quantity that is actually ill-posed, the regularized inverse \(\left( A^{T}A + \lambda I \right)^{- 1}\), does \emph{not} depend on the desired function. It depends only on the neurons. It is computed once; for a different function, only the right-hand side \(A^{T}Y_{g}\) changes. This makes the procedure very efficient: \emph{with different decoders, the same neural population can provide different functions of the variable it represents.}

\begin{figure}[htbp]
\centering
\includegraphics[width=0.8\textwidth]{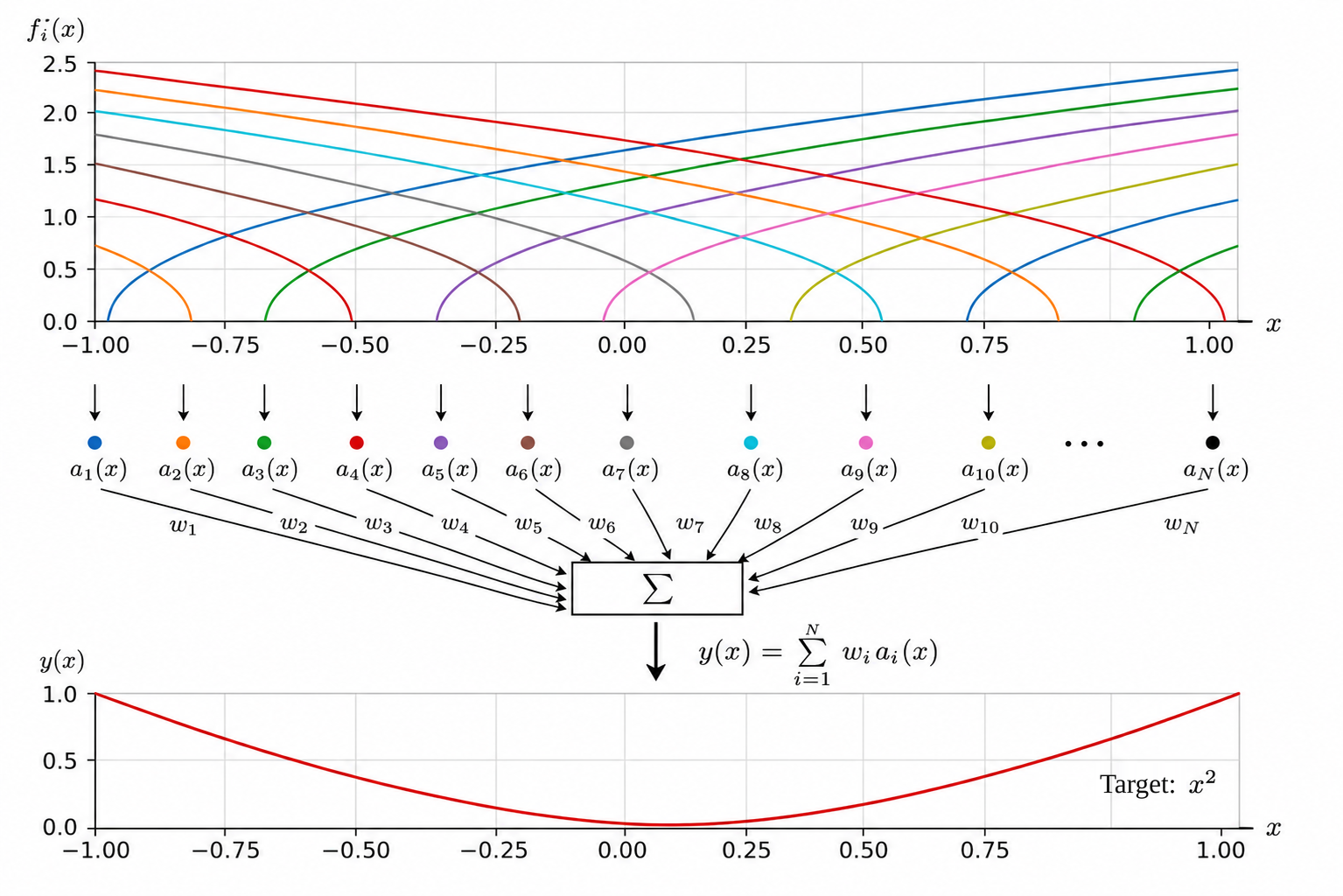}
\caption*{Fig. 4.4 Tuning curves for representing the input variable}
\medskip
\medskip
\includegraphics[width=0.8\textwidth]{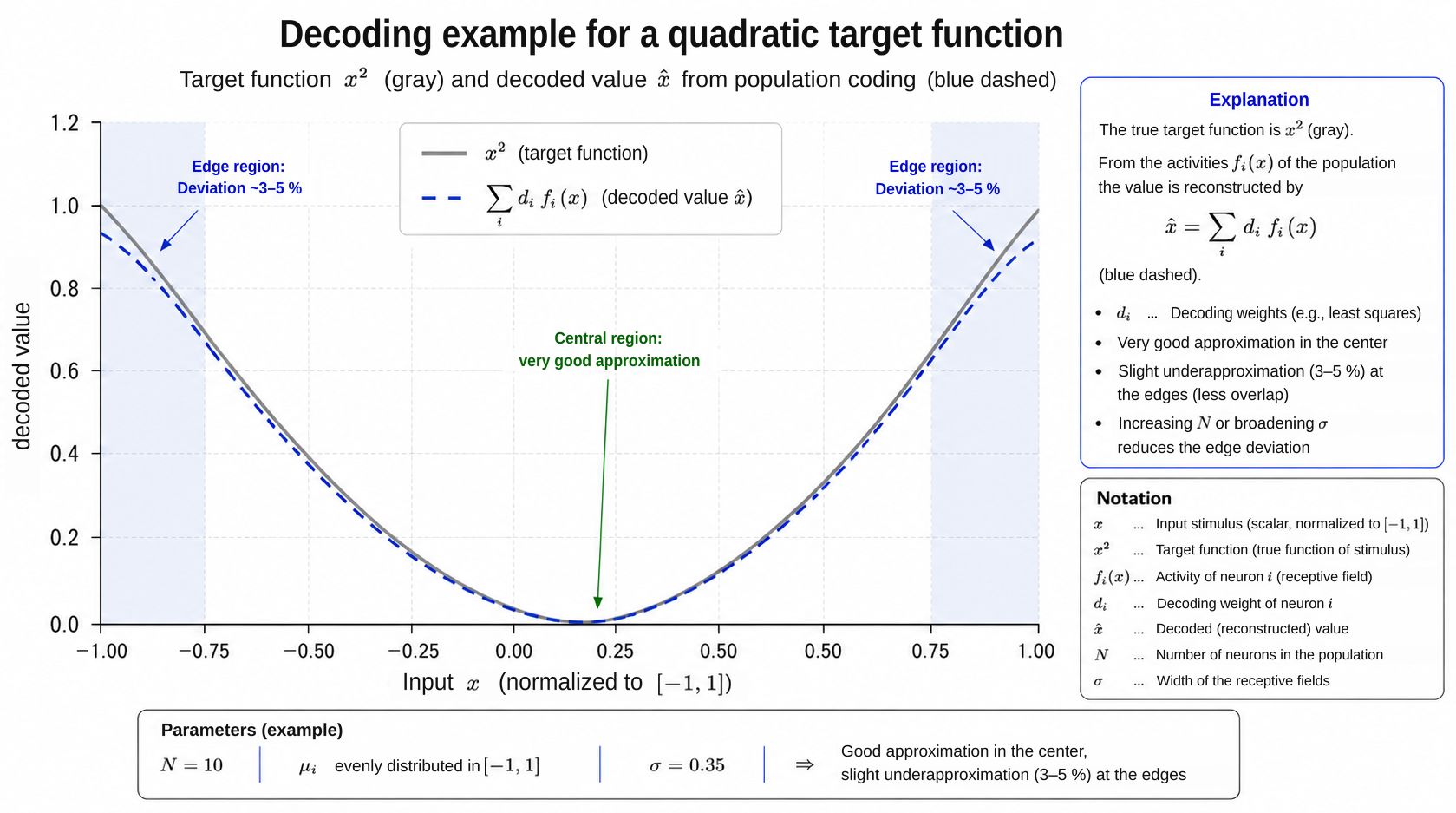}
\caption*{Fig. 4.5\quad Approximation of the function $y=x^2$}
\end{figure}

The NEF thus builds a remarkable bridge between the numerical and the neural domains. On one side is an ordinary mathematical function \(g\left( \mathbf{x} \right)\); on the other side there are only neurons, currents, and spikes. The decoders translate between the two descriptions.

The procedure is easiest to see in the example of \textbf{squaring a number}, \(g(x) = x^{2}\):

A population of differently tuned neurons, as in Fig. 4.4 above, first represents the input value \(x\). Each neuron has its own tuning curve and therefore responds differently to the same value. Together, these responses form a characteristic population pattern.

The decoders are then determined for the target function \(x^{2}\). Fig. 4.5 shows clearly what the method actually does: no individual neuron computes a square function. It is only the combined nonlinear response of the population that provides a set of basis functions, and a suitable linear combination of these approximates the desired function.

\textbf{Transfer to the next neuronal population:}

The result \(g\left( \mathbf{x} \right)\) is usually not read out as an external number but processed further by another neural population. For this, a neuron \(j\) in the target population needs an input of the form
\begin{equation}\tag{4.26}
J_j
=
\alpha_j
\mathbf{e}_j\cdot g(\mathbf{x})
+
J_j^{\mathrm{bias}}.
\end{equation}
Substituting the decoder representation of \(g\left( \mathbf{x} \right)\),
\begin{equation}\tag{4.27}
g(\mathbf{x})
\approx
\sum_i f_i(\mathbf{x})\mathbf{d}^{\,g}_i
\end{equation}
gives
\begin{equation}\tag{4.28}
J_j
\approx
\sum_i
f_i(\mathbf{x})
\alpha_j
\mathbf{e}_j\cdot\mathbf{d}^{\,g}_i
+
J_j^{\mathrm{bias}}.
\end{equation}
The synaptic weights can now be read off directly:
\begin{equation}\tag{4.29}
w_{ji}
=
\alpha_j\,
\mathbf{e}_j\cdot
\mathbf{d}^{\,g}_i.
\end{equation}
This is the crucial transition from the mathematical description of a function to the neural network. The function \(g\left( \mathbf{x} \right)\) is translated into the connection weights between two groups of neurons. The running network no longer computes the operation \(\mathbf{x} \rightarrow g\left( \mathbf{x} \right)\) as an explicit mathematical formula. The operation is embodied in the structure of the synaptic connections.

Another operation that takes place at the population level is the \textbf{multiplication of vectors}, explained here for a two-dimensional vector. Let the two-dimensional input vector be
\[
\mathbf{x}
=
\begin{pmatrix}
x_1\\
x_2
\end{pmatrix}
\]
and let the desired function be \(g\left( \mathbf{x} \right) = x_{1}x_{2}\). A population of neurons jointly represents the two components \(x_{1}\) and \(x_{2}\). The neural activities \(f_{i}\left( x_{1},x_{2} \right)\) are computed for a sufficiently large number of sample points in the permitted range of values. The target values are now the products
\[
y^{(m)}
=
x_1^{(m)}x_2^{(m)}.
\]
The corresponding function decoders are determined from the least-squares problem
\[
A\mathbf{d}^{\,g}
\approx
\mathbf{y}
\]
The network then approximately implements the multiplication
\[
(x_1,x_2)\longrightarrow x_1x_2,
\]
even though no individual neuron executes a multiplication algorithm. The computation arises from the combined nonlinear response of the neural population and the weighted combination of its activities.

\bigskip

\textbf{Branching and merging of signal paths:}

More complex computational structures can be built in the same way. The activity of one population can be passed to several downstream populations at the same time, and different decoders can produce different functions from the same representation.

Conversely, several populations can combine their results in a shared target population. In this way, larger functional systems can be assembled from elementary transformations. What matters is not a rigid layered architecture but the way populations are interconnected: each represents a particular variable, and the populations carry out defined transformations between these representations.

\textbf{Dynamical systems:}

The NEF can do more than compute static functions. Its implementation of dynamical systems is particularly important. Consider a differential equation
\begin{equation}\tag{4.30}
\frac{d\mathbf{x}}{dt}
=
f(\mathbf{x},\mathbf{u}),
\end{equation}
where \(\mathbf{x}\) is the current state and \(\mathbf{u}\) an external input. If a neural network is to implement this system, its future state must follow from its current state and the input. To achieve this, the population representing the state \(\mathbf{x}\) is connected recurrently to itself.

The recurrent connections do not simply feed the current activity back unchanged. Their weights are chosen so that the feedback, together with the synaptic dynamics, produces the desired time course of the state. The differential equation is thereby translated into a suitable recurrent transformation.

The synaptic filters play a fundamental role here. A spike does not act at a single mathematical instant; it produces a postsynaptic current that extends over time. The dynamics of these filters are therefore taken into account when the recurrent transformation is derived.

In this way, SNNs can implement integrators, control systems, memories, and other dynamic processes. This is where the particular significance of the NEF becomes clear: not only a static mapping but the desired \emph{temporal evolution of a system} can be translated into the structure of a spiking network.

\textbf{The price of approximation:}

The elegance of the NEF comes at a cost, however. The mathematical variable is represented not by a single exact computational operation but by the collective activity of a finite number of neurons. Every reconstruction, and every function derived from it, therefore remains fundamentally an approximation.

The achievable accuracy depends mainly on the number and diversity of the available tuning curves. Adding more suitably distributed neurons generally reduces the error. A finite population, however, provides only a finite set of basis functions and so limits how accurate the approximation can be.

For higher dimensions and more complex functions, the number of neurons required can grow considerably. In addition, the range over which the function is to be approximated must already be known well enough at design time. Outside this range, a good approximation is not automatically guaranteed.

The NEF thus trades part of the algorithmic computational cost for neural redundancy. A mathematical operation is not carried out step by step but approximated through the parallel activity of many simple computing elements. This is attractive on neuromorphic hardware in particular, provided that a sufficient number of very low-energy neurons is available.

\textbf{Field of application and limits:}

These properties also define where the NEF is currently applied. So far it is not a competitor to high-dimensional deep learning architectures. For tasks with very large representation spaces, the number of neurons required would quickly become a major limiting factor.

Its particular strength lies instead in a smaller but technically interesting area: tractable dynamical systems that can be described mathematically and in which real-time processing, parallelism, and energy efficiency are especially important.

In this sense, the NEF can at present be regarded as a specialized niche technology. This niche, however, is of particular interest for neuromorphic systems, because here the highest numerical precision need not be the priority. What matters is the ability to execute known dynamic functions directly in a spiking network with low energy demand.

\textbf{Construction and learning:}

The restriction to known functions does not mean that learning becomes fundamentally unnecessary in the NEF. In real tasks, the required transformations are often only partly known, change during operation, or must be adapted to an unknown environment. Predefined and learned structures can therefore be combined.

Part of the network can be constructed analytically and thus has a defined basic function from the outset. Other connections can then be adapted through local learning rules, error signals, or rewards.

This creates an interesting connection to the preceding sections. There, the central question was how a network develops its structure itself through learning. The NEF asks, in a sense, the reverse question:

\begin{center}

\emph{If the desired function is known, what neural structure must implement it?}

\end{center}

The two approaches are not mutually exclusive. It is their combination that makes it possible to specify known functional structures and to let learning determine only what is actually unknown.

\textbf{Summary:}

The regularized pseudoinverse gives the Neural Engineering Framework a highly effective tool for determining the connection weights of a spiking network for known functions in a single construction step; to that extent, iterative gradient-based training becomes unnecessary. It demonstrates that spiking neural networks need not be understood exclusively as trainable classifiers. It forms a bridge between the continuous mathematical world of numbers, vectors, and differential equations and the event-driven world of neural spikes. This bridge does not have to be built anew for every processing step. Once built, it is anchored in the connection structure of the network, the synaptic weights, and from then on it carries every transition between the world of numbers and neural dynamics.

The price is the approximation by finite neural populations and the neuron count that grows with task complexity. This limits the practical range of application of the method, but at the same time it makes clear where the method's particular significance lies: a spiking network is not only a system whose function must be found through training, but a \emph{designable computing medium whose function can be derived as well as learned}.

\subsection{4.2 The SNN landscape and Intel's Loihi system}

\subsubsection{4.2.1 The world of SNNs}

Almost every major scientific or technical university today has a research program on AI, and increasingly on neuromorphic approaches and SNN systems as well. University and industrial research programs frequently combine neurophysiological and cognitive-science studies with the development of neuromorphic approaches to the design of AI systems.

Typical results of this work can be seen in a series of neuromorphic chips and computing systems:

\begin{itemize}

\item

\emph{IBM TrueNorth (2014):} A pioneering chip with one million programmable spiking neurons \cite{merolla2014truenorth}. It demonstrated that complex neural processing tasks can be carried out with extremely low power consumption; for typical applications, figures of the order of 70 mW were reported.

\item

\emph{Intel Loihi (2017):} Loihi \cite{davies2018loihi} extended the neuromorphic approach primarily with programmable mechanisms for local synaptic plasticity and on-chip learning, that is, the adaptation of synaptic connections during operation. The thoroughly revised second generation, Loihi 2, is discussed in more detail in the next section.

\item

\emph{SpiNNaker and SpiNNaker2:} The SpiNNaker system \cite{furber2014spinnaker}, originally developed at the University of Manchester, takes a different route. A very large number of programmable ARM processor cores makes it possible to simulate large spiking neural networks in real time. The million-core machine, which went into operation in 2018, is designed to simulate networks in real time with about 1\% of the number of neurons in the human brain. SpiNNaker2 developed this concept further, with major contributions from research groups in Dresden, through new processor cores and specialized accelerators, and is designed to simulate biological structures on a scale of up to 50\% of the human brain (50 billion neurons) \cite{mayr2019spinnaker2}.

\item

\emph{IBM NorthPole (2023):} With \emph{NorthPole} \cite{modha2023northpole}, IBM presented an architecture that does not rely exclusively on spikes, and so is not an SNN in the strict sense, but that rigorously applies the principle of tightly integrating memory and processing. By largely avoiding external memory accesses, it achieved, for certain image recognition benchmarks, a reported energy efficiency up to 25 times higher than that of GPU systems built with comparable fabrication technology.

\end{itemize}

Without losing sight of the ultimate goal of broad application, these systems were initially aimed mainly at basic research, architecture testing, and large research platforms, rather than directly at end users. In recent years, however, SNN development has increasingly expanded from pure research prototyping toward systems that are practically available and in some cases offered commercially.

A crucial step toward practical use came from commercial edge vendors that actually offer neuromorphic chips or license the corresponding chip IP.

The company \textbf{BrainChip}, for instance, offers several generations of the \emph{Akida} family. The AKD1500 \cite{cnx_akd1500,brainchip_akd1500}, an edge AI co-processor based on the Akida architecture, was presented in November 2025 and has been commercially available since the end of June 2026. It shows high energy efficiency in inference and can be connected to various host systems via standard interfaces \cite{brainchip_products,brainchip_metatf}. Its neural processing model, however, does not simply correspond to the classical LIF model of the preceding chapters but is tailored to efficient event-based processing \cite{onm_akida}.

The Swiss-Chinese company \textbf{SynSense} \cite{ncore_synsense} was founded in 2017 by, among others, Giacomo Indiveri, a pioneer of neuromorphic circuit design. It does not produce general-purpose SNN chips but rather several application-specific system-on-chip solutions. One example is the \emph{Speck} chip \cite{richter2023speck}. It combines an event-driven neuromorphic processor with an event-based vision sensor and contains 328,000 spiking integrate-and-fire neurons. Its power consumption is in the sub-milliwatt range \cite{onm_speck}. Further product lines such as \emph{Xylo} are aimed at processing weak sensor signals with a power consumption of a few hundred microwatts \cite{xylo2_power}.

Other manufacturers such as \textbf{Innatera} likewise concentrate on selected applications. The \emph{Pulsar} chip is a commercially available, brain-inspired microcontroller for sensing at the network edge. It combines an SNN engine with a RISC-V microcontroller and CNN acceleration (CNN: convolutional neural network). For single always-on inference tasks, power consumption in the microwatt range is reported \cite{onm_pulsar}.

These examples also point to an important trend. The route to practical use of SNNs does not necessarily have to pass through a fully neuromorphic general-purpose computer. Instead, hybrid systems are increasingly emerging in which spiking networks, classical microcontrollers, and conventional AI accelerators work together, each contributing its own strengths.

For many potential users of SNNs, moreover, access to development platforms without neuromorphic hardware of their own is essential. Only then can new methods be developed and tested at low risk. A growing variety of solutions is now available for this. Intel gives research groups access to Loihi hardware through its neuromorphic research infrastructure; other manufacturers also provide their own development platforms, some of them cloud-based.

Intel's open-source \emph{Lava} framework supports hardware-software co-development with a particular focus on Loihi. Similar development environments are also offered for other neuromorphic platforms.

In addition, there are many options for software-only simulation on conventional hardware (see \cite{spikingjelly,ncore_awesome,qin2026mlxsnn}). Besides biologically oriented simulators, deep-learning-oriented frameworks can be used.

\emph{Nengo}, the development environment of the Neural Engineering Framework (Section 4.1.5), stands apart. It makes it possible to design neural systems analytically first and to simulate them independently of hardware. Through different backends, the resulting models can then be implemented on various execution platforms.

Finally, software tools also support the conversion of classical neural networks into SNNs and their subsequent simulation or transfer to suitable target hardware \cite{snntoolbox}. Besides training spiking networks directly, this provides another way into neuromorphic processing.

As this shows, the neuromorphic field is no longer a closed community but a tiered ecosystem of purchasable chips, research platforms, development environments, and freely available simulation tools. Access to SNNs is therefore no longer limited to a few specialized hardware developers but is open to an increasingly broad community of researchers and users.

These systems do not yet form a uniform class of computers. Their very diversity, however, shows that the basic ideas of SNNs (event-driven processing, temporal coding of information, local memory, and local learning) are now being implemented along very different technical routes; overviews and roadmaps for this field are given in \cite{schuman2022opportunities,mehonic2022brain,christensen2022roadmap}.

\subsubsection{4.2.2 Loihi 2: a neuromorphic symbiosis}

Intel's Loihi system, already mentioned above, is a particularly advanced example of how these principles can be combined in a programmable neuromorphic research architecture.

The \emph{Loihi SNN chip} \cite{intel2021loihi2} is especially well suited to illustrating the development of neuromorphic hardware, and it will later allow a comparison with classical AI architectures.

The first Loihi generation was presented in 2017, and from 2018 it was made available as a research platform for broader application testing. In 2021 it was followed by Loihi 2 \cite{orchard2021loihi2}, a thoroughly revised second generation. Together with the open software framework \emph{Lava}, it forms a research platform for developing future practical neuromorphic systems. Universities, research institutions, and industrial partners collaborate on it, for example within the Intel Neuromorphic Research Community (INRC).

Loihi 2 is manufactured in the Intel 4 process (formerly designated the 7 nm class) and contains around 2.3 billion transistors on a chip area of about \(31\,{mm}^{2}\). The chip provides 128 neuromorphic computing cores with \(2^{13} = 8192\) neurons each and can implement up to about one million logical neurons in total. The number of neurons and synaptic connections that can actually be used depends on the configuration of the particular network.

\begin{figure}[htbp]
\centering
\includegraphics[width=0.85\linewidth]{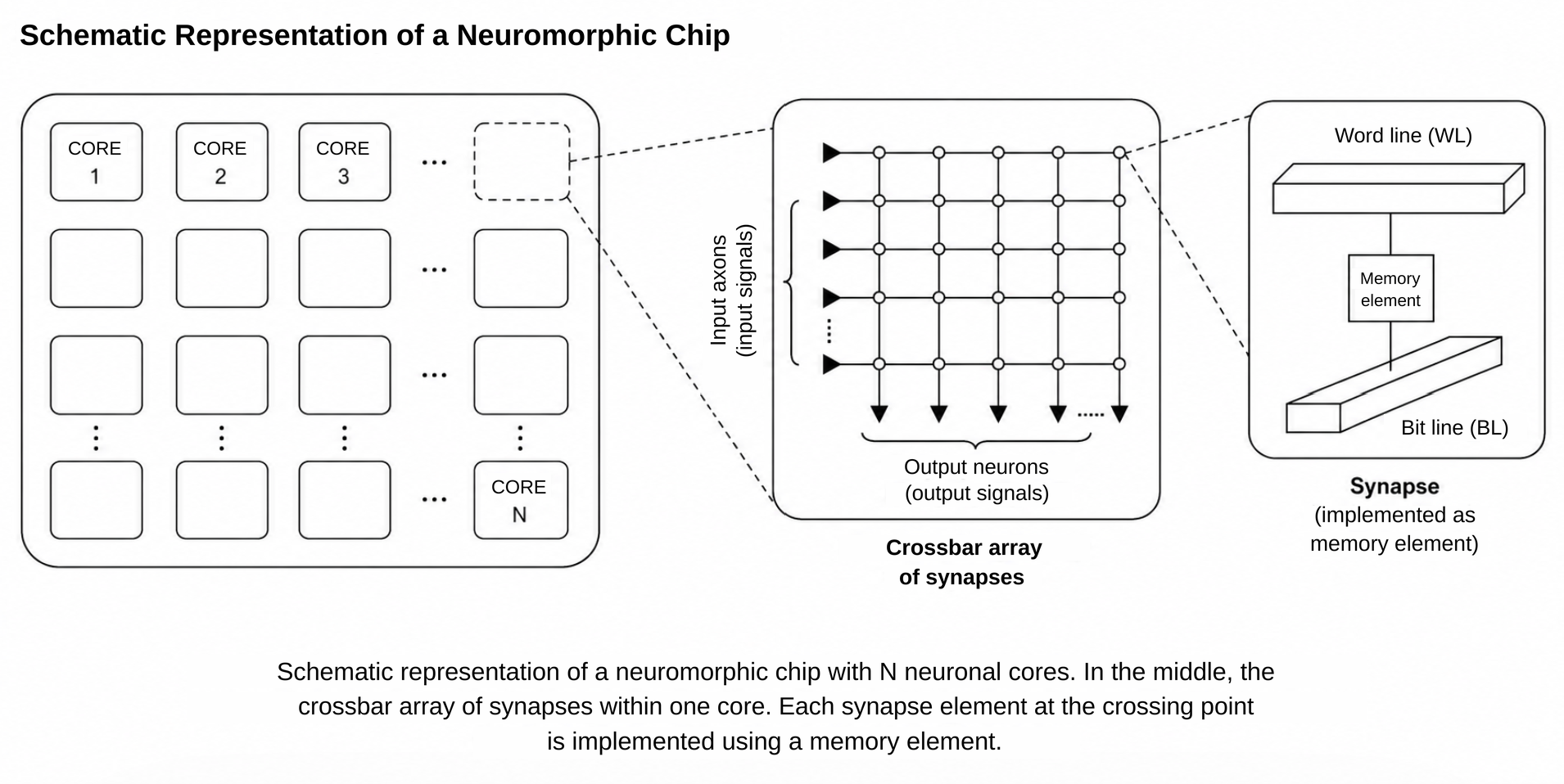}
\caption*{Fig. 4.6 Schematic representation of the Loihi\,2 system}
\end{figure}

The crossbar connection matrix shown in Fig. 4.6 is schematic. It illustrates how incoming spike events are logically assigned to the states and synaptic weights of the target neurons, but it does not correspond to a simple physical implementation of a complete crossbar for one million neurons.

SNN structures are created by programming which target-neuron inputs receive the output events of each sending neuron. Connections can run within a neurocore, across several cores, and in larger systems across several Loihi chips.

\textbf{Chip architecture:}

The basic structure of Loihi differs from the classical von Neumann architecture mainly in the tight coupling of processing, local state, and memory.

The system consists of a two-dimensional network of interconnected neurocores. Each core is a largely autonomous functional unit in which neuronal states, synaptic information, and the associated processing mechanisms are kept in close physical proximity. Synaptic weights therefore do not have to be fetched from a central main memory for every operation.

This proximity of \emph{compute} and \emph{memory} reduces energy-intensive data movement and enables massive parallelism. The cores communicate over a network-on-chip (NoC), which carries the events generated by the neurons to their respective destinations.

In keeping with the basic idea of spiking networks, neuromorphic processing is event-driven and largely asynchronous. A neurocore does not have to perform every possible neuronal operation continuously. Computation happens mainly where events occur and neuronal states need to be updated.

A substantial part of the computational effort therefore scales with the actual activity of the network. For sparse, event-based data streams in particular, this is a crucial efficiency advantage of neuromorphic architectures.

\textbf{Addressing and routing of spikes:}

Communication in such a distributed system requires an engineered counterpart to biological axonal connections. A spike event must be routed through the communication network so that it reaches the target neurons configured for the sending neuron.

Communication follows the basic principle of \emph{address-event representation} (AER). The occurrence of a spike is represented as a digital event, and the routing infrastructure delivers this event to its recipients on the basis of the configured connection information.

An individual sending neuron does not have to send a complete list of all destination addresses with every spike. The connection and routing information is configured in the hardware infrastructure beforehand. The event can then be replicated in the network and routed to the intended recipients (\emph{fan-out}).

The physical communication fabric of the network is therefore fixed at the outset. Within this fabric, however, neuronal states, synaptic weights, and with them the functional dynamics of the network can be changed.

\textbf{Graded spikes - more information per event:}

One major advance of Loihi 2 over the first generation is a more flexible representation and processing of spike events. An event no longer has to be interpreted solely as a binary pulse; it can also carry a numerical value, or payload.

In addition to the time of the event, a further quantity can be transmitted. Such \emph{graded spikes} can represent, for example, amplitude values, modulation values, or other algorithmically defined information.

The receiving unit can process this payload together with the locally stored synaptic parameters. Loihi 2 thereby adds a further dimension of information to the classical binary spike principle.

On the one hand, this is a departure from strict biological fidelity; on the other, it considerably extends the algorithmic toolbox. Temporal spike coding and numerical information transmission can be combined. This also makes it possible to implement algorithms that would be far more costly to implement with purely binary spike events.

\textbf{Local learning mechanisms:}

Another central property of Loihi is its support for programmable local learning rules. The neuromorphic cores have mechanisms for changing synaptic states on the basis of locally available variables.

These variables include pre- and postsynaptic spike events, time-dependent traces, and other state variables. This allows, for example, STDP-like learning rules and more general multi-factor learning methods to be implemented.

Loihi 2 considerably extends these capabilities. Additional information can be fed into local learning rules via events, so that modulation or error signals, for example, can be transmitted to the relevant learning units.

This does not mean, however, that Loihi 2 automatically runs classical backpropagation or a complete gradient descent locally. Rather, the hardware provides programmable mechanisms with which more complex local and error-driven learning methods can also be implemented. This ties in directly with the eligibility traces and three-factor learning rules discussed in Section 4.1.3.

\textbf{The processing pipeline:}

Fig. 4.7 gives an intuitive picture of how the circuit works. The illustration should be read as a functional scheme; in the actual hardware, the individual steps are executed in a highly parallel manner and partly by time multiplexing.

\begin{figure}[htbp]
\centering
\includegraphics[width=0.85\linewidth]{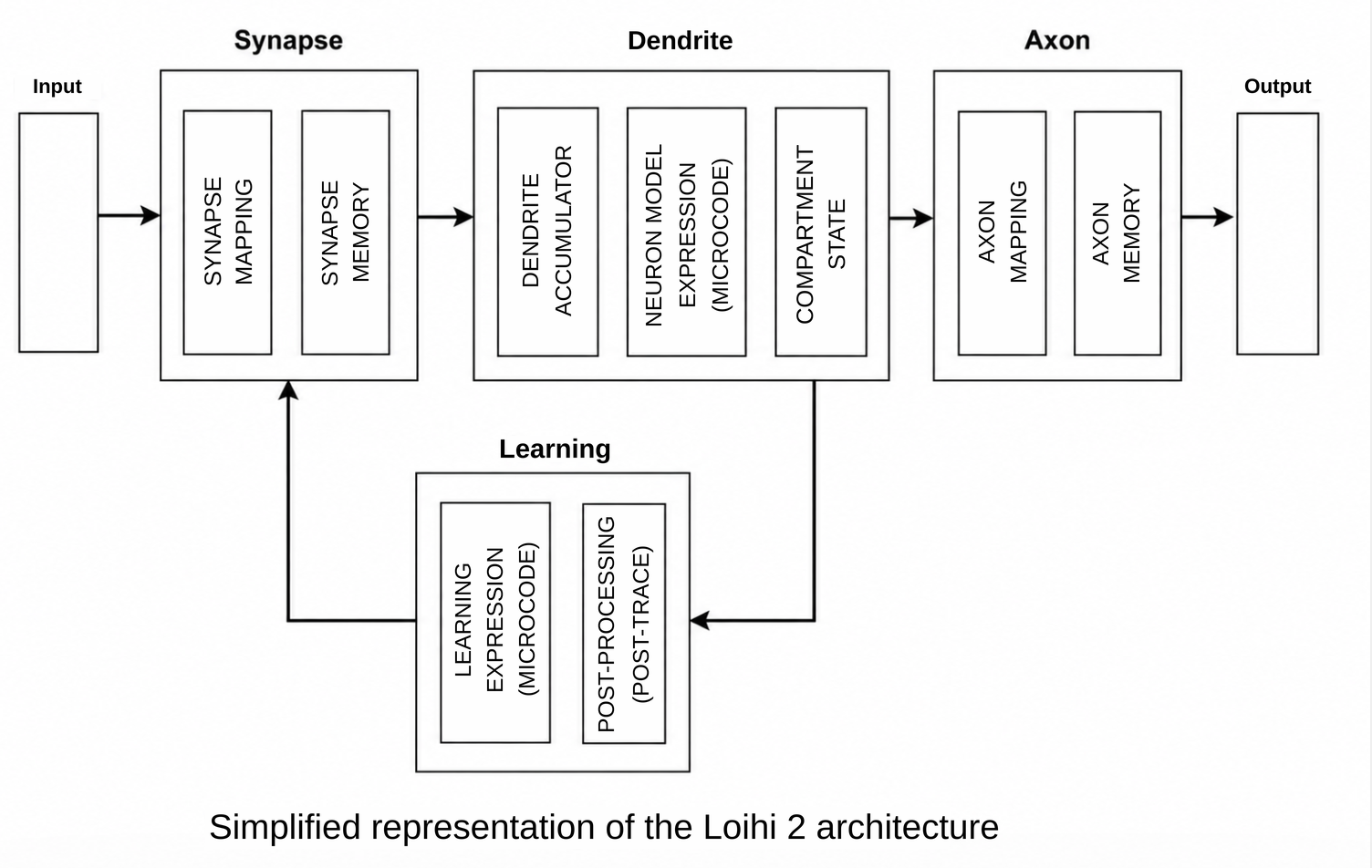}
\caption*{Fig. 4.7 The processing pipeline of a Loihi\,2 core}
\end{figure}

When a spike event arrives, the core first determines which synaptic inputs or target states the event applies to. The associated locally stored synaptic parameters are retrieved and combined with the incoming information.

The resulting contributions act on the state variables of the target neuron. In the simplest case, this corresponds to the integration of weighted input signals into a membrane potential, as in the familiar LIF model. Loihi 2, however, is not restricted to a single fixed neuron model. Because its state dynamics are programmable, it can implement different neuron models as well as dynamic computing elements that go beyond them.

If the neuronal state meets the programmed trigger condition, the neuron generates a new event. The routing infrastructure then distributes it to the configured recipients. A single event can be forwarded to many target neurons at once.

In parallel, pre- and post-traces and other local state variables can be updated. A programmed learning rule evaluates these quantities and, where required, modifies the synaptic parameters.

The result is a closed local cycle:
\[
\text{spike input}
\rightarrow
\text{synaptic weighting}
\rightarrow
\text{state change}
\rightarrow\]
\[\rightarrow\text{new spike event}
\rightarrow
\text{update of learning states}.
\]
Processing and learning can therefore be tightly coupled and, in principle, carried out during normal operation. This is a fundamental difference from classical computer architectures, in which training, parameter storage, and inference are usually kept much more strictly separate.

\textbf{A symbiosis of biological and engineering principles:}

Loihi 2 does not attempt to reconstruct a biological nervous system in the greatest possible detail. Instead, it adopts selected functional principles from biology and combines them with the capabilities of digital circuit technology.

From biology it takes, in particular,

\begin{itemize}

\item

event-driven processing,

\item

massive parallelism,

\item

local memory for neuronal and synaptic states,

\item

communication via spike events, and

\item

the possibility of local plasticity.

\end{itemize}

Engineering adds precise digital routing, programmable neuron models, additional event payloads, and scalability over a digital communication network.

It is this combination that makes Loihi 2 a neuromorphic symbiosis: the architecture does not copy the brain, but it uses biological organizational principles where they offer computational advantages and supplements them with engineering methods where those are more powerful or more flexible.

Loihi 2 and the larger systems that have since grown out of it mark a major step from the theoretical idea of spiking networks to a scalable neuromorphic computer architecture.

\textbf{Communication, routing, and emergence: a remark}

Loihi 2 is an extraordinarily flexible research platform for building a wide variety of SNN structures, neuron models, and learning methods. Its architecture is optimized not primarily for the fastest possible execution of dense matrix multiplications but for sparse, event-driven, and time-dependent computations.

One structural difference from the biological brain stands out, however: the physical communication and routing fabric of an implemented network is configured at the outset. Synaptic plasticity can change the functional significance of these connections, but it cannot create arbitrary new physical communication paths outside the routing structure provided.

For fully self-organizing systems, this can in principle be a limitation. A communication path excluded by the original network configuration cannot emerge spontaneously through later local plasticity either. On the other hand, sufficiently rich and redundant initial structures can provide a very large space of possible functional interconnections in which emergent dynamics can indeed arise.

The key open question is therefore not so much whether \emph{pre-routing} prevents emergence in principle, but how open the predefined structure must be for the network to develop, in the course of operation, new functional forms of organization that were not explicitly designed beforehand.

\subsubsection{4.2.3 The path to practice}

The application studies to date confirm that neuromorphic systems are strongest where information arrives in a time-dependent, sparse, and event-driven form. This applies in particular to real-time processing of sensor data, sensor fusion, robotics, and event-based image processing.

In a range of applications, Loihi systems have shown considerable advantages over conventional processor architectures. For selected inference and optimization tasks, Intel reports up to 100 times lower energy consumption and up to 50 times higher processing speed than conventional CPU and GPU systems. Such figures, however, depend heavily on the task, the comparison hardware, and the implementation, and they should not be read as a general performance factor for SNNs.

One especially natural use is in vision systems for high-speed object tracking and gesture recognition, where only changes need to be processed as events. Adaptive prosthetics, wearable gesture recognition, robust speech recognition in noisy environments, artificial cochleas, and the classification of signals from olfactory sensors are further applications in which event-driven processing promises particular advantages.

Individual studies have reported very large improvements over conventional solutions, for example about 200 times lower energy consumption and ten times lower latency in wearable gesture and vision applications.

These results are probably characteristic of an important practical application area for small and medium-sized SNNs: low-power mobile systems, controllers, and monitoring tasks in which continuous data streams are mostly redundant and only isolated, precisely timed events need to be processed. The commercial edge products described in Section 4.2.1 confirm this trend, which may also become important for the use of SNNs as peripheral agents in world-model-based systems.

The capabilities of neuromorphic systems are not confined to such applications, however. Researchers are increasingly asking whether their event-driven, highly parallel mode of operation can also be applied to large scientific computing problems. Studies using Loihi-based algorithms address, for example, physical simulations, heat propagation, optimization problems, and turbulence modeling \cite{Talha2025Exploring}. For real-time turbulence simulations, for instance, more than an order of magnitude higher computational efficiency and 34 times lower energy consumption than CPUs have been reported, along with 1.3 to 1.4 times higher efficiency and 10 times lower energy consumption than GPUs. The demand for mobile, energy-saving simulation tools that can be used in the field is often what motivates work on such applications.

A second line of development is therefore emerging. Neuromorphic systems are no longer seen solely as very low-power computers for small sensor systems; increasingly, they are also being examined to determine whether their advantages can be scaled up to highly complex computing problems.

\textbf{SNNs and language models:}

Spiking principles have also been applied to language models. An early example is \emph{SpikeGPT} \cite{zhu2023spikegpt}. The model was trained in variants with 45 million and 216 million parameters and, according to its authors, was at the time of publication the largest SNN model trained with backpropagation.

SpikeGPT is inspired by the RWKV architecture (RWKV: receptance weighted key value, an architecture that combines the advantages of classical transformers and recurrent neural networks). It replaces multi-head self-attention, which scales quadratically with sequence length, with a recurrent, linearly scaling mechanism built from spiking activation units. This makes the model suitable both for language generation and for language understanding tasks.

In initial experiments, performance on the benchmarks examined was competitive with that of comparable non-spiking models. At the same time, the authors report about 20 times fewer elementary operations, provided that processing runs on hardware that can actually exploit the sparse, event-driven activations.

\emph{SpikingBERT} went on to show that BERT-like spiking language models can also be trained by knowledge distillation from classical models \cite{bal2024spikingbert}.

These early results are still clearly far from the scale of today's large language models. They are nevertheless significant as a proof of principle: time-dependent spiking processing is not confined to classical sensing and control tasks but can, in principle, also be transferred to sequential symbolic information processing.

In \cite{abreu2025loihillm}, a matrix-multiplication-free architecture adapted to Loihi 2 chips is reported and used to implement an LLM with 370 million parameters. Initial results show up to three times the throughput at only half the energy consumption of a transformer-based LLM on edge GPUs. This is an impressive demonstration of the potential for fast, efficient inference when generating long texts on neuromorphic hardware.

The work on implementing world models with SNNs is particularly interesting. Spiking world models already exist, and they reach parity with ANNs. Sun et al.~\cite{sun2025spikingwm} report the construction of a true spiking world model with dendrite-like multi-compartment neurons and its evaluation on the DeepMind Control Suite. It outperforms earlier SNN approaches and is comparable with GRU-based ANN world models.

The starting point of this remarkable approach is a more accurate treatment of the complex integration of electrical potentials within the neuronal structure, combined with a neuron's ability to process long sequences: a fine example of successful neuromorphic thinking.

\textbf{SNNs for neuromorphic research:}

With the \emph{Hala Point} project, Intel continued the scaling begun with Loihi on a much larger scale. The system, first installed at Sandia National Laboratories in 2024, combines 1152 Loihi 2 processors and supports up to 1.15 billion artificial neurons and 128 billion synapses. Its neuron capacity is of the same order of magnitude as an owl's brain, or the cortex of a capuchin monkey.

Hala Point serves as a research platform for investigating neuromorphic methods at an unprecedented scale. Alongside new AI methods, the focus is on scientific simulations, optimization problems, and the question of how the efficiency advantages of neuromorphic processing can be carried over to very large systems.

The significance of this step therefore lies less in having already created a neuromorphic general-purpose computer. What matters is that it is now possible, for the first time, to investigate which properties of spiking systems emerge only at very large network sizes. This opens up to experiment a question of central importance for the further development of SNNs: whether the particular advantages of event-driven, temporally coded networks are preserved as the networks grow, or whether new capabilities actually emerge.

\emph{In summary, the Loihi development and comparable projects have convincingly demonstrated the technical feasibility of efficient, energy-saving SNN computing systems. So far, their particular strengths lie mainly in the energy-efficient processing of time-dependent, sparse data streams. At the same time, increasing scalability, programmable local learning mechanisms, and initial applications to more complex computing and language problems open the way to substantially larger applications.}

General-purpose programmable neuromorphic components are therefore available today, on which application-specific SNN learning machines with local processing and online learning capability can be developed. Whether an independent class of computers for highly complex AI tasks will emerge from this in the long run remains open; but this scale is now experimentally accessible.

%% file: chapters/05_zusammenfassung.tex
\section{5. Summary}

\enlargethispage{\baselineskip}

Two major lines of development currently shape connectionist artificial intelligence. On one side is the world of established artificial neural networks, which now has an enormous body of experience and methods for learning, abstraction, network design, and scaling. On the other side, a second world is developing around spiking neural networks, one that follows the operating principles of biological nervous systems more closely.

The two lines of development are not independent of one another; they are beginning to complement each other. Classical AI provides powerful mathematical methods for learning and optimization. Neuroscience is uncovering the organizational principles by which biological nervous systems process and store information at extremely low energy cost and use it to adapt their behavior. Neuromorphic engineering attempts to carry suitable principles from both worlds over into new computer architectures.\footnote{Maturing advances in bringing quantum computing into practical use, for example, may influence further developments.}

Using text generation by large language models as an example, this book first showed the level of performance that classical neural networks have now reached. Transformers process information in high-dimensional representation spaces in which relations between features and contexts are formed during training. The attention mechanism makes it possible to weight dynamically those relations among the representations that are relevant to the current context.

This performance comes at an enormous computational cost, however. Global training by backpropagation in particular requires very large quantities of data to be processed repeatedly and gradients to be computed across the entire network. As models grow, the hardware, energy, and infrastructure costs rise to levels that are increasingly becoming a critical challenge even for the further development of classical AI.

Biological neural networks offer a different perspective. The brain has no central controller that computes a global error gradient for billions of synapses. Instead, information is processed through the activity of large, interconnected assemblies of neurons, while learning and structure formation rely largely on local processes.

A fundamental mechanism here is spike-timing-dependent plasticity (STDP). It links the temporal causality of neuronal events to the local modification of synaptic connections. Recurring temporal relationships can stabilize preferred signaling pathways, so that larger functional structures can arise from local plastic processes, a basic example of self-organization in neural systems.

Eligibility traces add a further form of local memory. Past activity leaves a temporary trace at a synapse, so that an evaluative or reward signal arriving later can still act on the connections that were involved in an earlier processing step. This creates a possible bridge between strictly local learning and higher-level evaluation, without a full gradient necessarily having to be backpropagated through the whole network.

Spiking neural networks carry selected principles of this kind into engineered systems. They add a further fundamental dimension to the toolbox of connectionist AI: time itself becomes a carrier of information.

Whereas classical neural networks process information predominantly as numerical values in synchronous, sequential computational steps, SNNs can additionally represent information through the timing, the order, and the temporal structure of spike events. Temporal codes in particular can carry a potentially high information content even in a single event. Exploiting them in practice, however, places high demands on temporal precision, robustness, learning methods, and hardware.

Processing is event-driven: computational operations are triggered mainly where neuronal activity actually occurs. With sufficiently sparse event streams, a large part of the operations that conventional architectures execute constantly can be dispensed with. Considerable advantages in energy consumption and latency have therefore been demonstrated, particularly in sensing, control, and other time-dependent tasks.

The initial skepticism about whether spiking networks could be made to work in practice has since given way to an extensive methodological repertoire. It ranges from the conversion of classically trained networks, through STDP and local trace-based learning rules, to surrogate-gradient methods. In addition, there are methods for network stabilization and for targeted architectural pre-structuring, and, with the Neural Engineering Framework, even methods for translating known mathematical functions and dynamical systems directly into neuronal populations and their interconnections.

What emerges from this is not a replacement of classical ANNs by SNNs. Rather, a continuum of approaches is emerging, ranging from fully specified networks through globally trained ones to locally self-organizing networks. Depending on the task, mathematically constructed structures, classical training, and local neuronal plasticity can be combined.

Neuromorphic hardware platforms show that this approach is technically feasible. Intel's Loihi project in particular demonstrates how event-driven communication, programmable neuronal dynamics, local states, and local learning mechanisms can be brought together in a scalable digital computer architecture. With systems capable of implementing more than a billion neurons, it has now also become technically possible to investigate very large neuromorphic networks.

At the same time, specialized neuromorphic processors are beginning to be used commercially, in particular for low-energy edge and sensor applications. Developments so far also show that the future need not lie in an exclusively neuromorphic general-purpose computer. Hybrid systems can deploy classical processors, conventional AI accelerators, and spiking networks wherever each has its particular strengths.

It is still unclear what role SNNs can take on in highly complex, higher-level AI tasks such as knowledge-based system integration, the construction of world models, or the creation of autonomous agents. Early work on spiking language processing shows that time-based neural architectures can also be transferred to complex sequential and symbolic tasks. Such systems are, however, still a long way from the scale and performance of today's large language models. On the other hand, first SNN world models, and first uses of SNNs in substantial roles within ensembles of autonomous agents, already indicate that spiking systems could be given more than a marginal, supplementary role here.

It matters for further development that what engineering has so far harnessed of the structural and methodological potential of biological nervous systems captures only a small fraction of their organizational principles. Neuromodulation, different forms of synaptic and intrinsic plasticity, memory consolidation, structural plasticity, and the interaction of local and global control mechanisms are known in principle, but their overall effect is still far from understood.

This may be a key path for the further development of artificial intelligence. The crucial lesson from biology may not be to reconstruct the brain as faithfully as possible. It seems more promising to understand its general organizational principles and to transfer them to engineered systems that enable new forms of efficient, adaptive, and self-organizing information processing.

Of course, the development and acceptance of new research directions is never a purely scientific question. It is embedded in a diverse landscape of economic, social, and individual positions and interests, which can have a considerable influence.

The AI gigafactories currently being built show, for example, the enormous technological and energy expenditure required to continue along the chosen path of ever larger classical models. There are two reasons to doubt that there is no alternative to this path. First, the limits to the seemingly unlimited scaling of computing power and energy consumption can become a strong incentive to push forward the development of SNNs and other neuromorphic architectures and to seek fundamentally more efficient forms of learning and computing. Second, the foreseeable next generation of commercially available neuromorphic hardware can be expected not only to open a first practical route to implementing large classical AI systems such as LLMs on energy-efficient spiking networks, but also to make hardware and software development for SNNs a broader field of foundational AI development.

The technological basis of information processing is itself evolving, and this will feed into the development of classical and neuromorphic systems alike. Optical computing is currently gaining momentum of its own: in 2025, for the first time, special-purpose photonic processors that execute matrix operations by the interference of light ran large network models with accuracy close to that of electronic hardware \cite{harris2025universal}, and such processors are now being offered commercially \cite{qant2025photonic}. They address the energy problem of classical ANNs at the component level, by making each individual dense operation cheaper, while SNNs address it at the architecture level, by avoiding unnecessary operations in the first place. The two approaches are not mutually exclusive: with excitable laser neurons and photonic synapses, experimental building blocks of fully optical SNNs already exist that combine event-driven sparsity with spike times in the sub-nanosecond range \cite{feldmann2019alloptical,shastri2021photonics}. A new substrate, however, will not replace the many options offered by neuromorphic design (sparse computation, local plasticity, temporal coding); it will open up additional routes to their implementation.

\emph{In summary, two major lines of development are currently converging: the mathematically and technologically highly developed world of classical artificial neural networks, and the world of time-based, locally learning, and neuromorphically organized systems, whose technological exploitation is only beginning. In the long run, their combination could matter more than the further development of either direction on its own.} \emph{It can only be conjectured whether systems that fundamentally extend today's artificial intelligence not only in energy efficiency but also in learning capacity, adaptability, and autonomous structure formation will actually emerge from this. Biological systems show, however, that such principles are fundamentally capable, and neuromorphic development is beginning to make them accessible to engineering.}

The further development of artificial intelligence, however, is not a technical question alone. As its capability grows, and with it the capital and infrastructure it requires, its societal significance continues to grow as well. It therefore remains to be hoped that this development will not be subordinated predominantly to the interests of individuals in increasing their power and capital but will serve the goal of providing knowledge and technical capabilities for the benefit of as many people as possible.